\documentclass{article}

\usepackage{arxiv}
\usepackage[utf8]{inputenc} 
\usepackage[T1]{fontenc}    
\usepackage{url}            
\usepackage{booktabs}       
\usepackage{amsfonts}       
\usepackage{xcolor}
\usepackage{graphicx}
\usepackage{amsmath}
\usepackage{upgreek}
\usepackage{amsthm}
\newtheorem{definition}{Definition}
\newtheorem{theorem}{Theorem}
\newtheorem{proposition}{Proposition}
\newtheorem{lemma}{Lemma}

\usepackage{tikz}
\usetikzlibrary{matrix}

\title{Is K-fold cross validation the best model selection method for Machine Learning?}

\author{
  J.M. Gorriz, R. Martin Clemente, F. Segovia, J Ramirez\thanks{gorriz@ugr.es, jg528@cam.ac.uk} \\
  Department of Signal Theory and Communications\\
  University of Granada\\
  Granada, Spain \\
   \And
 A. Ortiz \\
  Department of Communications Engineering\\
  University of Malaga\\
  Malaga, Spain\\
  \texttt{aortiz@ic.uma.es} \\
     \And
 John Suckling \\
  Department of Psychiatry\\
  University of Cambridge\\
  Cambridge, UK\\
  \texttt{js369@cam.ac.uk} \\
       \And
       International Initiatives\\
for the Alzheimer's Disease Neuroimaging Initiative (ADNI)\\
}

\begin{document}
\maketitle

\begin{abstract}
As a technique that can compactly represent complex patterns, machine learning has significant potential for predictive inference. K-fold cross-validation (CV) is the most common approach to ascertaining the likelihood that a machine learning outcome is generated by chance, and it frequently outperforms conventional hypothesis testing. This improvement uses measures directly obtained from machine learning classifications, such as accuracy, that do not have a parametric description. To approach a frequentist analysis within machine learning pipelines, a permutation test or simple statistics from data partitions (i.e., folds) can be added to estimate confidence intervals. Unfortunately, neither parametric nor non-parametric tests solve the inherent problems of partitioning small sample-size datasets and learning from heterogeneous data sources. The fact that machine learning strongly depends on the learning parameters and the distribution of data across folds recapitulates familiar difficulties around excess false positives and replication. A novel statistical test based on K-fold CV and the Upper Bound of the actual risk (K-fold CUBV) is proposed, where uncertain predictions of machine learning with CV are bounded by the \emph{worst case} through the evaluation of concentration inequalities. Probably Approximately Correct-Bayesian upper bounds for linear classifiers in combination with K-fold CV are derived and used to estimate the actual risk. Additionally, the origins of the replication problem are demonstrated by modeling and simulating common experimental circumstances, including small sample sizes, low numbers of predictors, and heterogeneous data sources. The performance with simulated and neuroimaging datasets suggests that K-fold CUBV is a robust criterion for detecting effects and validating accuracy values obtained from machine learning and classical CV schemes, while avoiding excess false positives.
\end{abstract}


\section{Introduction}

Inflated type I error rates, p-hacking\footnote{The deliberate selection of the level of significance in the seek of the desired effect}, bad experimental designs, etc. have all become problematic affecting reproducibility and replicability across science and engineering \cite{NAS2019}, including neuroimaging \cite{Eklund16,Noble20}. For example, between-group analyses in neuroimaging (a binary classification task) are typically conducted with the General Linear Model (GLM) and classical statistics \cite{Friston95}. Although straightforward to interpret \cite{Friston02}, the assumptions of the univariate GLM are frequently violated \cite{Rosenblatt14} making the classical tests more conservative and encouraging less than optimal analytic practices. 

Machine learning (ML) systems are agnostic models \cite{Ribeiro2016} that have clear potential for processing complex datasets. In other words, they are designed to solve tasks without knowledge of the underlying distributions of the data \cite{Lecun15}. The unknown probability density function (pdf) in high-dimensional spaces, from which data samples are drawn, is modeled by fixing the ML system complexity, typically the number of parameters of the neural network architectures \cite{Grohs22}, and optimising the parameters through learning on a training set. At the decision stage, high dimensional features are commonly projected onto low-dimensional sub-spaces \cite{vandermaaten08} easing the learning task and permitting inference. These multivariate, data-driven approaches offer an attractive alternative with high detection rates available in several research fields, e.g. neuroscience \cite{Mouro-Miranda05,Zhang14,Gorriz2021b,Gorriz2022}.  In this context, migration to ML \cite{Wang07,Wang09} does not preclude false positives and misleading results. For example, outputs from some ML systems with multivariate features are strongly affected by high-noise levels in the data \cite{Jollans2019,McKeown03}. Moreover, although questionable research practices such as p-hacking are not problematic for ML, there is a strong dependence of performance on the construction of the learning and validation datasets \cite{Gorgen18}. 

Cross Validation (CV) with limited sample sizes working with heterogeneous datasets can still lead to large standard errors \cite{Varoquaux18,Gorgen18}. In the context of small sample-sizes and heterogeneous data, CV strongly underestimates the actual risk, which can be interpreted as the violation of the assumption of \emph{ergodicity} \cite{Gallavotti1995}; that is, the average behavior (performance) of any system can be described from a collection of random samples (a set of random performances in each CV loop). What would happen if our process was not ergodic? We can make use of the concept of a stable inducer \cite{Kohavi95} to describe the consequential effect. An inducer is the algorithm that creates a classifier from a training set. It is stable for a given dataset if it creates classifiers that make the same predictions when given perturbed datasets. If the sample is strongly perturbed, e.g. due to small-sized and heterogeneous samples drawn from multivariate distributions, the process of learning from specific training folds cannot be efficiently extrapolated, in statistical terms, to test folds.

\subsection{Related work}

System performance in ML is frequently assessed using CV techniques \cite{Allen74,Geisser75}, which divide the overall data into partitions (i.e., folds) and then systematically train and test the ML system on each fold, averaging the resulting outcome measures. Although several meritorious approaches have been undertaken to estimate the CV prediction error and its confidence intervals, they were evaluated under ideal conditions using model-driven approaches, e.g. homoskedascity, Gaussianity in the model error \cite{Bates2023}, the independent decomposition of the variance of error when sample size is large \cite{Rodriguez2010}, etc. Moreover, confidence intervals for a broad class of estimates of the out-of sample prediction error, such as data splitting, nested cross validation or bootstrapping, are usually derived based on the desired quantile of the standard normal distribution. 

Nonetheless, there have been several commentaries in the literature \cite{Varoquaux18,Eklund16,Jollans2019,Gorriz18} about the high variability (or variance in statistical terms) of performance found across CV folds in several neuroimaging data analyses with clear implications for predictive inference. However, no quantitative solution has been given to this issue beyond increasing sample size and warning of errors across CV folds. 

Permutation analyses \cite{Phipson2010} can augment ML systems to test for statistical significance by the computation of p-values from the distribution of a statistic under the null hypothesis. Assuming the data is i.i.d., this null distribution is obtained by permuting the class labels a large number of times and training and testing the ML system each time. However, the estimation of the statistics could be biased with heterogeneous datasets due to the use of only a single instance of the K-folds in the CV, limited amounts of data, overfitting, etc. A high degree of confidence in the results cannot be obtained if the essential element of these analyses, the predictive inference based on an averaged accuracy from specific folds, is flawed. 

\subsection{Our contribution}

Although, no closed solution to this common problem in data analysis has been found in the literature, several results within statistical learning theory (SLT) could alleviate the issue by defining confidence intervals without classical model assumptions \cite{Vapnik98,Boucheron13}. In this case, we change the inference paradigm by combining the out-of-sample prediction error (from testing folds) with inferences on the whole sample, thus avoiding the splitting of the small sample. In this sense, we construct more conservative upper bounds (confidence intervals) by assuming the maximum deviation of the actual risk or error from the empirical error to validate learning models.

In this paper we propose the use of K-fold CV in combination with a non-parametric statistical test based on the analysis of the \emph{worst case}. This method, named K-fold Cross Upper Bounding Validation (CUBV) allows control over the performance of the system. Given the empirical K-fold CV-error we calculate the upper-bound of the actual risk to assess the degree of significance of the classifier's performance. Upper-bounding the actual risk is a statistical learning technique based on concentration inequalities (CI) that constructs efficient algorithms for prediction, estimation, clustering, and feature learning. CI deal with deviations of functions of independent random variables from their expectations; here, the deviation of empirical errors in each training fold from the real error achieved by the classifier given the complete dataset. This solution for evaluating error variability produces protective confidence intervals to control the excess of false positives in the inference problem. In this sense, it is similar to multiple comparisons p-value corrections, such as Bonferroni or Random Field Theory, which are used within hypothesis-driven methods for statistical inference \cite{Frackowiak04, Nichols12}.

\section{K-fold Cross Validation: theory and practice}\label{sec:methods}

\subsection{Cross Validation: setting and notation}\label{sec:CVmethods}

We consider the supervised learning setting for binary classification with the input and output spaces $Z=(\mathcal{X},\mathcal{Y})\in(\mathbb{R}^n,\{0,1\})$, a class $\mathcal{F}$ of functions $f:\mathcal{X}\rightarrow \mathcal{Y}$ and $\mathcal{P}(Z)$ the space of all probability distributions on $Z$. We aim to predict any $y\in\mathcal{Y}$ from $\mathbf{x}\in\mathcal{X}$ using a function $f^*$ that minimizes the expected loss:
\begin{equation}
   \mathcal{R}(f)=\mathbb{E}_{Z\sim \mathcal{P}}[\mathcal{L}(y,f(\mathbf{x})]=\mathbb{P}(f(\mathbf{x})\neq y) 
\end{equation}
where the loss is usually selected as the squared error $\mathcal{L}(y,f(\mathbf{x}))=(y-f(\mathbf{x}))^2$. 

However we only observe $N$ i.i.d. pairs $Z^N=(X,Y)$ sampled according to an unknown distribution $P\in\mathcal{P}(Z)$, thus the actual risk cannot be exactly determined. Instead, classical theory assesses other estimands from statistical measures with fixed-size datasets, such as the predictive error given the observed data and labels or the average error of the fitting function on same-sized datasets: 
\begin{equation}
\begin{array}{l}
     \mathcal{R}_{XY}(f)=\mathbb{E}_{(\mathbf{x},y)\in Z^N\sim P}[\mathcal{L}(y,f(\mathbf{x}))|(X,Y)]  \\
        \mathcal{R}_N(f)=\mathbb{E}_{Z^N\sim P}[\mathcal{R}_{XY}(f)] 
\end{array}
\end{equation}
Based on estimations of these measures, we approximate $f^*$ using the Empirical Risk Minimization (ERM) algorithm, e.g. $f_N=\underset{f\in\mathcal{F}}{ \arg\min }\mathcal{R}_N(f)$. Under this setting, we validate the model performance using the K-fold CV estimated error. In CV we partition $Z^N$ into $K$ mutually exclusive and stratified subsets or folds $D_i$, $i=1\ldots,K$ at random and fit a model for each of them $f_{N/K}^i$. The CV estimated error is given by:
\begin{equation}\label{eq:CVloss}
\hat{\mathcal{R}}_N(f)=\frac{1}{N}\sum_{i;(\mathbf{x},y)\in D_i} 1-\delta(f_{N/K}^i(\mathbf{x}),y)
\end{equation} 
where $\delta(.)$ is the Kronecker´s delta. Typically, $K = 10$ is used, representing a trade-off between the variance and bias in the estimation of the error. Another common choice is $K = N$, known as Leave-One-Out (LOO) cross-validation.

It has be recently shown that $\hat{R}_N(f)$ is a better estimator of $\mathcal{R}_N(f)$ than $\mathcal{R}_{XY}(f)$ assuming the homoskedastic Gaussian model \cite{Bates2023}. On the other hand, a more accurate estimation of the error using CV would include all the possible combinations, $\binom{N}{\frac{N}{K}}$, although this is usually too computationally expensive \cite{Kohavi95}. One solution to this issue is resampling the available data (a single instance from the unknown $P$) with $F$ repetitions and averaging the resulting outcomes as shown in the experimental part. This technique is commonly referred to as bootstrapping \cite{Efron1993}. Another possibility is a nested cross-validation procedure that provides theoretical confidence intervals by combining the corrected inner CV error estimation of a regular K-fold cross-validation using $K-1$ folds (a smaller subset) and the predicted error on the remaining fold \cite{Bates2023}. With these kinds of model-driven approaches, the expected deviation of the nested CV error with respect to the actual out-of-sample error is accurate for simple data distributions and slightly lower than the naive CV using real datasets. However, the computational cost is a significant drawback, as nested CV requires running multiple cross-validation processes, making it substantially more resource-intensive compared to naive CV.

Nevertheless, the main goal in model selection is to evaluate how well the fitting function on a finite dataset would perform on an infinite out-of-sample set. SLT allows us to provide conservative (and simple) data-dependent upper bounds for the actual risk $\mathcal{R}(f)$, which is the error of the fitting algorithm applied to the entire population given $P$. This is essential when model assumptions are not valid, such as when using real data, and CV confidence intervals fail.

\subsection{Performance of K-fold Cross Validation in Common Experimental Designs}\label{sec:preliminary}

The following examples illustrate the key problems found in the literature that affect the reproducibility of ML results in neuroimaging and other related areas. In summary, contemporary results are based on a single instance of the pdf from the training folds as well as the selection of a particular CV fold as the test set with which predictive inference is assessed. Thus, even with the same datasets, the performance of one study informed by another can differ substantially with potentially contradictory interpretations if the data folds are not identical.  

\subsubsection{The null experiment – control of type I errors}\label{sec:null}

\begin{figure*}
\centering
\includegraphics[width=0.32\textwidth]{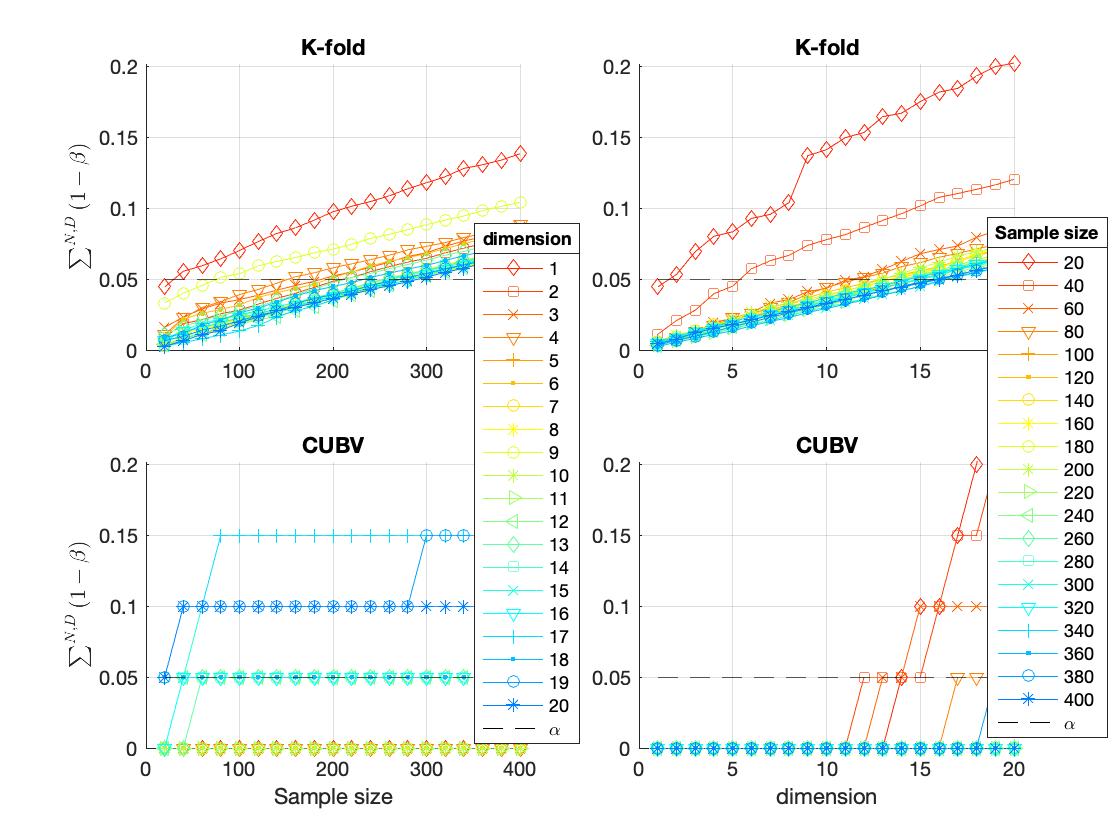}
\includegraphics[width=0.32\textwidth]{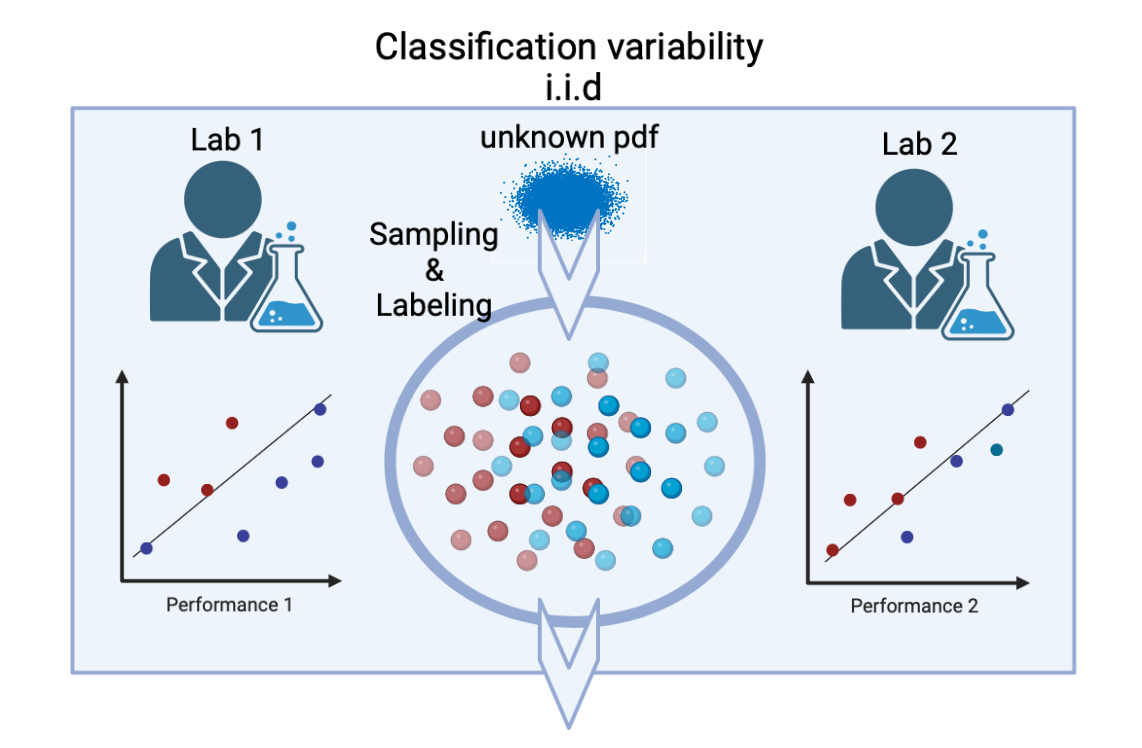}
\includegraphics[width=0.32\textwidth]{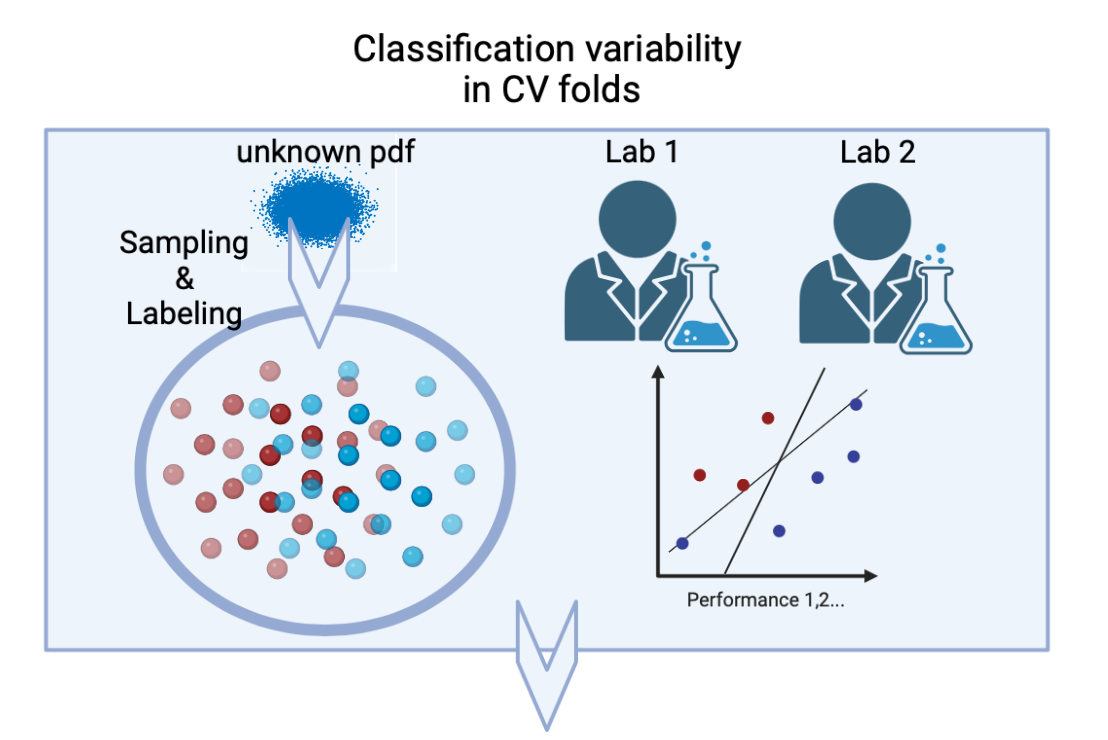}\\
\includegraphics[width=0.32\textwidth]{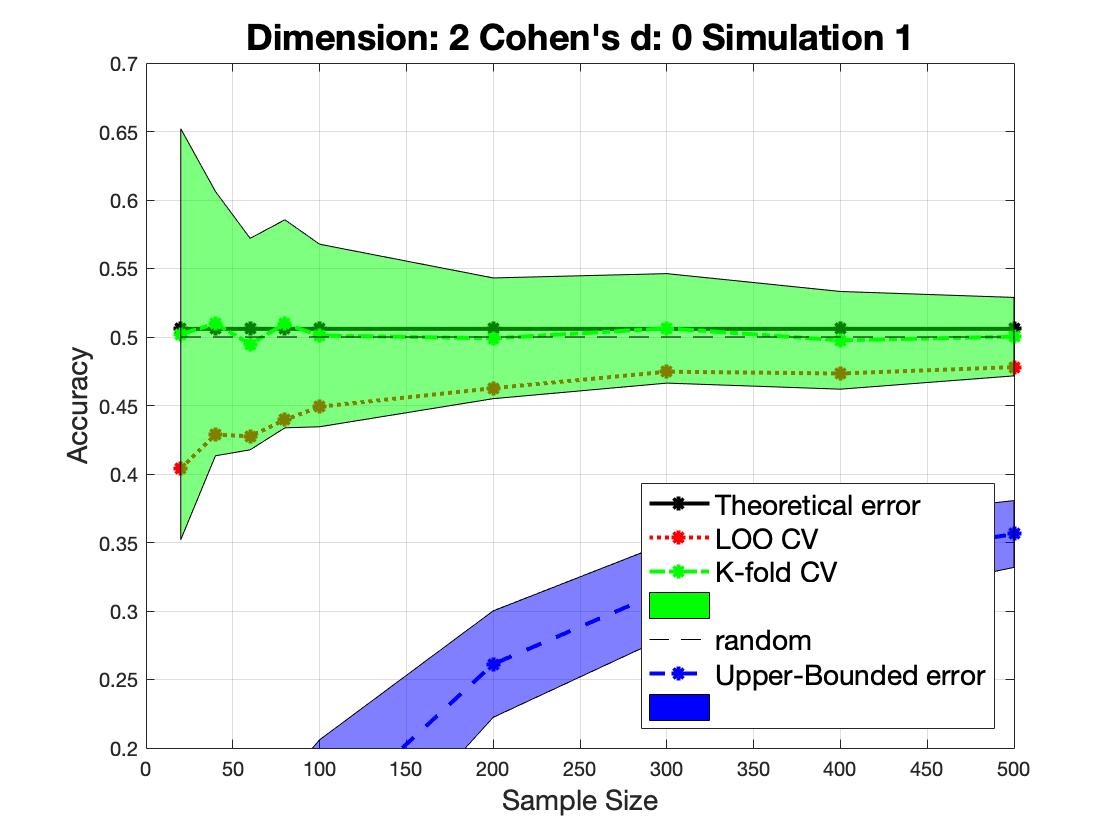}
\includegraphics[width=0.32\textwidth]{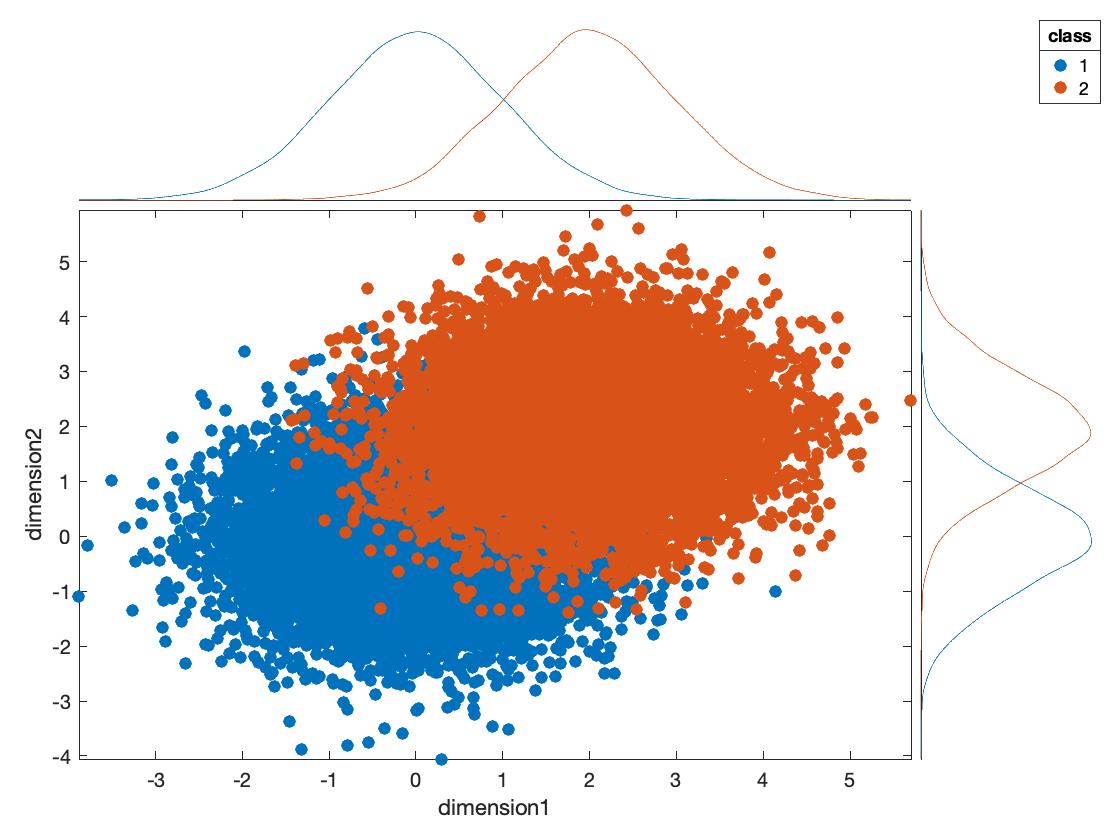}
\includegraphics[width=0.32\textwidth]{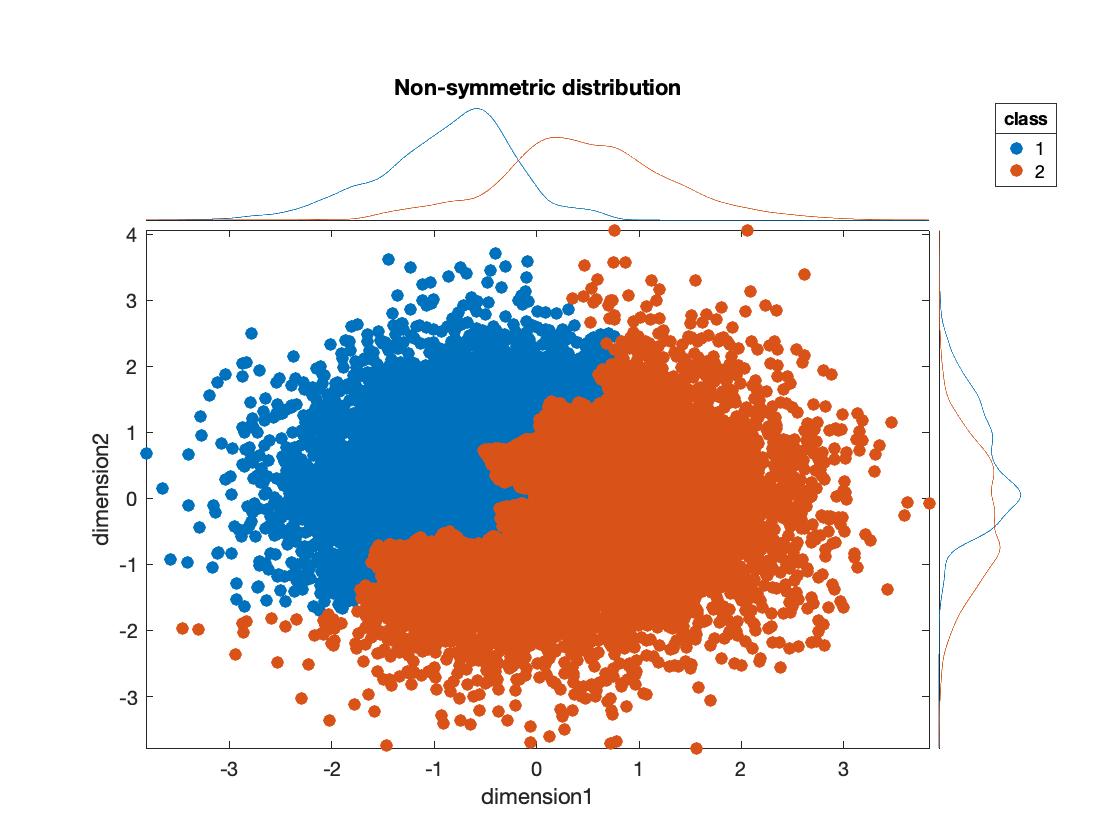}
\includegraphics[width=0.32\textwidth]{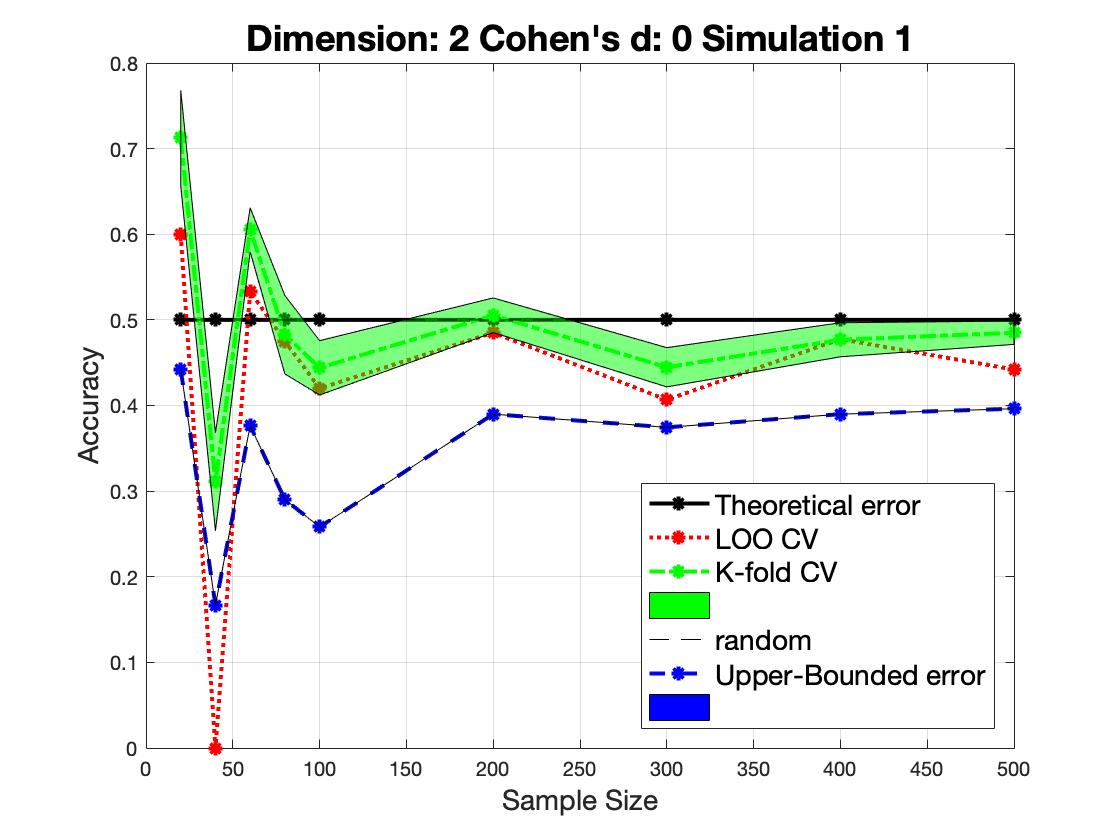}
\includegraphics[width=0.32\textwidth]{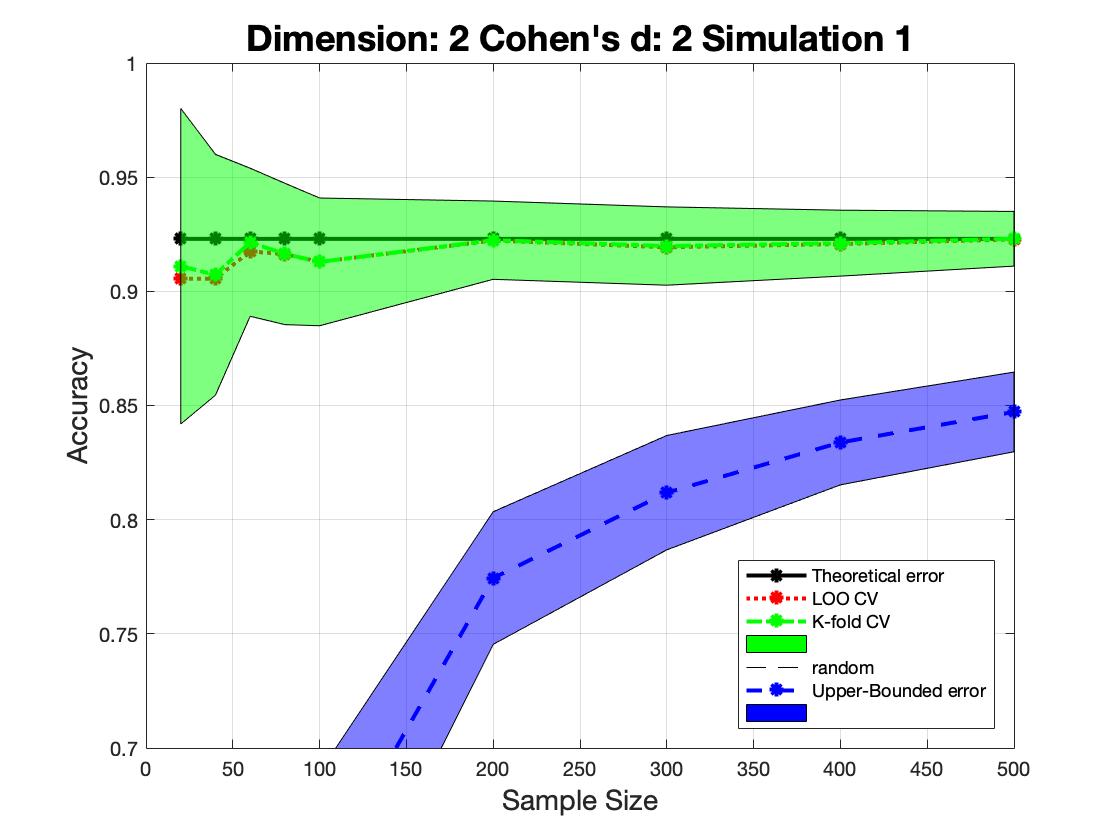}
\includegraphics[width=0.32\textwidth]{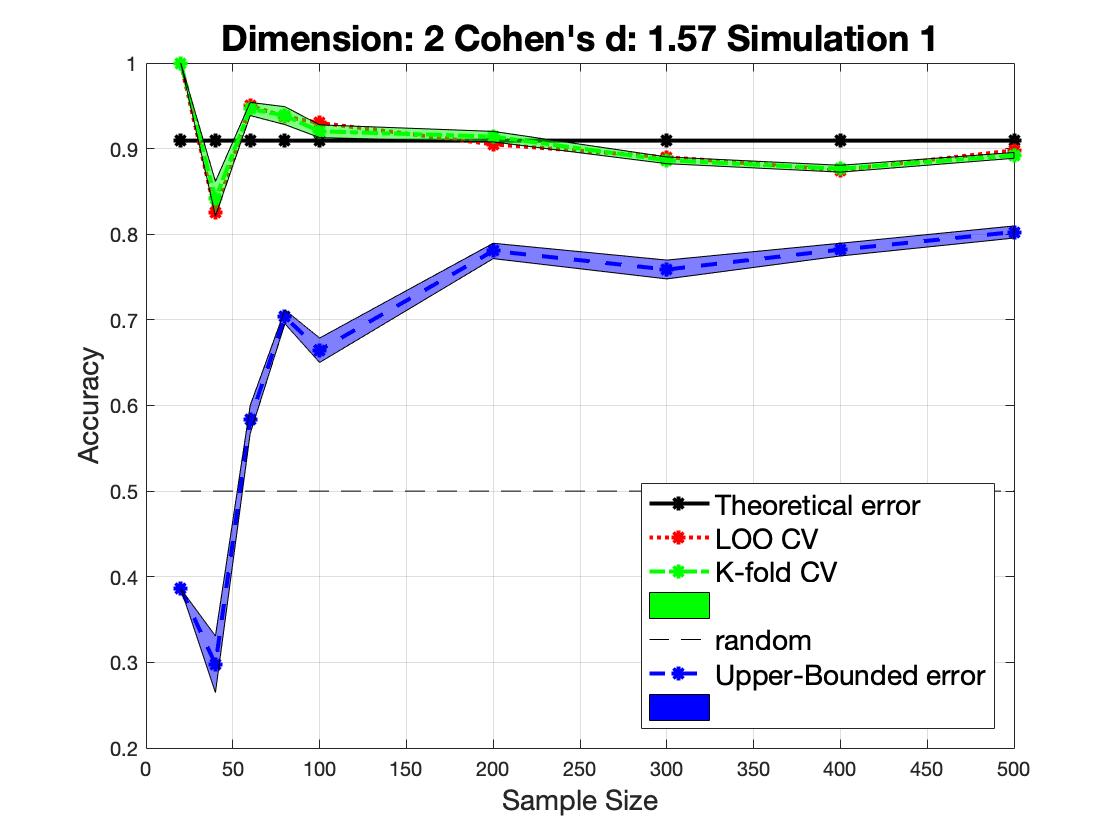}

\caption{Left Column: Null distribution of accuracy values using K-fold CV (in green font) obtained from sampling the pdf (middle) and permuting the fold distribution (bottom). In blue we show the proposed K-fold CUBV method to control FP in this null experiment. Note that in this example dimension $n=2$ and Cohen's $d=0$. Middle column: example of a classification problem using linear decision functions and samples drawn from two Gaussian pdfs with Cohen's $d=2$ similar to the problem described in section \ref{sec:null}. Averaged accuracy and its standard deviation versus sample size are displayed in green font for the standard K-fold CV, $K=10$. The theoretical error achieved by linear classifiers and the whole dataset ($2*10^4$ samples) in this problem is displayed by the black line. Right column: Example of a classification problem with a single sample generated following the procedure described in \cite{Gorriz19} and a $F=100$-fold accuracy distribution for $K=10$.}
\label{fig:example4}
\end{figure*}

Assume that two laboratories A and B acquire experimental data from two populations and seek an effect in a group comparison that is actually trivially small, i.e. dependent on the sample realization (figure \ref{fig:example4}). Both laboratories believe that an effect exists although the patterns belonging to both classes are drawn from the same, unknown, Gaussian distribution: $\mathcal{N}(0,1)$. Given their sample realizations, they apply their ML algorithms for pattern classification using a properly established CV-scheme and obtain accuracy values. The results obtained by the two laboratories are, of course, not identical (a different sample realization in each lab) and \emph{normally} distributed around $0.5$. This expected theoretical result is achieved if we follow the same scheme with an increasing number of laboratories and average the set of accuracy values.  

As an illustration, K-fold CV was evaluated in a classification task consisting of synthetic i.i.d. data drawn from two completely overlapped Gaussian distributions, representing two groups to be compared, in $n=2$ dimensions; that is, with Cohen's $d=0$ between the distribution centroids. We assessed the control of False Positives (FP) in this experiment where any effects detected are type I errors as a function of sample size. As shown in figure \ref{fig:example4}, the experiment on the left shows a symmetric distribution of the statistic when we sample the Gaussian pdf and directly compare the results. On the right of \ref{fig:example4} is the \emph{estimated} null-distribution of classifier error obtained when permuting labels $F=100$ times using a single-point realization of the sample. It is non-symmetrically distributed and biased around $50\%$. In short, the statistical tests performed in a permutation analysis  result in excess FP (or false negatives depending on the sign of the bias) above or below random chance. 

\subsubsection{Classification variability across independent (multi-sample) experiments}


Assume two laboratories A and B are exploring a specific neurological condition versus a control pattern by a classification task. Datasets obtained from different experimental setups and signal processing pipelines may be modelled by a fixed, but unknown pdf. Each laboratory enrols a different cohort of participants and after data acquisition and processing they obtain noisy i.i.d samples (theoretically) drawn from different samples of the same pdf. We may simulate this scenario with synthetic data by randomly generating two samples of sample size $N$ from 2-dimensional normal distributions $\mathcal{N}(0,1)$ and $\mathcal{N}(d,1)$, where $d$ is the Cohen's distance between the centroids. Unfortunately, even following the same protocols, laboratories A and B obtain different performance accuracies averaged across CV folds. 

In general, classification results obtained from K-fold CV reflects the variability in the sample at different sample sizes, $N$, even when the effect is large ($d=2$) as shown in figure \ref{fig:example4}. Note in this figure how the accuracy is symmetrically distributed around the real effect (black line) and the decreasing variability with increasing $N$, as expected. However, with a realization of $20$ samples we readily see that the effect could be overestimated in half of the laboratories that undertake this experiment (values above $95\%$) or underestimated on the other half (values around $85\%$) when the true effect is around $90\%$. It would be desirable to provide an additional test to evaluate if under the specific experimental conditions (sample size, number of predictors, classifier complexity, etc.) laboratories A or B give performances that could be extrapolated elsewhere and are close to the real effect. 

\subsubsection{Classification variability across CV-folds in single sample experiments }


An increasing number of international challenges and initiatives as well as open source databases foster the common situation whereby laboratory A shares data with B to test the same hypothesis \cite{sarica_editorial_2018,NIA18,Gorriz2020}. Although both laboratories have the same single realization of the sample, neither obtained the same classification results or estimated the same effect level in data, \emph{if any}. This is because the selected fold distribution in the CV was different, thus predictive accuracy in the test data and its average may be slightly or substantially dissimilar. This scenario can be modelled in the same manner as in the preceding examples by randomly permuting the training and test folds $F$ times.  

As an illustration, we simulated two data distributions with $d=1.57$ following the procedure shown in \cite{Gorriz19} (figure \ref{fig:example4}). We observed  a ``non-symmetric'' distribution of accuracy values around the real effect level with increasing sample size. The results obtained by many laboratories are similar (less variability than the previous example), but \emph{biased} mainly due to small sample sizes and the single realization of the sample.

Finally, data may be non-Gaussian distributed, e.g. see an example in $n=2$ dimensions in figure \ref{fig:example4}, generating imbalanced modes. In the previous examples, we assumed that samples were drawn from one dominant mode and an almost symmetric Gaussian pdf.  Complex data following imbalanced multi-modal pdfs \cite{Gorriz19} further increases the variability of the performance obtained at each lab. Moreover, the size of effect strongly influences the variability of the accuracy results, with small samples increasing variability, as shown in the following sections; see figure \ref{fig:complexity}.

\section{The K-fold Cross Upper Bound Validation test}\label{sec:kcubv}

\subsection{A statistical test based on K-fold CV and upper bounding}
\label{sec:novelstat}
The proposed statistical test in this section, the K-fold CUBV test, formalizes the selection of the accuracy threshold (typically $50\%$) as the criterion for detecting an effect in between-group analysis. The rationale behind this test is that when a classifier with fixed complexity and a fixed number of predictors (dimensions of the input pattern) is fitted to a small sample-size dataset, statistical significance cannot be guaranteed if the effect size is small. This limitation exists despite using empirical error and model-driven confidence intervals. 

Instead, we propose to reject the null-hypothesis if the analysis of the \emph{worst case} with at least a probability $1-\eta$ provides an upper bound that satisfies:
\begin{equation}\label{eq:test}
\mathcal{R}(f_N)=\mathcal{R}_N(f_N)+\Delta(N,\mathcal{F},Q)\leq \eta
\end{equation} 
where we use CV to estimate the empirical risk $\mathcal{R}_N(f_N)$ and $\Delta(N,\mathcal{F},Q)$ is an (agnostic) upper bound based on SLT analysed in the following sections. Note that this probability $\eta$ refers to the analysis of the worst case and is not related to the level of significance $\alpha$ of a classical statistical test. With an $\alpha < 0.05$ only $1$ out of $20$ random effects would be considered as a real effect by rejecting the null hypothesis. In the worst case analysis we could soften this condition to $\eta=0.5$ as it refers to the supremum of the deviation between actual and empirical errors \cite{Gorriz2021}, a value that an efficient SLT algorithm, such as a linear Support Vector Machine (SVM) \cite{Burges98}, could not achieve given the sample set.

\begin{figure*}
\centering
\includegraphics[width=\textwidth]{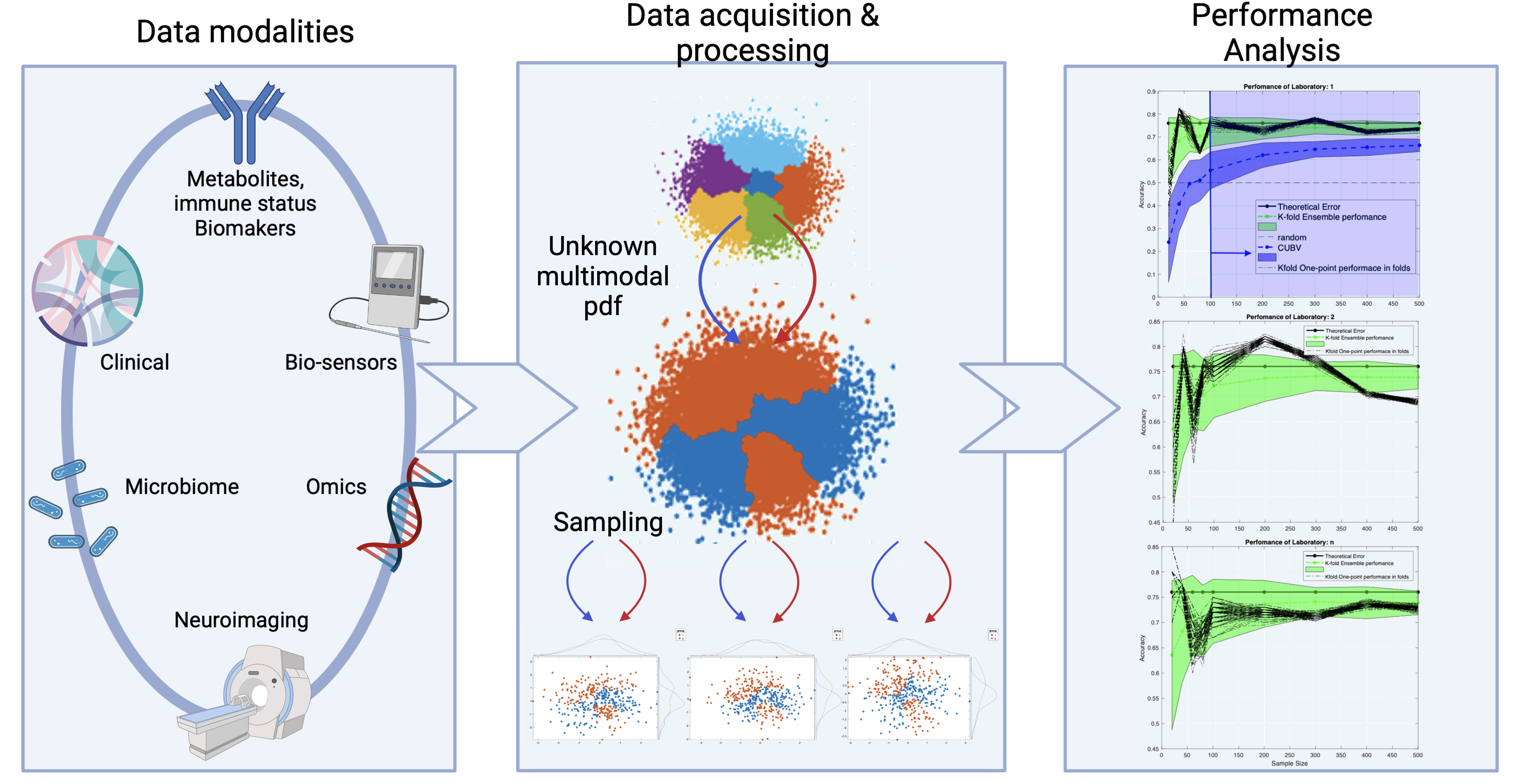}
\caption{Performance of K-fold CV in common experimental designs. Typical large biobanks include data across modalities including neuroimaging, biosensors, genetic, clinical, omics, etc. data. A set of synthetic and real MRI samples, obtained from $N_c$ sources and expressed as $n$ dimensional features, are analysed. The theoretical error achieved by a (linear) classifier can be assessed by resubstitution on the infinite population (pdf). Then, the K-fold CV error is estimated by (theoretically) sampling this pdf ($M$ times) or by (realistically) permuting the learning folds using a single realization of the sample ($F$ times). The proposed CUBV test rejects the null-hypothesis in the blue-shaded area.} 
\label{fig:summary}
\end{figure*}

In figure \ref{fig:summary} we collect all the ideas presented in previous illustrations of different experimental scenarios. We show the operation of the proposed CUBV test for detecting effects with good performance using the regular K-fold CV and a single-point realization of the sample. Here, performance is described in terms of variability of the error and its deviation from the actual risk. As shown in figure \ref{fig:summary}, theoretically sampling the unknown pdf (in green font) provides a biased estimation of the theoretical error. Moreover, using a single realization (dash-dotted lines) does not improve the situation, but quite the opposite especially at small sample sizes. Finally, when the deviation obtained by the CUBV technique is in the majority above $\eta=0.5$ (blue-shaded area on the right) one can be assured that performance of a single-point realization based K-fold CV in ``laboratory 1'' can be ``extrapolated'' to other laboratories. Note that the size of effect influences this decision (real classification rates of around $75\%$). 

\subsection{Confidence intervals with SLT: The Chernoff bound}
Probabilistic methods in combinatorics have provided useful tools and results for deriving powerful and robust bounds for learning algorithms based on bounded loss functions. The Chernoff bounding trick \cite{Chernoff52} is one of them, allowing us to bound the probability $\mathbb{P}[X\geq \mathbb{E}X+t]$ of a random variable $X$ for some $t>0$. 

\begin{definition}
    Given a random sample $Z^N$ according to some $P$ and a class $\mathcal{F}$ of functions, we define the uniform deviation as:
\begin{equation}
    \Delta(N,\mathcal{F})\equiv\sup_{f\in\mathcal{F}}|\mathcal{R_N}(f)-\mathcal{R}(f)|
\end{equation}
\end{definition} 
\noindent where the empirical risk $\mathcal{R_N}(f)$ is estimated by averaging the loss on the sample of size N (similar to eq. \ref{eq:CVloss}). It is easy to see that for any $f\in\mathcal{F}$:
\begin{equation}\label{eq:boundrisk}
   \mathcal{R}(f)\leq \mathcal{R_N}(f)+ \Delta(N,\mathcal{F})
\end{equation}
and, in particular, for the fitted function obtained from data by ERM. The Chernoff bound applied to the random variable $\Delta(N,\mathcal{F})$, which is bounded by $1/N$; is given by the McDiarmid's inequality \cite{McDiarmid89}:
\begin{equation}
   \mathbb{P}\left(\Delta(N,\mathcal{F})\geq \mathbb{E}[\Delta(N,\mathcal{F})]+t\right)\leq e^{-2Nt^2}
\end{equation}
for any $t>0$. Letting $t=\sqrt{\frac{log(1/\eta)}{2N}}$, we conclude that:
\begin{equation}\label{eq:boundDelta}
   \mathbb{P}\left(\Delta(N,\mathcal{F})\leq \mathbb{E}[\Delta(N,\mathcal{F})]+\sqrt{\frac{log(1/\eta)}{2N}}\right)\geq 1-\eta
\end{equation}

\subsection{Upper Bounding the actual risk under the worst-case scenario}
\label{sec:GLMLast}

The main goal of SLT is to provide a framework for addressing the problem of statistical inference \cite{Vapnik82,Haussler92}. One of its most notable achievements is to establish simple and powerful confidence intervals for bounding the actual risk of misclassification, $\mathcal{R}(f)$ \cite{Boucheron13}. In particular, we are interested in bounding the actual risk of $f_N$ from the empirical quantity obtained on the sample $Z^N$ with probability at least $1-\eta$. Combining inequations \ref{eq:boundrisk} and \ref{eq:boundDelta}, we see that:
\begin{equation}\label{eq:dos}
\mathcal{R}(f_N)\leq\mathcal{R}_N(f_N)+\mathbb{E}[\Delta(N,\mathcal{F})]+\sqrt{\frac{log(1/\eta)}{2N}}
\end{equation} 
with probability at least $1-\eta$. $f_N$ is estimated to prevent overfitting \footnote{A classifier that perfectly predicts the labels of the training data but often fails to predict them on the test set.}, e.g. restricting the class of functions, and $\mathbb{E}[\Delta(N,\mathcal{F})]$ is an upper bound of the actual risk. In the \emph{worst case} analysed in section \ref{sec:novelstat}, the inequality turns into an equality. This deviation or inequality can be interpreted from several perspectives of classical probability theory \cite{Vapnik82} in order to assess how close the sum of independent random variables (empirical risk) are to their expectations (actual risk). 

\subsection{A Probably Approximately Correct Bayesian bound}

The previous statistical approach for upper bounding, similar to the classical statistical approaches found in the literature \cite{Bates2023, Rodriguez2010}, focuses on functional estimands and the relationship between empirical error and its expected value on new unseen patterns. The PAC-Bayesian approach takes a broader perspective by allowing the estimator $f$ to be drawn randomly from an auxiliary source of randomness, which smooths the dependence of $f$ on the sample \cite{Cantoni07}. Here, we employ one of the major advances in this field based on PAC-Bayesian theory \cite{MacAllester2013}. In particular, we evaluate a dropout bound inspired by the recent success of dropout training in deep neural networks. The bound depicted in equation \ref{eq:dos} is expressed in terms of the underlying distribution $Q$ that draws the function $f$ from the set of 'rules', $\mathcal{F}$.

\begin{definition}
    Given a random sample $Z^N$ and some $\gamma\in \mathbb{R}$, we define the Bayesian deviation as:
\begin{equation}\label{eq:Bdev}
    \Delta(N,\mathcal{F},Q)\equiv|\gamma \mathcal{R}_N(f)-\ln\left(1-\mathcal{R}(f)+\mathcal{R}(f)e^{\gamma}\right)|
\end{equation}
\end{definition}
Note that we omit the dependence of the risks on Q, where both are random variables $\in [0,1]$.

\begin{theorem}
For any constant $\lambda>1/2$, and a class of classifiers $\mathcal{F}$ that are selected according to the distribution $Q$, we have that with probability at least $1-\eta$ over the draw of the sample, the following CI hold for all the distributions $Q$:
\begin{equation}
\begin{array}{l}
\mathcal{R}(f)\leq \mathcal{R}_N(f) \\+ \min_\lambda\frac{1}{2\lambda-1}\left(\mathcal{R}_N(f)+\frac{2\lambda^2}{N}\left(D_{KL}(Q||Q_u)+\ln\frac{1}{\eta}\right)\right)
\end{array}
\end{equation}
where $D_{KL}(Q||Q_u)\equiv\mathbb{E}_{f\sim Q}[\ln \frac{Q(f)}{Q_u(f)}]$ is the Kullback-Leibler divergence from $Q$ to the uniform distribution $Q_u$. 

Proof:  Using the Jensen inequality for any convex function, i.e. $f(\mathbb{E}[x])\leq\mathbb{E}[f(x)]$, the Bayesian deviation, scaled by $N$, can be bounded by:
\begin{equation}\label{eq:jensen}
    N\Delta(N,\mathcal{F},Q)\leq \mathbb{E}_{f\sim Q}[N\Delta(N,\mathcal{F},Q)] 
\end{equation}
Combining equations \ref{eq:Bdev} and \ref{eq:jensen}, we get:
\begin{equation}\label{eq:boundB}
\begin{array}{l}
    \mathcal{R}(f)\leq\frac{1-e^{\gamma\mathcal{R}_N(f)-\frac{1}{N}\mathbb{E}_{f\sim Q}[N\Delta(N,\mathcal{F},Q)] }}{1-e^\gamma}\\
    \text{Subject to: } \mathcal{R}(f)\in[0,1]
        \end{array}
\end{equation}
Taking into account the Taylor's series of the term $e^\gamma$, if $\gamma<0$, we can additionally bound eq. \ref{eq:boundB} as:
\begin{equation}\label{eq:boundB2}
    \mathcal{R}(f)\leq\frac{-\sum_{j=1}^\infty\frac{g(\gamma)^j}{j!}}{-\sum_{l=1}^\infty\frac{\gamma^l}{l!}}\leq\frac{-\sum_{j=1}^{j_0}\frac{g(\gamma)^j}{j!}}{-\sum_{l=1}^{l_0}\frac{\gamma^l}{l!}}
\end{equation}
where $g(\gamma)=-\gamma \mathcal{R}_N(f)+\frac{1}{N}\mathbb{E}_{f\sim Q}[N\Delta(N,\mathcal{F},Q)]$, $j_0$ is odd and $l_0$ is even. We continue the proof selecting $j_0=1$ and any even $l_0$, then:
\begin{equation}\label{eq:boundB2}
       \mathcal{R}(f)\leq\frac{\mathcal{R}_N(f)-\frac{1}{N\gamma}\mathbb{E}_{f\sim Q}[N\Delta(N,\mathcal{F},Q)]}{1+\sum_{l=2}^{l_0}\frac{\gamma^{l-1}}{l!}}\
\end{equation}
where $\gamma<0$. We then apply the Donsker's variational formula \cite{Donsker83} to bound the expected term in eq. \ref{eq:boundB2} as:
\begin{equation}\label{eq:boundB3}
    \mathbb{E}_{f\sim Q}[h]\leq D_{KL}(Q||Q_u)+\ln\left( \mathbb{E}_{f\sim Q_u}[e^{h}]\right)
\end{equation}
where $h=N\Delta(N,\mathcal{F},Q)$. The resulting inequality:
\begin{equation}\label{eq:boundB2bis}
       \mathcal{R}(f)\leq\frac{\mathcal{R}_N(f)-\frac{1}{N\gamma}\left(D_{KL}(Q||Q_u)+\ln\left( \mathbb{E}_{f\sim Q_u}[e^{h}]\right)\right)}{1+\sum_{l=2}^{l_0}\frac{\gamma^{l-1}}{l!}}\
\end{equation}
 holds with probability one for any $\gamma<0$ subject to $\mathcal{R}(f)\in[0,1]$. 

From the definition of \(\Delta(N, \mathcal{F}, Q)\), it is easy to see that \(\mathbb{E}_{f \sim Q_u}[e^{h}] < 1\). Using Markov's inequality for a non-negative random variable, i.e., for any \(t > 0\), \(\mathbb{P}(X \geq t) \leq \frac{\mathbb{E}[X]}{t}\), and letting \(t = \frac{1}{\eta}\), we get:
\begin{equation}\label{eq:boundB4}
    \mathbb{P}\left(\mathbb{E}_{f\sim Q_u}[e^{h}]>1/\eta\right)<\mathbb{E}_{Z^N\sim P}\left[\mathbb{E}_{f\sim Q_u}[e^{h}]\right] \eta\leq \eta
\end{equation}
Using eq. \ref{eq:boundB4} in eq. \ref{eq:boundB2bis} we have that with at least a probability $1-\eta$:
\begin{equation}\label{eq:boundB5}
    \mathcal{R}(f)\leq\frac{\mathcal{R}_N(f)-\frac{1}{N\gamma}\left(D_{KL}(Q||Q_u)+\ln\frac{1}{\eta}\right)}{1+\sum_{l=2}^{l_0}\frac{\gamma^{l-1}}{l!}}
\end{equation}
The latter CI hold simultaneously for all $\gamma<0$ and $Q$. Then, Theorem 1 now follows by setting $\gamma=-\frac{1}{\lambda}>-2$, $l_0=2$ and selecting the minimum bound by minimizing with respect to $\lambda$. 
\end{theorem}
Moreover, technical details on the minimisation of Bayesian bounds can be found in \cite{MacAllester2013}. For example $Q_u$ can be formalised as $\mathcal{N}(0,1)^n$ and $D_{KL}(Q||Q_u)=\frac{1-\delta}{2}||\boldsymbol{\upomega}||^2$ for linear classifiers $f(\mathbf{x})=\boldsymbol{\upomega}\cdot \mathbf{x}$, where $\boldsymbol{\upomega} \in \mathbb{R}^n$ and $\delta$ is a dropout rate. In this sense, we follow the recommendation found in \cite{Cantoni07} on the implementation of simple Bayesian bounds, which provide a good trade-off between performance and complexity.

\section{Materials and Methods}

\subsection{Statistical Inference with permutation analyses}\label{sec:statinfer}

Statistical inference using ML systems is usually performed by means of predictive inference. Datasets are split into CV folds (training and test sets) and the ability of statistical classifiers to extrapolate from the samples in the training sets is assessed on the test set.  Performance is measured in terms of the classifier error that is then used as a statistic to test for between-group differences; e.g. by estimating p-values from permutation testing. The degree of reliability of this approach will depend on the sample size and the data complexity with both affecting the quality of the learning procedure. A major drawback in using a predictive inference approach is that the CV is sub-optimal \footnote{Proof: the Neyman-Pearson lemma.} when performing a hypothesis test for a given \emph{sample size} and threshold \cite{Friston13}. Indeed, the variance of the error strongly depends on $K$ if the classifier $f$ is unstable under perturbations. This happens when deleting instances of the training folds provokes a substantial change in the outcome, mainly due to data heterogeneity.

A p-value estimation by means of a permutation test \cite{Bullmore99} can be conducted on the classification results from K-fold CV  \cite{Reiss15}. In this way, the error of the trained ML system is the statistic used to test for between-group differences. We evaluate how likely is the performance of the system using the labeled data and empirical distribution compared to the distribution from a number, $M$, of label permutations that simulate ``no effect', i.e. samples are i.i.d (the null distribution) \cite{Jimenez22}. In other words, we assess the probability of the observed value of the error under the null hypothesis by permutation $\pi$, as the following:
\begin{equation}\label{ec:pvalue}
p_{value}=\frac{\#(\mathcal{R^\pi}\leq\mathcal{R})}{M+1}
\end{equation}
where $\#(.)$ is the number of times the error in the permutation is less or equal to the error obtained with the observed effect.  If this value is less than our level of significance, e.g. $\alpha=0.05$, then the distribution of data significantly differs between groups \cite{Phipson2010}.

To sum up, the state-of-the-art in this field often employs only a single-instance estimation from a particular set of CV folds and, at most, a subsequent permutation analysis to simulate the null distribution and test for statistical significance. However, the whole procedure, i.e. the assessment of the true effect and the null distribution from incomplete datasets, is substantially \emph{biased} if the samples are small and heterogeneous. 

\subsection{Monte Carlo Performance Evaluation and Power Calculations}

Monte Carlo (MC) simulation was employed to determine the number of trials of a random variable, or statistic $T,$  required to achieve detection with a given a threshold \cite{Kay98}, e.g. $\mathbb{P}(T>\gamma)>0.5$.  The probability that a $T$-statistic, consisting of an average of normally distributed random variables, is greater than a threshold is equal to $\mathbb{P}(T>\gamma)=\mathbb{Q}\left(\frac{\gamma}{\sqrt{\sigma^2/N})}\right)$, where $\mathbb{Q}$ denotes here the right-tail probability and $\sigma$ is the standard deviation of the random variable. Then,  by estimating this probability using a number of $M$ trials we can evaluate the minimum value to achieve detection at a given significance level $\alpha$ as:
\begin{equation}\label{eq:montecarlo}
M\geq\frac{[\mathbb{Q}^{-1}(\alpha/2)]^2(1-p)}{\epsilon^2p}
\end{equation} 
where $\epsilon$ is the maximum deviation allowed between $p=\mathbb{P}(T>\gamma)$ and its estimation $\hat{P}$ is computed by counting the number of times $T$ is greater than the selected threshold $\gamma$ in $M$ trials. MC simulations were performed in controlled experiments to undertake power calculations.  We estimated the power of the test ($1-\beta$) across permutations as the number of times the p-value (equation \ref{ec:pvalue}) was less than the significance level, 0.05, divided by the total number of permutations.  This was done for different sample sizes under ideal conditions, i.e. mono-modal Gaussian pdf per class, and with imbalanced and multimodal Gaussian data (see the experimental section \ref{sec:experiments}).

\subsection{Synthetic data to model heterogeneous and multi-cluster datasets}
\label{sec:complex}

Current biomedical data provide multi-modal/dimensional sources of information which result in complex and heterogeneous datasets \cite{Zhang2020,Acosta22} (see figure \ref{fig:summary}). Real data in between-group analyses are usually characterized by a multimodal pdf that draws samples from several sources of variability (covariates) per group such as sex, age, socioecomic status, genetic profiles, and so on \cite{Hyatt20} that all influence the effect under observation \cite{Gorriz18,Leming20}. Assuming data from each group is defined by a balanced set of Gaussian clusters, each cluster representing a data source, we generate realistic data following the procedure described in \cite{Gorriz19}, or simply by generating samples given each Gaussian cluster in a $n$-dimensional space. 

SLT provides additional statistical measures to control: i) the degree of reality of the pattern generated by these procedures (typically used in the extant literature to evaluate models in a controlled, experimental fashion), or simply; ii) the ability of the selected classifier complexity to perform well in a classification task (see supplementary materials).

\subsubsection{Simulation of single cluster Gaussian PDFs}

We generated synthetic data to model different scenarios in neuroimaging data analysis whereby observation vectors $\mathbf{x}$ were simulated for each group from Gaussian distributions (similar to figure \ref{fig:example4}) with different effect sizes ($d:0-2$), sample sizes ($N:20 - 500$), and dimensions ($n:1-6$). To complement the results presented in the preliminary example in section \ref{sec:preliminary}, we computed the power of the k-fold CV-based permutation test and the upper-bounding test (considered as an independent test) in these commonly encountered experiments.  We drew samples from $n=2$ Gaussian pdfs with only two clusters (one per group) and increasing $d$. Note that the aim of this paper is the combination of both; that is, using the CUBV technique to validate the K-fold CV test within a permutation analysis. 

\subsubsection{Simulation of complex multi-cluster Gaussian PDFs}

We repeated the simulations increasing the number of clusters, $N_c$, per group (see figure \ref{fig:example3}) and constructing pdfs for each group from an imbalanced number of samples drawn from each cluster. This procedure increases the complexity of the simulation affecting the generalisation of performance of K-Fold CV. The idea behind this simulation is quite simple: by chance the fitted classifiers at the training stage cannot perform well on the test set because patterns in both groups (training and test) are drawn from different pdf modes at different ratios.

\subsection{Neuroimaging dataset}\label{sec:datareal}

Datasets provided by the International challenge for automated prediction of \emph{mild cognitive impairment} (MCI) from MRI data (https://inclass.kaggle.com/c/mci-prediction) were considered for the evaluation of the proposed method in a neuroimaging context. MRI scans were selected from the Alzheimer's disease Neuroimaging Initiative (ADNI, http://www.adni-info.org) and preprocessed by Freesurfer (v5.3) \cite{sarica_editorial_2018}. The dataset consisted of 429 demographic, clinical, and cortical and subcortical MRI features for each participant. Participants were in four groups according to their diagnostic status: healthy control (HC), AD patients, MCI individuals whose diagnosis did not change in the follow-up, and converter MCI (cMCI) individuals that progressed from MCI to AD in the follow-up period. The training dataset contained $400$ ADNI individuals ($100$ HC, $100$ MCI, $100$ cMCI and $100$ AD). Up to $20$-dimensional partial least squares (PLS) based features were extracted from the MRI-study groups according to \cite{Gorriz19} and then combined to provide different classification problems with heterogeneous sources ($N_c=4$) by combining the resulting groups. 

\section{Results}\label{sec:experiments}

\subsection{Experimental Designs}

We performed an analysis of the scheme shown in Figure \ref{fig:summary} under the experimental conditions described in the illustrated scenarios (Section \ref{sec:preliminary}). For each experimental design, we used a linear SVM and three validation methods were compared: leave-one-out (LOO) CV, regular K-fold CV, and K-fold CUBV. We used $K = 10$-fold CV, $M = 100$ realizations (ensemble experiments), and $F = 100$ repetitions (single instance experiments). Note that the latter ``bootstrapping strategy'' based on repetitions, employed for all the CV procedures, substantially improves the estimation of the prediction error. Moreover, the impact of nested CV is relatively low with small datasets (due to re-splitting) and using linear classifiers (no optimization on the regularization parameter $C$ was performed in the experiments). Nonetheless, we added the theoretical confidence intervals of the Nested CV procedure in selected experiments to compare them with the empirical intervals, as described in \cite{Bates2023}, a theoretical method contemporary to our present work (see appendix \ref{sec:nestedapprndix}). The experimental designs were chosen to: i) evaluate the power of the tests; ii) assess the theoretical (ensemble) and real (single realization with fold resampling) performance of the regular K-fold CV test; and iii) carry out experiments on an MRI multiclass dataset \cite{sarica_editorial_2018}.

Power calculations for the null experiment, where $d=0$ (\ref{sec:null}), were undertaken to understand nominal FP control. These values were included within power calculations at different effect sizes $d=\{0\ldots 4\}$ in $n=2$ dimensions, data complexity $N_c=\{2,4\}$, samples sizes $N=\{20\ldots 500\}$, as shown in the experimental part. To assess the probability of detection we evaluated the MC performance of the regular K-fold CV (number of trials required to detect the effect) and the detection ability of the test based on CUBV.

All the experiments were carried out in the ``ideal case'' based on sampling the theoretical pdf $M=\{100,1000\}$ times or label permuting $F=\{100,1000\}$ times a single realization of the sample (real case). In addition, in both cases we generated data from single mode pdf and multimode pdf, thus increasing the complexity of the problem. Finally, real MRI data was analysed using this methodology to validate the findings achieved on a statistical significant level.

\subsection{Results on the null experiment}

Results for the null experiment, $d=0$, are displayed in figures \ref{fig:powertest} and \ref{fig:powertest2}. Power of the K-fold CV-based permutation test was above the significance level confirming a FP rate in some experimental setups beyond an admissible level, figures \ref{fig:powertestsingle} and \ref{fig:powertestmulti}. This effect is partly controlled by increasing the sample size with a low data complexity. The K-fold CUBV, used as a statistical test, always provided power below the significance level and thus can be considered conservative within ML-based inference approaches.

\subsubsection{\textbf{Control of type I errors across independent (multi-sample) experiments}}

\begin{figure*}
\centering
\begin{tikzpicture}
\matrix (a)[row sep=0mm, column sep=0mm, inner sep=1mm,  matrix of nodes] at (0,0) {
            \includegraphics[width=0.49\textwidth]{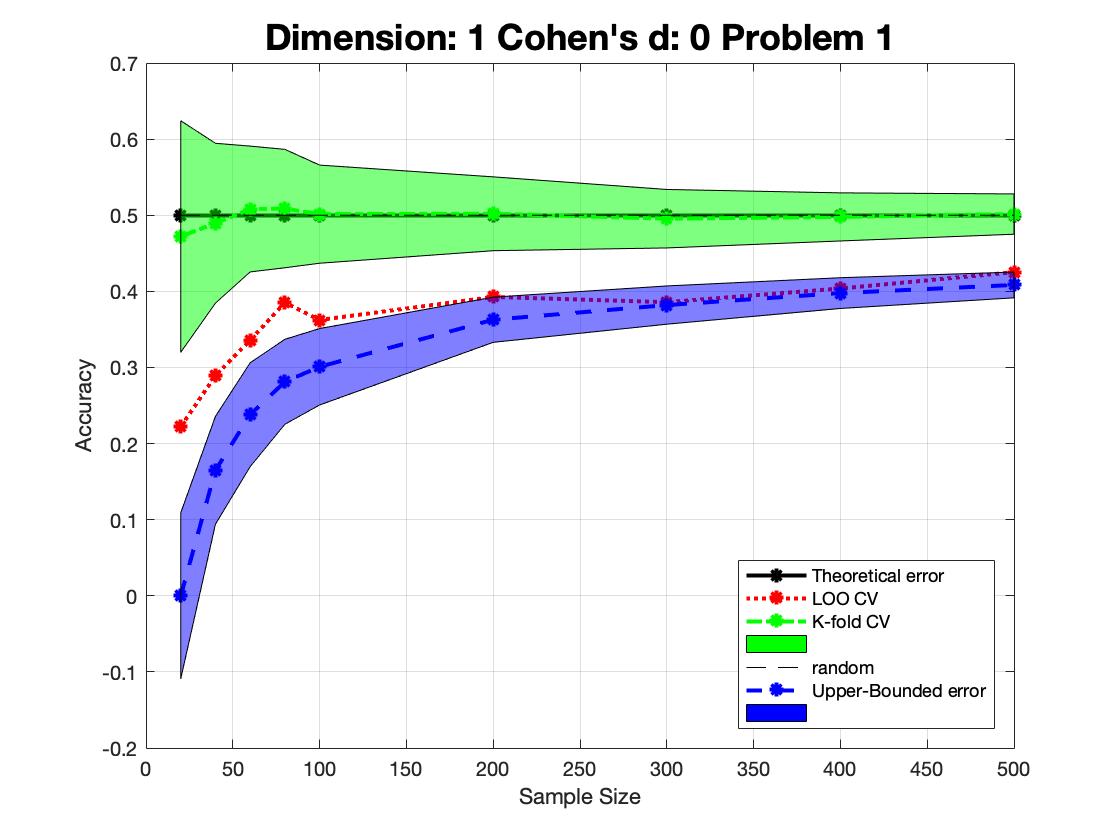} &
            \includegraphics[width=0.49\textwidth]{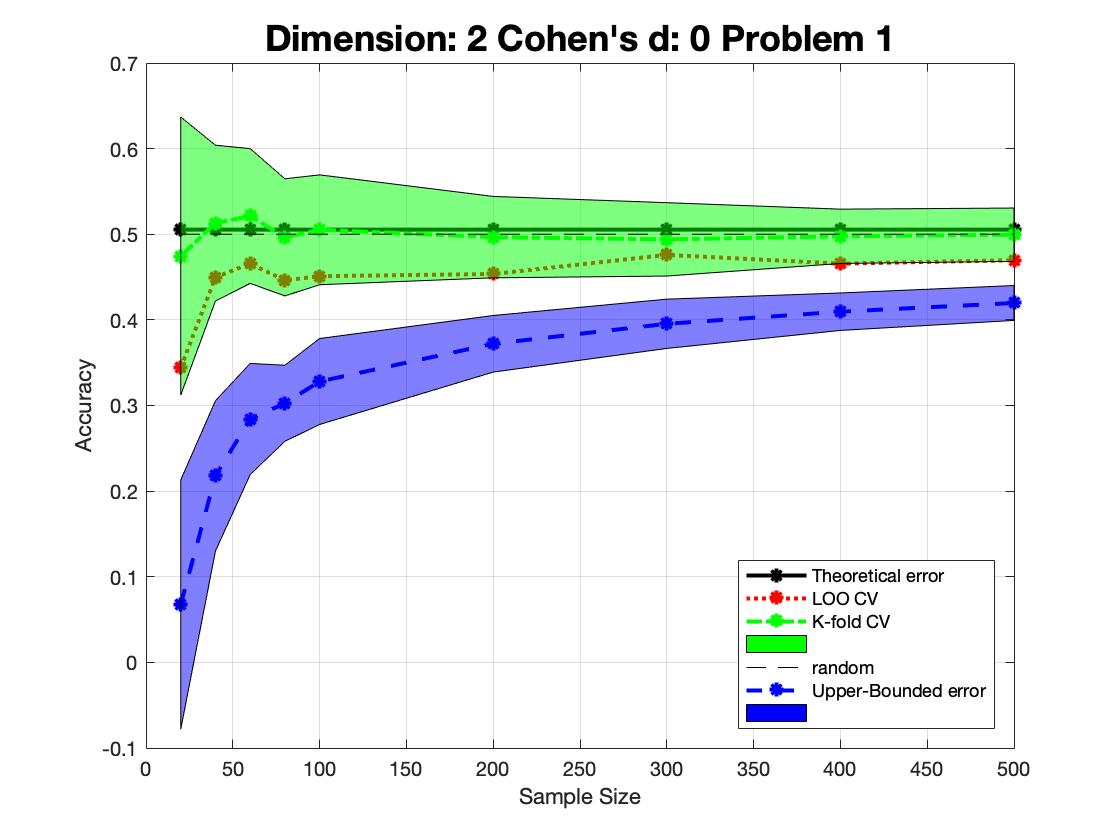} \\
            \includegraphics[width=0.49\textwidth]{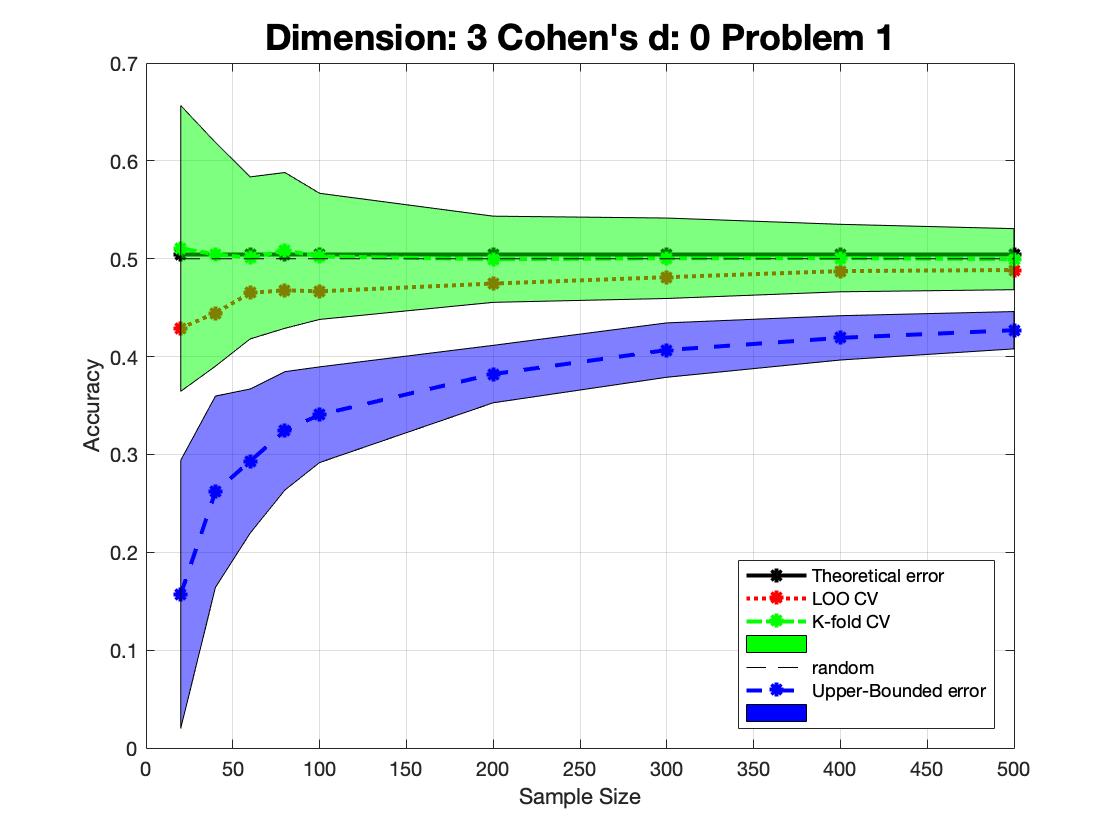} &
            \includegraphics[width=0.49\textwidth]{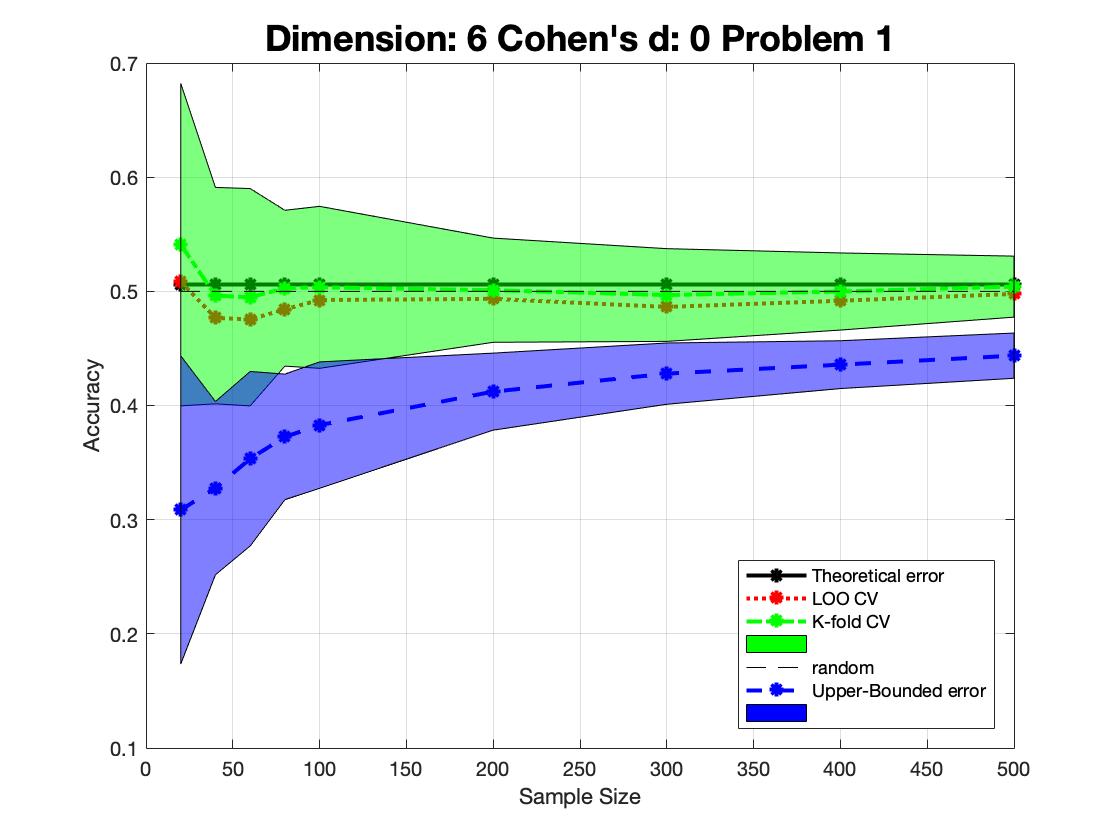} \\
            \includegraphics[width=0.49\textwidth]{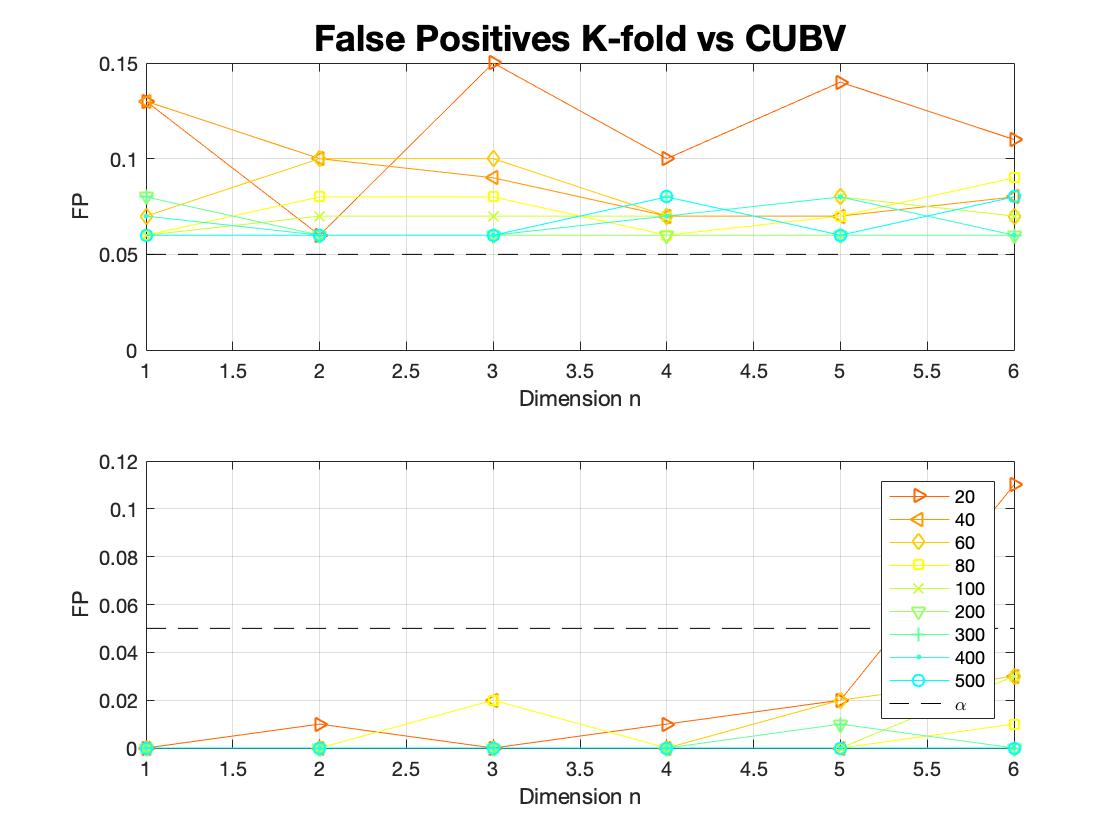} &
            \includegraphics[width=0.49\textwidth]{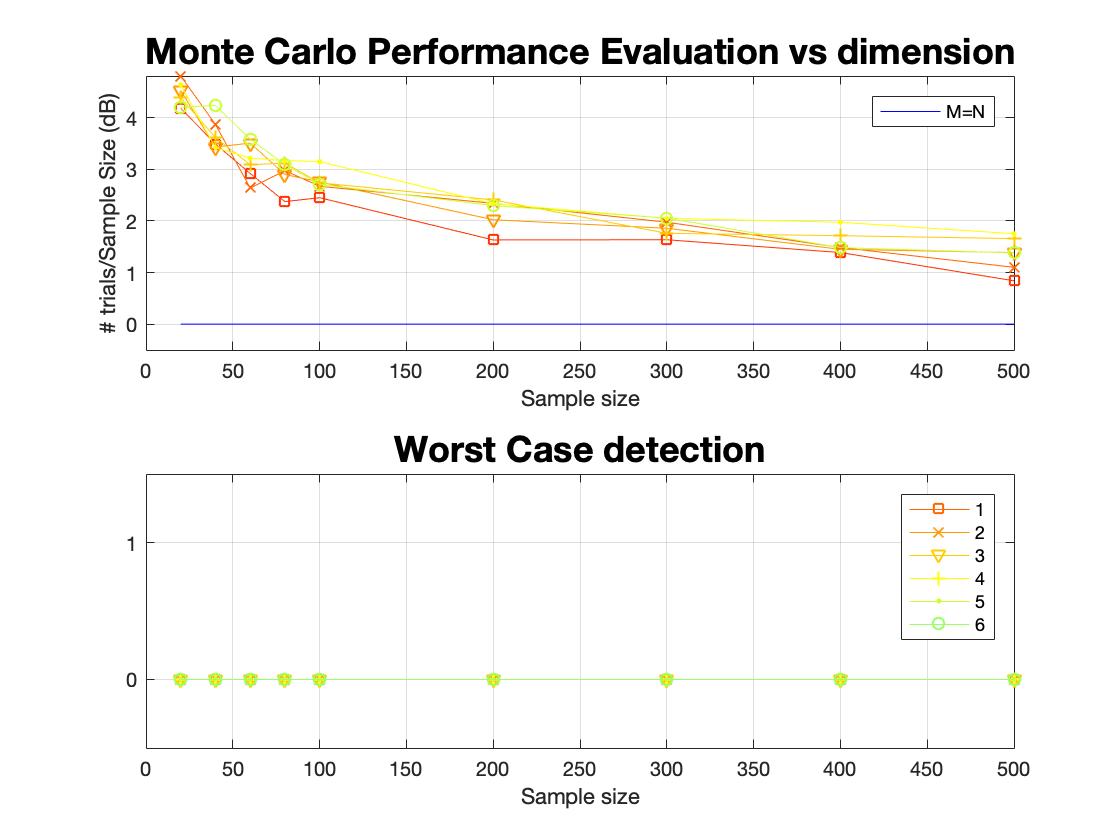}\\\\
        };
\draw[thick,blue!20] (a-1-1.north east) -- (a-3-1.south east);
\draw[thick,densely dashed,blue!20] (a-2-1.north west) -- (a-2-2.north east);
\draw[thick,densely dashed,blue!20] (a-3-1.north west) -- (a-3-2.north east);
\end{tikzpicture}
\caption{Examples of performance, FP rates and MC performance evaluation across independent (multi-sample) experiments}
\label{fig:powernull}
\end{figure*}

In figure \ref{fig:powernull} we display the  analysis of the FP rate under the null-hypothesis with increasing dimension $n$ and  sample size $N$. Performance is plotted for all the validation methods with $M=100$ samples obtained from theoretical Gaussian pdfs and $N_c=2$, the MC performance evaluation and the worst case detection analysis. CUBV controlled type I errors in all the analysed simulations. Note that both detection methods (MC and CUBV) yields similar results of no effect, but the FP rate in K-fold CV is always above the nominal value.

\subsubsection{\textbf{Control of type I errors in single sample experiments}}

\begin{figure*}
\centering
\begin{tikzpicture}
\matrix (a)[row sep=0mm, column sep=0mm, inner sep=1mm,  matrix of nodes] at (0,0) {
            \includegraphics[width=0.49\textwidth]{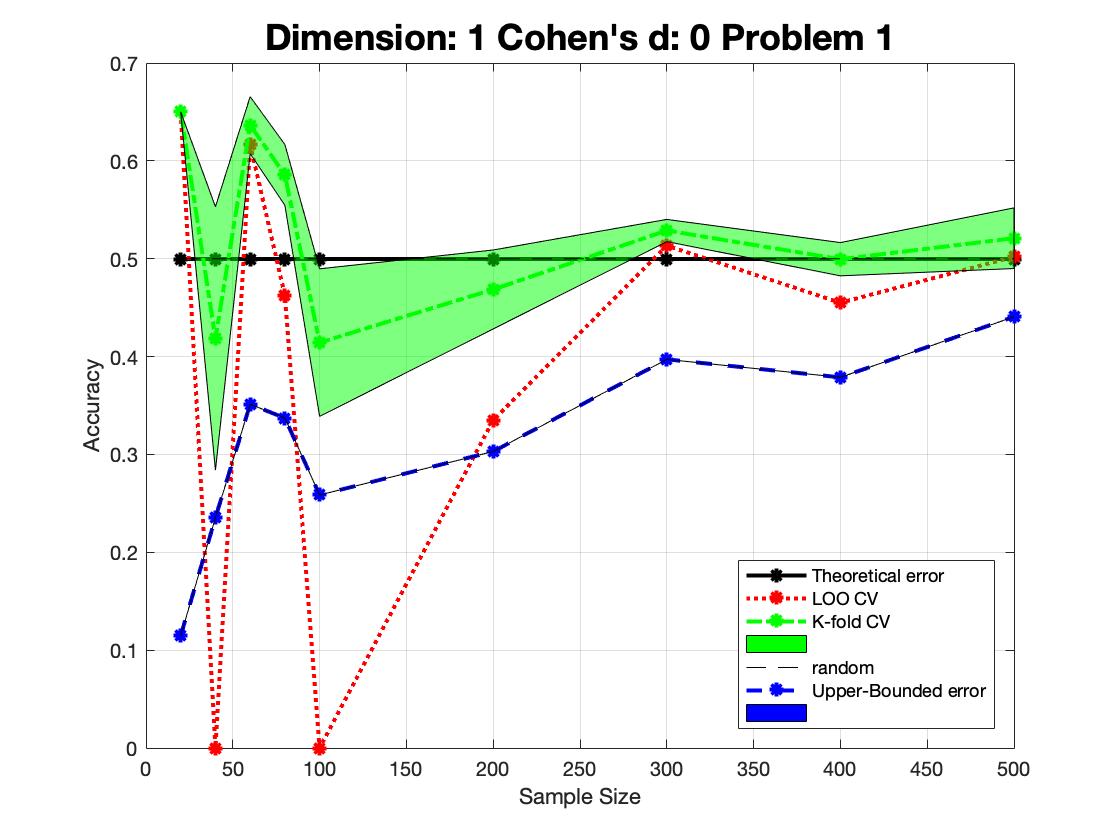} &
            \includegraphics[width=0.49\textwidth]{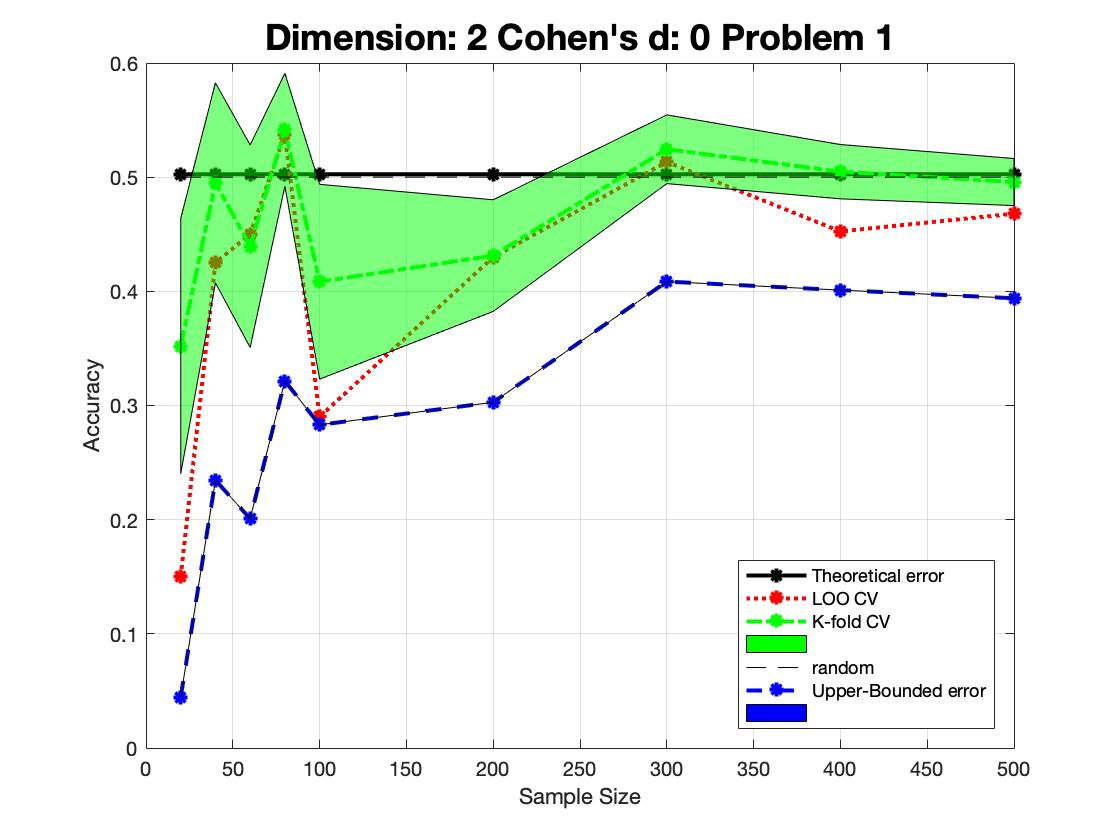} \\
            \includegraphics[width=0.49\textwidth]{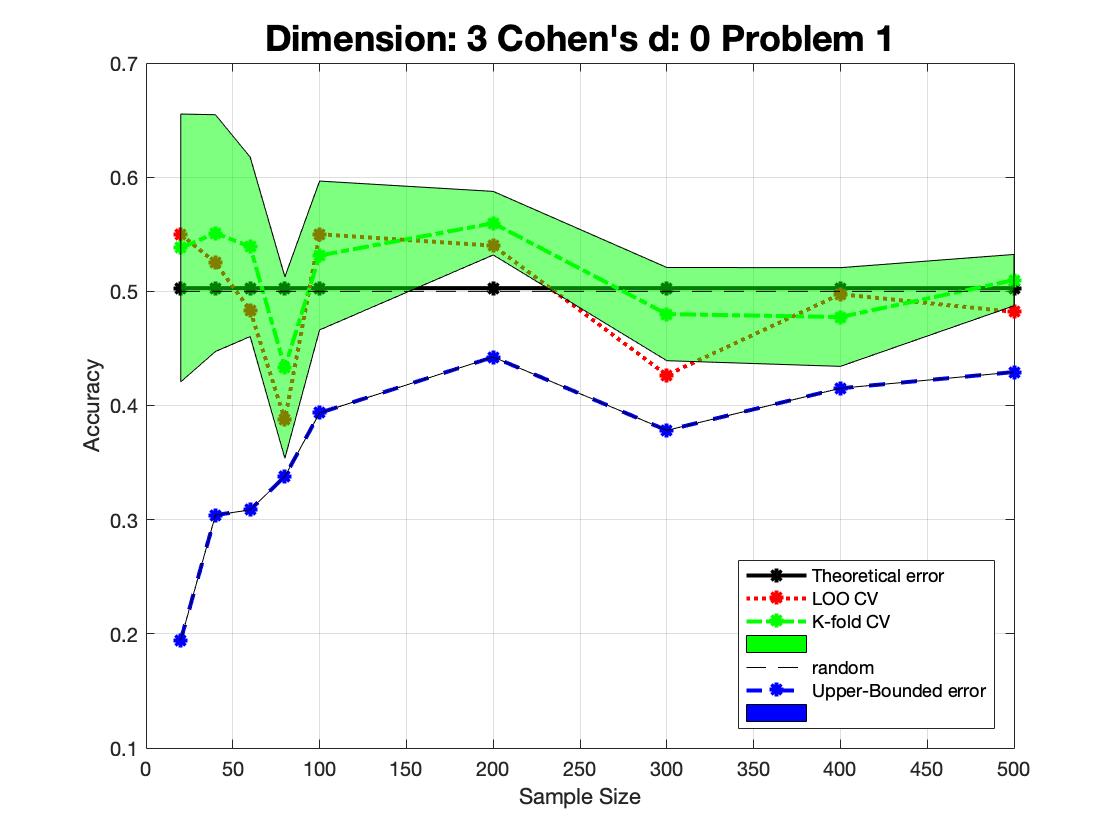} &
            \includegraphics[width=0.49\textwidth]{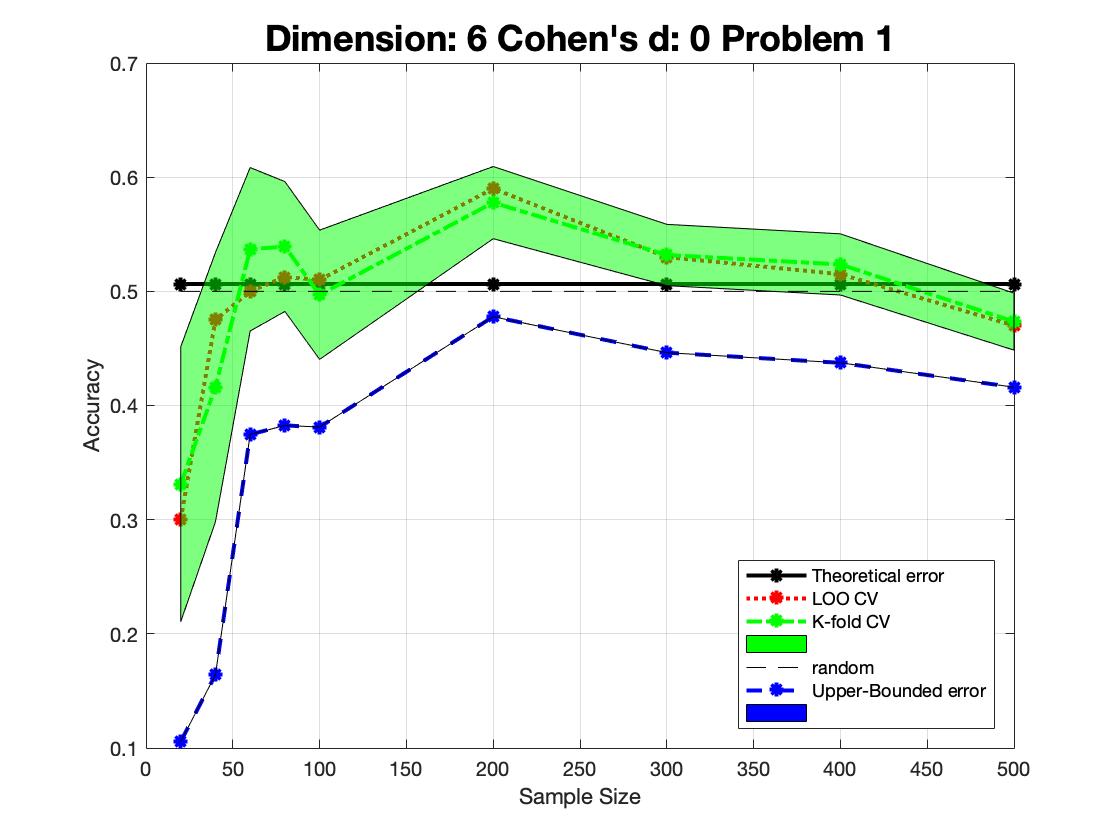} \\
            \includegraphics[width=0.49\textwidth]{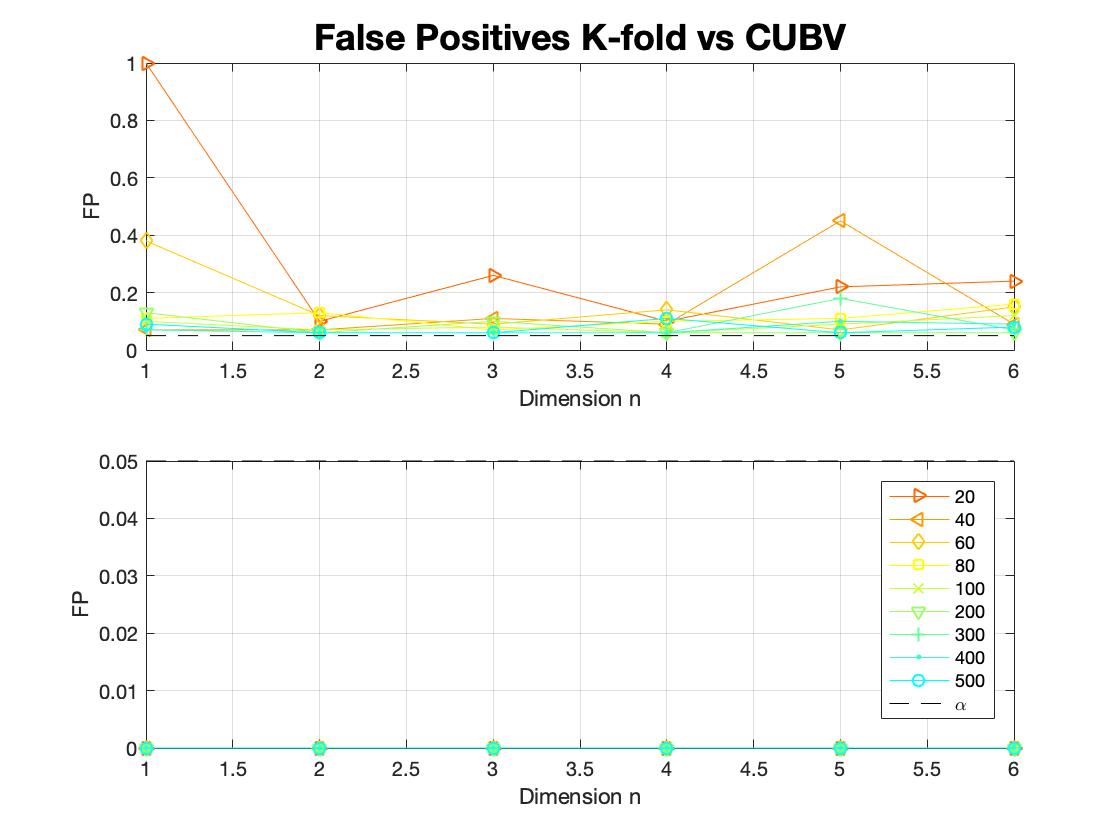} &
            \includegraphics[width=0.49\textwidth]{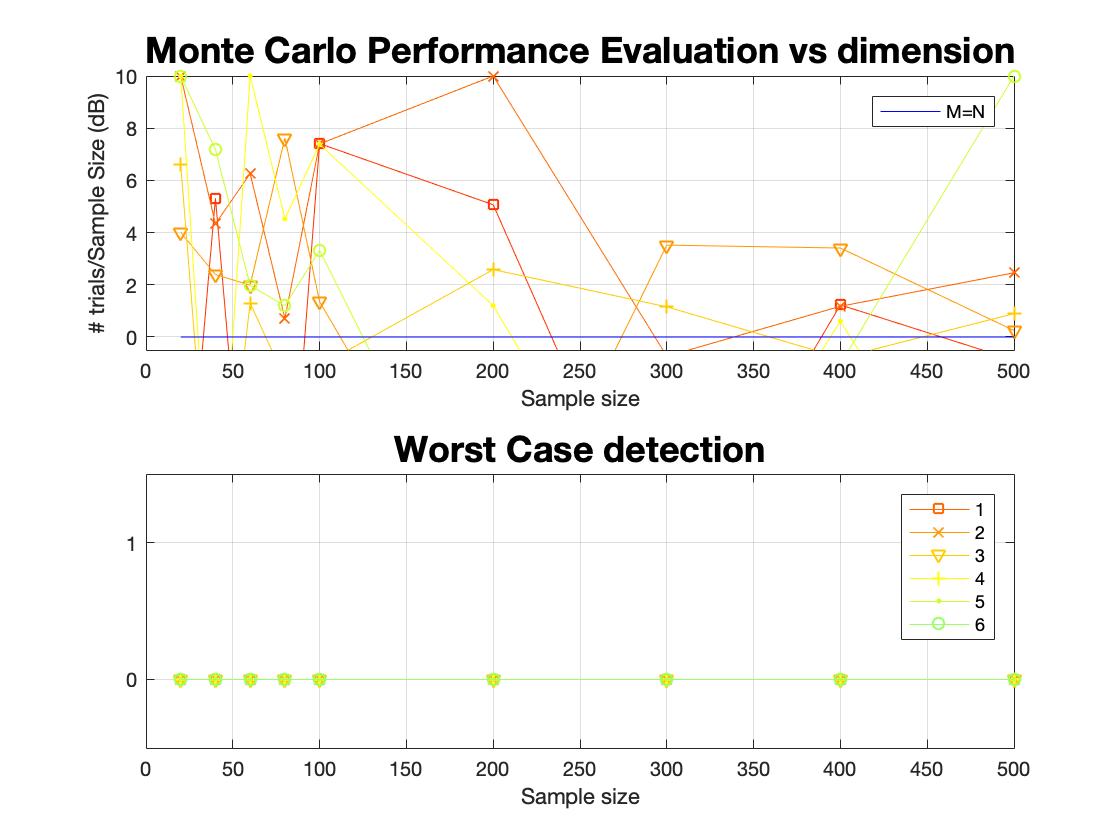}\\\\
        };
\draw[thick,blue!20] (a-1-1.north east) -- (a-3-1.south east);
\draw[thick,densely dashed,blue!20] (a-2-1.north west) -- (a-2-2.north east);
\draw[thick,densely dashed,blue!20] (a-3-1.north west) -- (a-3-2.north east);
\end{tikzpicture}
\caption{Examples of performance, FP rates and MC performance evaluation in single sample experiments.}
\label{fig:powernull2}
\end{figure*}

In figure \ref{fig:powernull2} we display the standalone analysis of the FP rate under the null-hypothesis with increasing dimension $n$ and sample size $N$. We plotted the performance of all the validation methods with $F=100$ samples obtained from the fold permutations of one single realization and $N_c=2$, the MC performance evaluation and the worst case detection analysis. This again highlights the ability of the K-fold CUBV to control type I errors unlike K-fold CV whose optimistic performance is confirmed by the MC performance evaluation (detection is achieved under the null hypothesis when colored lines go below the blue line: \ref{fig:powernull2}, bottom).

\subsection{Classification variability across independent (multi-sample) experiments}

\subsubsection{\textbf{K-fold variability vs complexity}}\label{sec:var}

\begin{figure*}
\centering
\begin{tikzpicture}
\matrix (a)[row sep=0mm, column sep=0mm, inner sep=1mm,  matrix of nodes] at (0,0) {
\includegraphics[width=0.75\textwidth]{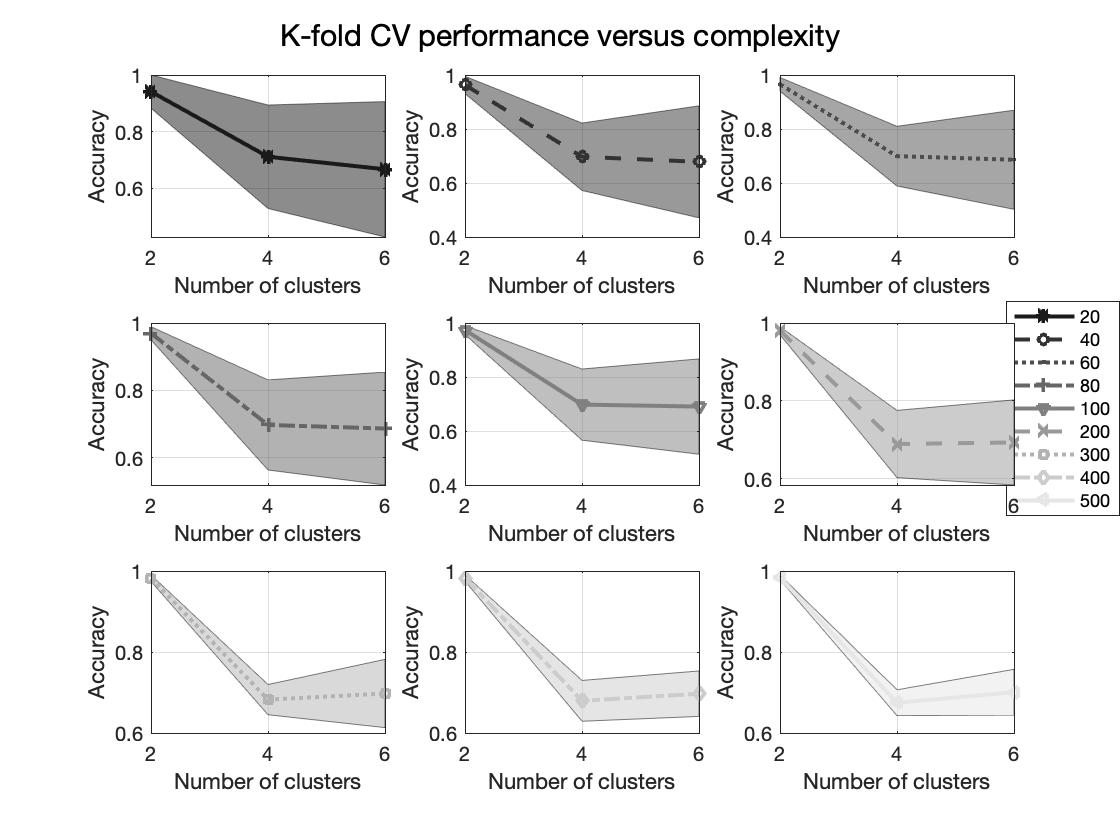}\\
\includegraphics[width=0.75\textwidth]{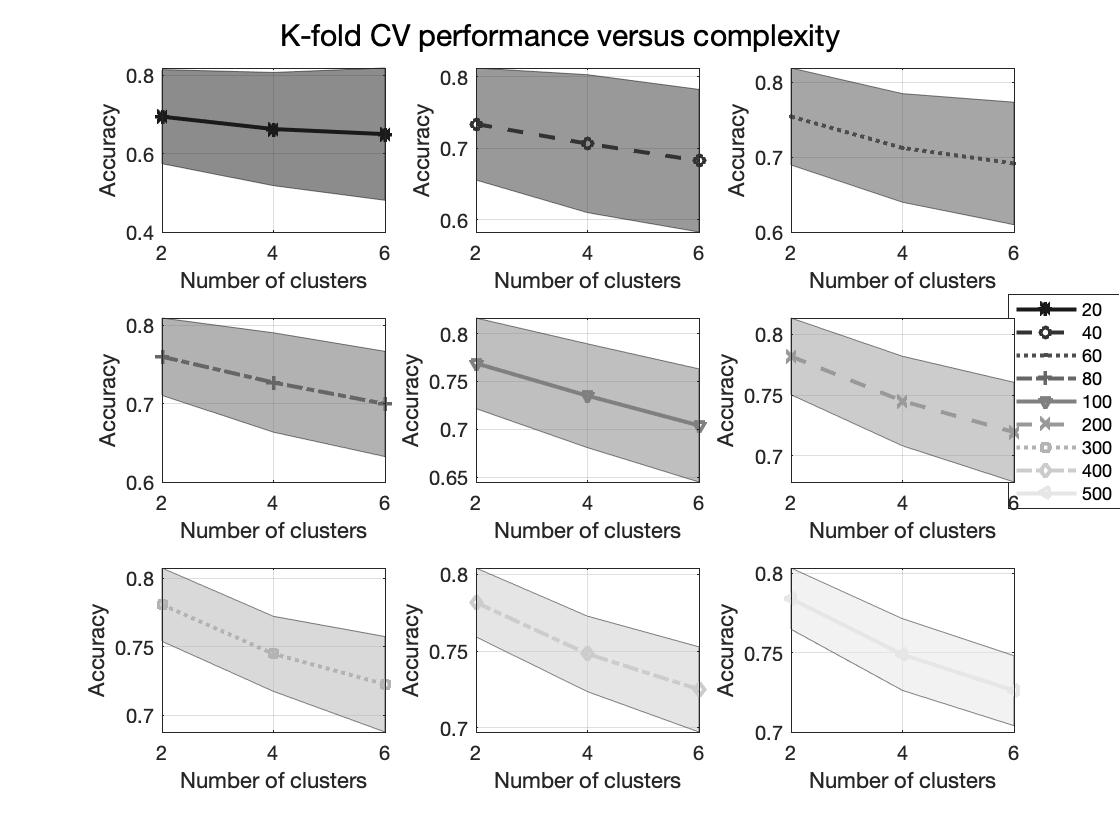}\\\\
        };
\draw[thick,gray] (a-2-1.north west) -- (a-2-1.north east);
\end{tikzpicture}
\caption{The accuracy values (average and standard deviation) obtained in K-fold CV versus complexity ($N_c$) and sample size $N$ with $M=1000$ and a large effect size, in a $n=1$ (top) and $n=6$ (bottom) binary classification task.}
\label{fig:complexity}
\end{figure*}

In figure \ref{fig:complexity} we show the classification values obtained averaging $M=1000$ experiments and $\{1,3,10\}$ simulations using K-fold CV as a function of the number of clusters (complexity) that generate the data, and sample size for an up to $6$ dimensional problem. The variability of the performance increases with complexity and sample size, although the dimensionality of the features can partly relieve this problem (see figure scale). Unfortunately, most of the methods usually employed in neuroimaging are univariate, typically voxelwise, and the conditions most frequently encountered are those reflected in the left side of figure \ref{fig:complexity}. The curves in this figure are compared to the estimated null-distribution to make an inference about the data, as shown in figure \ref{fig:powertest}.

\begin{figure*}
\centering
\begin{tikzpicture}
\matrix (a)[row sep=0mm, column sep=0mm, inner sep=1mm,  matrix of nodes] at (0,0) {
\includegraphics[width=0.75\textwidth]{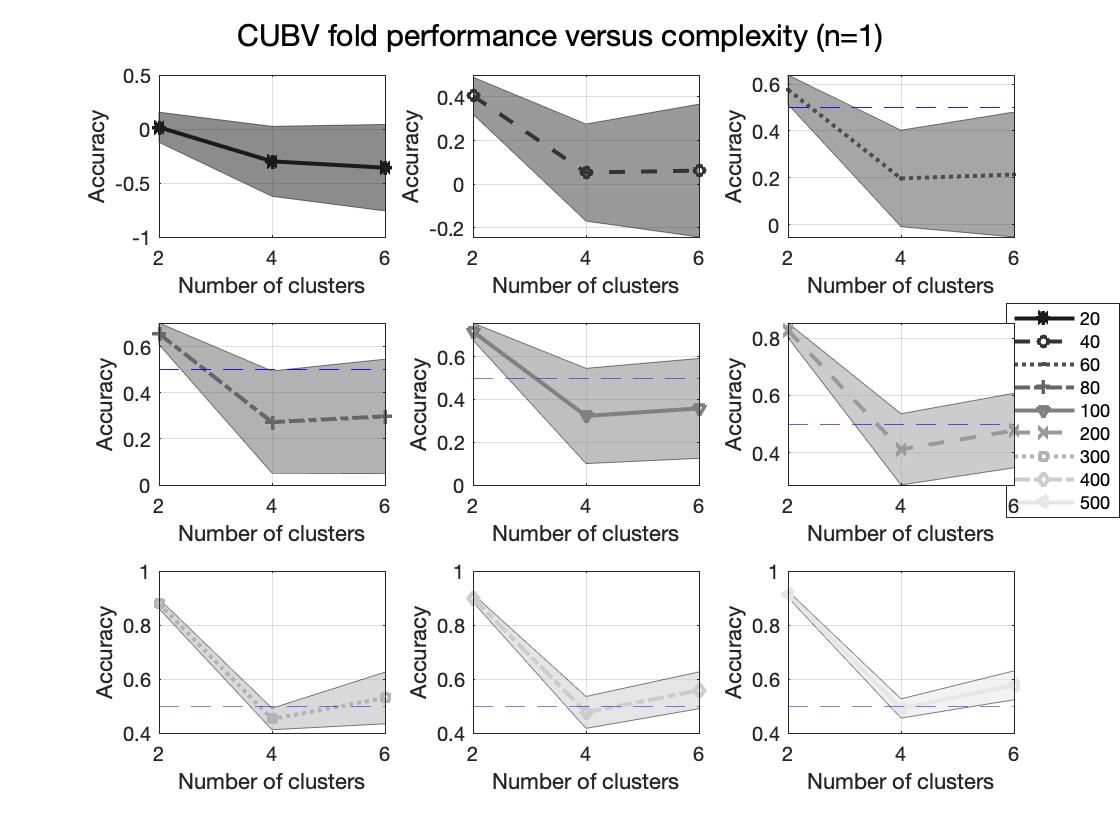}\\
\includegraphics[width=0.75\textwidth]{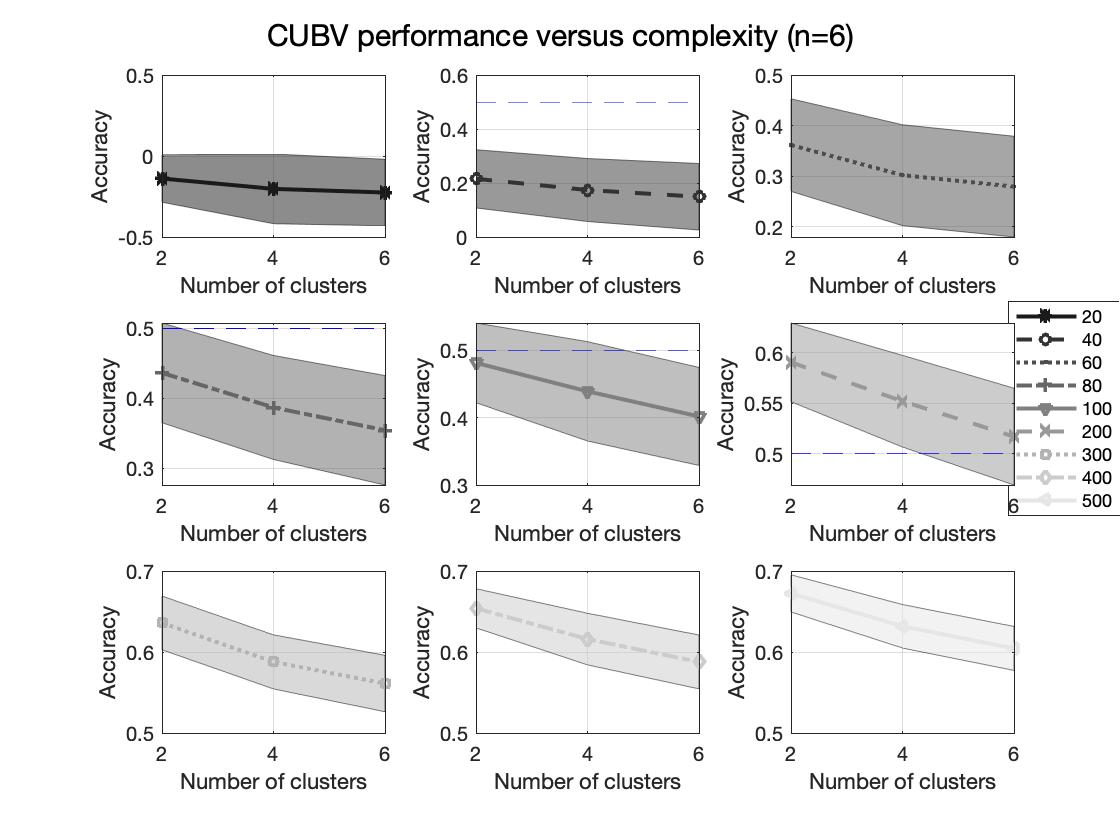}\\\\
        };
\draw[thick,gray] (a-2-1.north west) -- (a-2-1.north east);
\end{tikzpicture}
\caption{The accuracy values (average and standard deviation) obtained in CUBV versus complexity ($N_c$) and sample size $N$ with $M=1000$ and a large effect size, in a $n=1$ (top) and $n=6$ (bottom) binary classification task. Note the dashed line in blue corresponding to the random threshold.}
\label{fig:complexity2}
\end{figure*}

\subsubsection{\textbf{Power and detection analysis in single mode pdf}}\label{sec:powersingle}

\begin{figure*}
\centering
\begin{tikzpicture}
\matrix (a)[row sep=0mm, column sep=0mm, inner sep=1mm,  matrix of nodes] at (0,0) {
\includegraphics[width=0.75\textwidth]{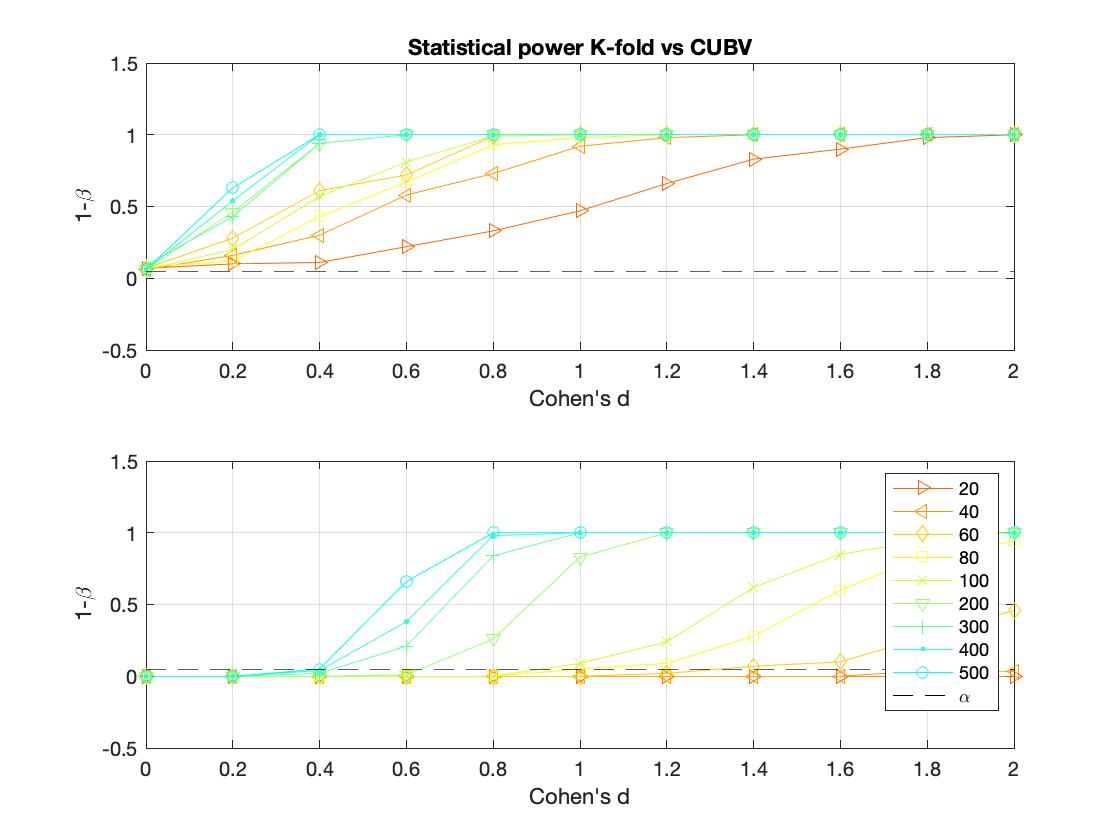}\\
\includegraphics[width=0.75\textwidth]{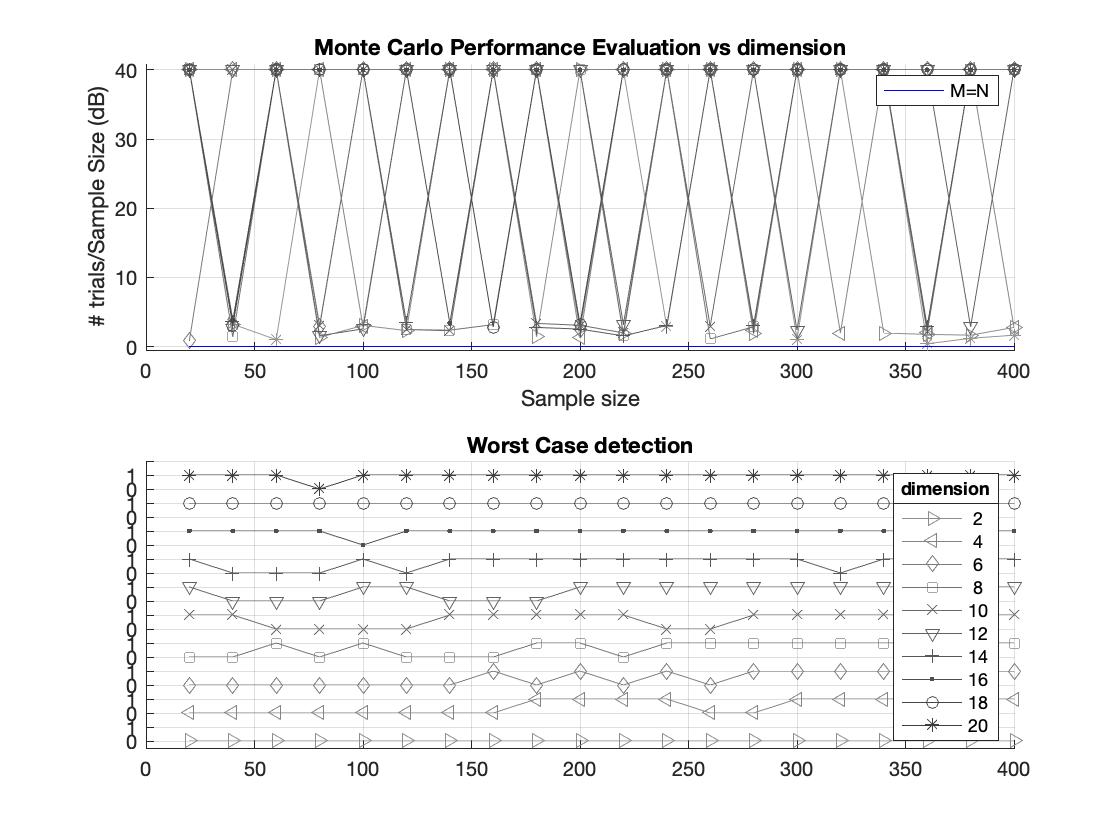}\\\\
        };
\draw[thick,blue!20] (a-2-1.north west) -- (a-2-1.north east);
\end{tikzpicture}
\caption{Analysis of the ideal case. Top-up: statistical power of K-fold and CUBV (top-down) CV permutation tests; bottom-up:  MC performance evaluation and CUBV detection (bottom-down).}
\label{fig:powertest}
\end{figure*}

In figure \ref{fig:powertest} we show the  behaviour of both approaches, K-fold CV and CUBV, with changes to Cohen's d and effect size in a $n=2$ classification task. Statistical power is computed against $d$ by evaluating the probability that $\text{p value}<0.05$  given the set of trials, $M$.  

When using the CUBV technique, the power of the test is computed by thresholding at the level of  random chance (equation \ref{eq:test}). However, if we compute the number of required MC trials to achieve detection with a degree of confidence $\alpha=0.05$ we obtain an unexpected result, as shown in figure \ref{fig:powertest} on the right. The number of required MC trials to achieve detection for small effect sizes is about $7$ times the sample size. Thus, for example, if $N=500$ then $M\sim 3700$, even under the controlled experimental conditions of this simulation (samples drawn from a Gaussian pdf). Conversely, the CUBV technique achieves a significant detection with just a few samples. It's clear that based on the information on the right of figure \ref{fig:powertest}, we can validate the results obtained with the classical ML-based permutation test using this technique, unlike the MC evaluation as it theoretically requires many trials to achieve detection.

\subsubsection{\textbf{Power and detection analysis in multi-mode pdf}}\label{sec:powermulti}

We repeated the power analysis of the previous section ($n=2$ and $N_{c}=4$) in the non-separable case predicted by the capacity measure (see figure \ref{fig:powertest2}) assuming a balanced sample per cluster and per group (i.e. sources have the same effect on the observed variable), and with a ratio $r=1/3$ per cluster in both classes (imbalanced case). 

Here, we readily see that the detection ability of the test using K-fold CV is less powerful than in the ``ideal'' case and greater than the level of significance with no effect ($d=0$), thus providing a FP rate greater than the nominal value. 

The MC performance is worse than in the previous case, e.g. the curve does not decrease with sample size at values between 2 and 3 dB in the worst scenario (balanced sample), thus the number of trials needed to achieve detection at the level of confidence is between $7-20$ times the sample size. The CUBV technique is expected to achieve detection with fewer samples ($(N,d)=(200,4)$ and $(300,3)$, respectively). In the imbalanced case, the benefits of the CUBV approach are even clearer as it preserves detection ability whilst controlling FP.

\begin{figure*}
\centering
\begin{tikzpicture}
\matrix (a)[row sep=0mm, column sep=0mm, inner sep=1mm,  matrix of nodes] at (0,0) {
            \includegraphics[width=0.49\textwidth]{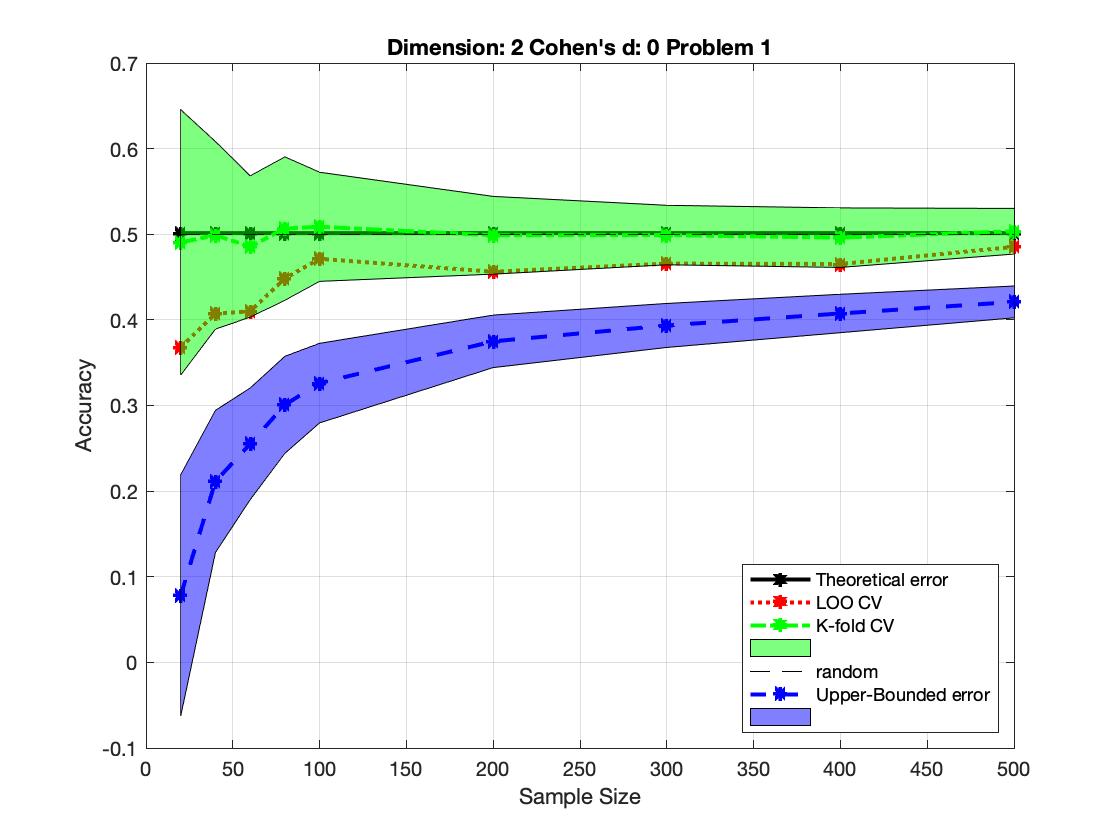}&
            \includegraphics[width=0.49\textwidth]{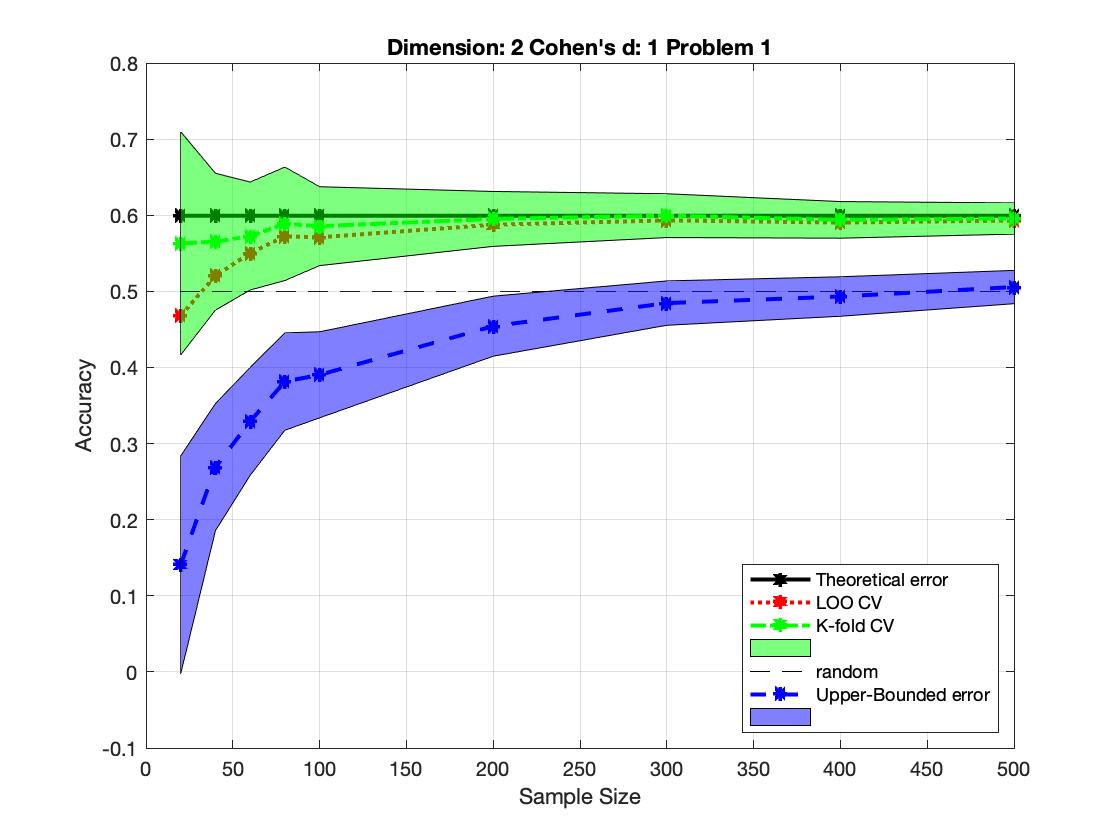}\\
            \includegraphics[width=0.49\textwidth]{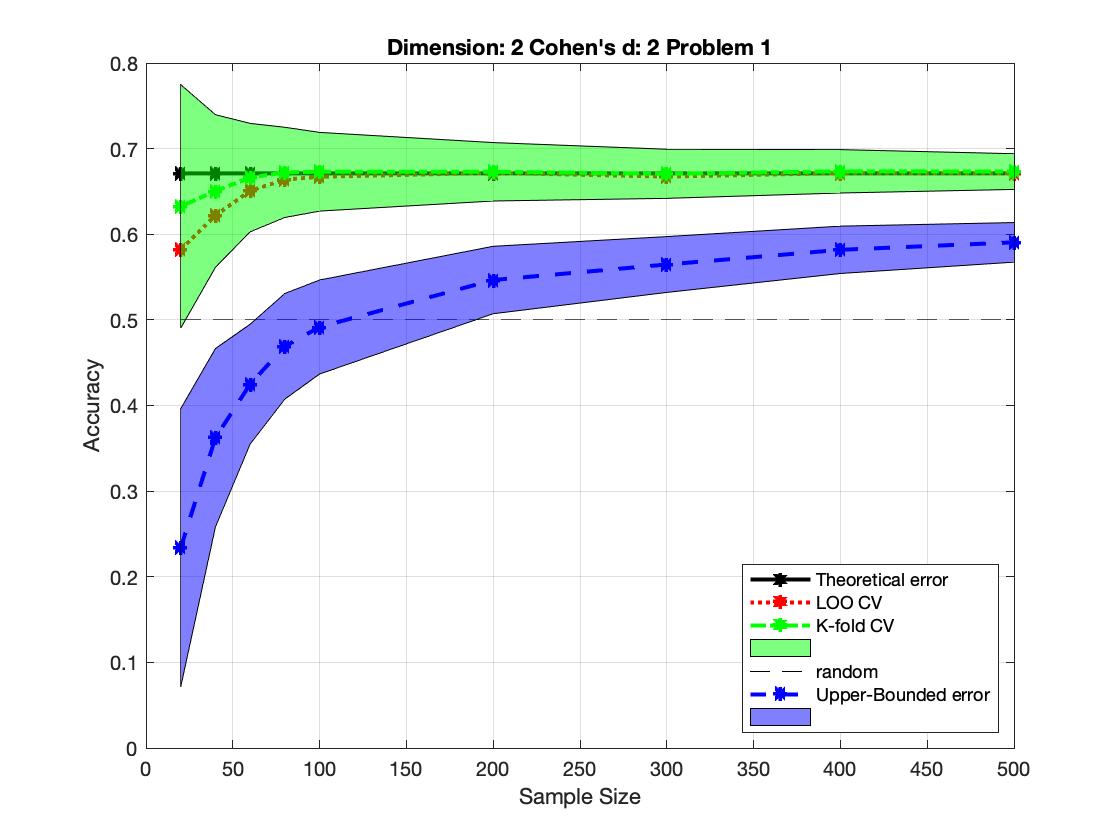}&
            \includegraphics[width=0.49\textwidth]{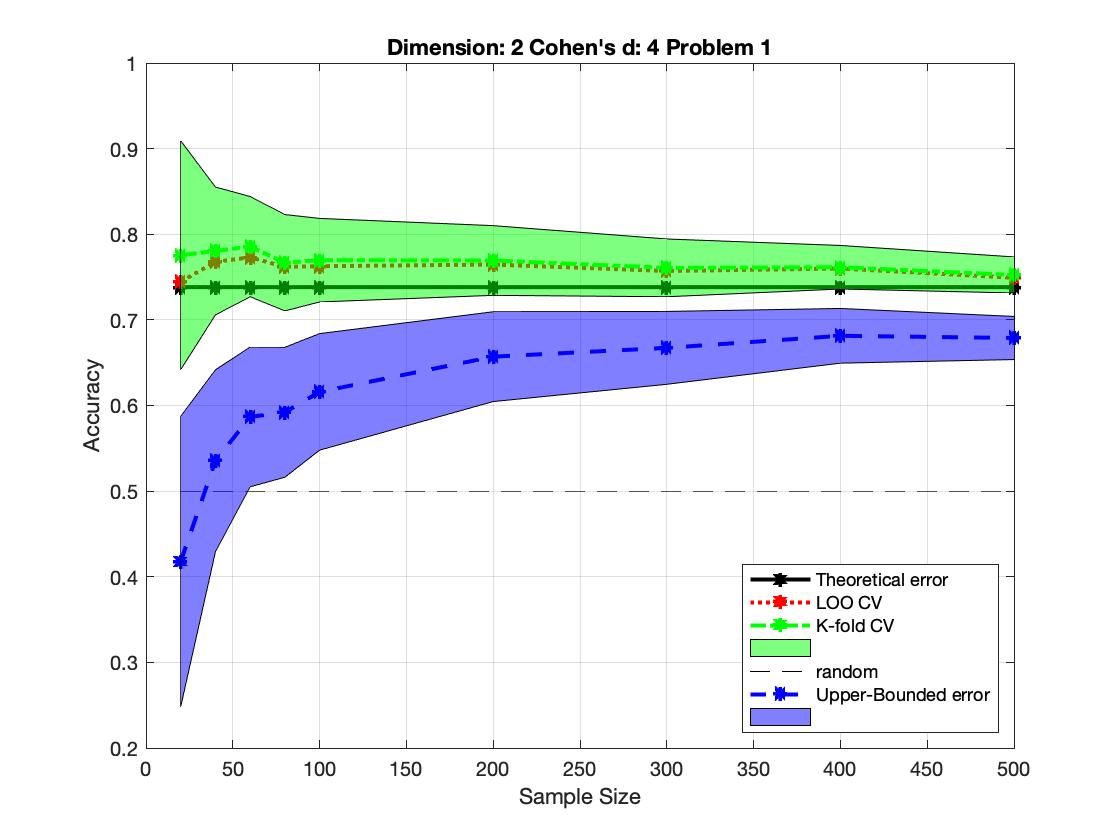}\\\\
        };
\draw[thick,blue!20] (a-1-1.north east) -- (a-2-1.south east);
\draw[thick,blue!20] (a-2-1.north west) -- (a-2-2.north east);
\end{tikzpicture}
\caption{Distribution of accuracy values ($M=100$) vs. sample size and $d$ for the non-ideal case. We show a $n=2$ classification problem sampling from $N_c=4$ Gaussian pdfs ($2$ per cluster) using an imbalanced dataset ($r=1/3$) and $d=\{0, 1, 2, 4\}$. Note the biased regions within the green area.}
\label{fig:exampleslast}
\end{figure*}

Finally, figure \ref{fig:exampleslast} illustrates the operation of the CUBV approach and the regular K-fold CV against $d$ for $n=2$. Substantial control of FP is achieved by the combination of both approaches; figure \ref{fig:exampleslast} top left. Optimistic or conservative estimations of the ``real effect'' (small, medium and large effects) are predicted by the CUBV method with errors below $0.5$ at small sample sizes.  

\subsection{Classification variability across CV-folds in single sample experiments}

\subsubsection{\textbf{Performance analysis including complexity}}

This simulation is close to empirical studies commonly found in the contemporary literature. Given one, and only one dataset with an increasing number of clusters or data complexity we ran a permutation test by randomly selecting training and testing folds repeatedly ($F=100$) under varying sample size ($N:20-500$), dimension ($n:1-6$) and complexity ($N_{c}:2-6$) using a prior procedure \cite{Gorriz19}. The effects considered were medium to large as shown in figure \ref{fig:example3}. 

\begin{figure*}
\centering
\includegraphics[width=0.49\textwidth]{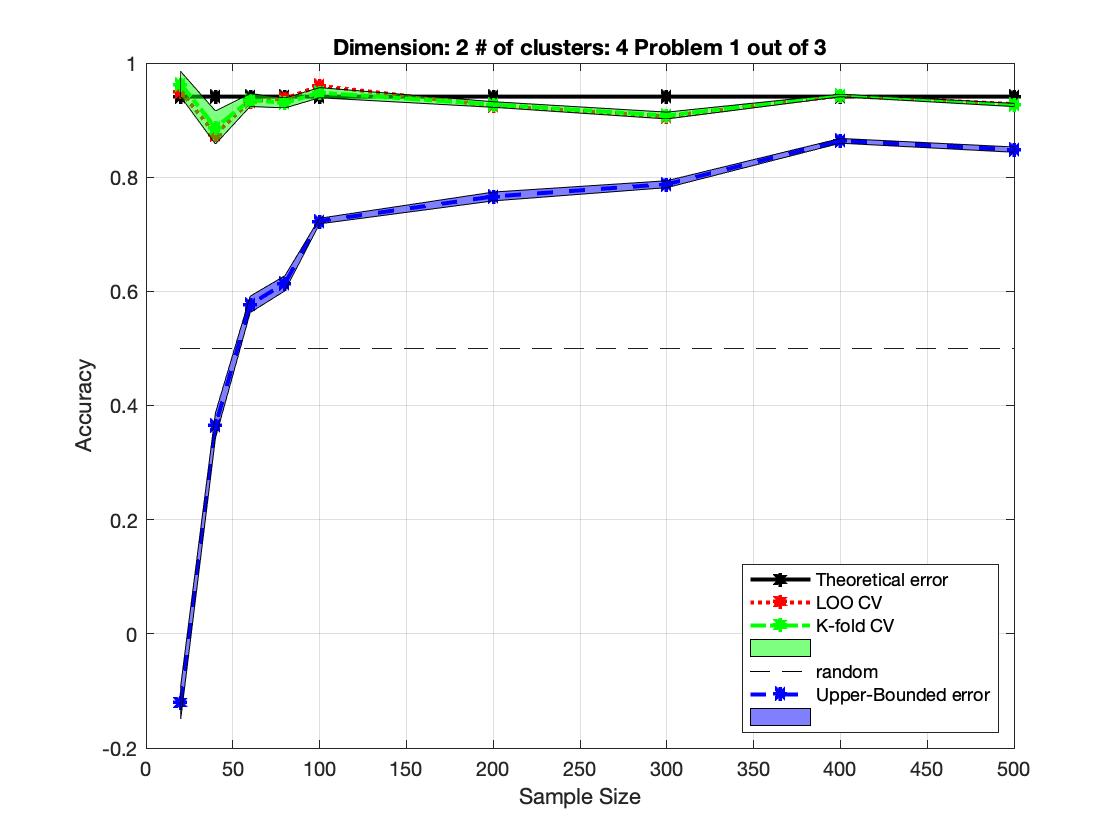}
\includegraphics[width=0.49\textwidth]{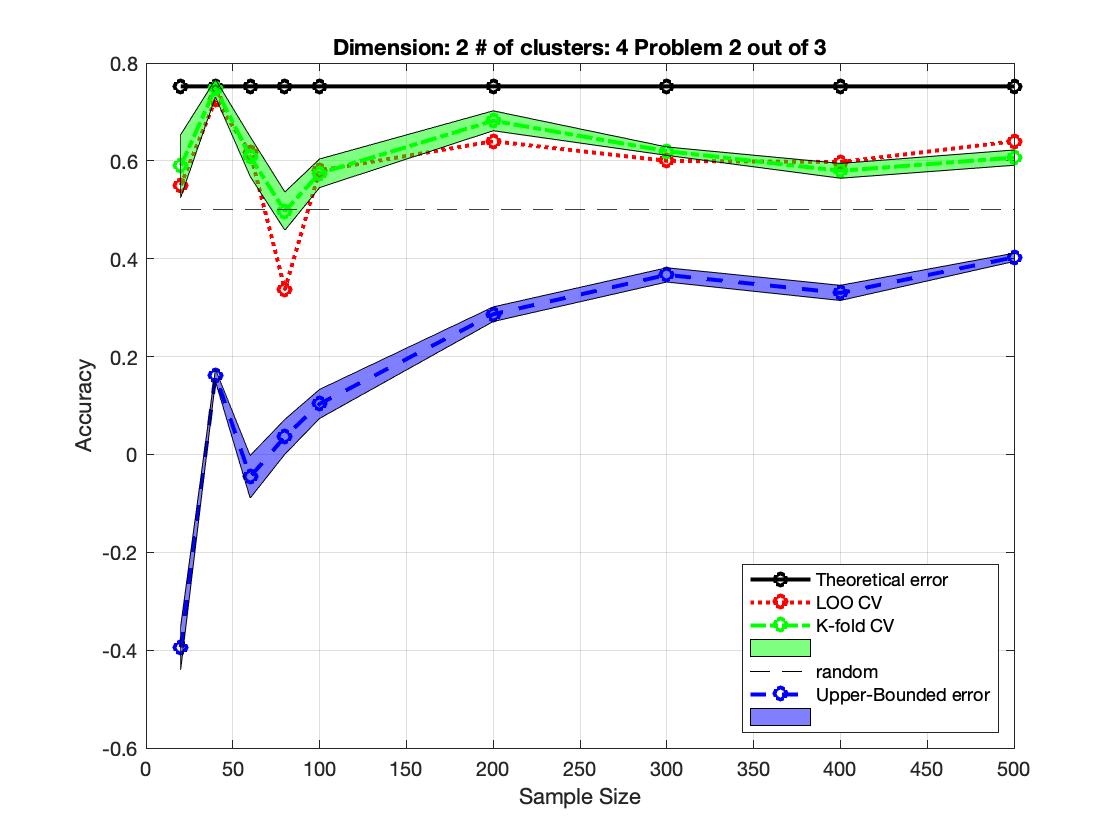}
\includegraphics[width=0.49\textwidth]{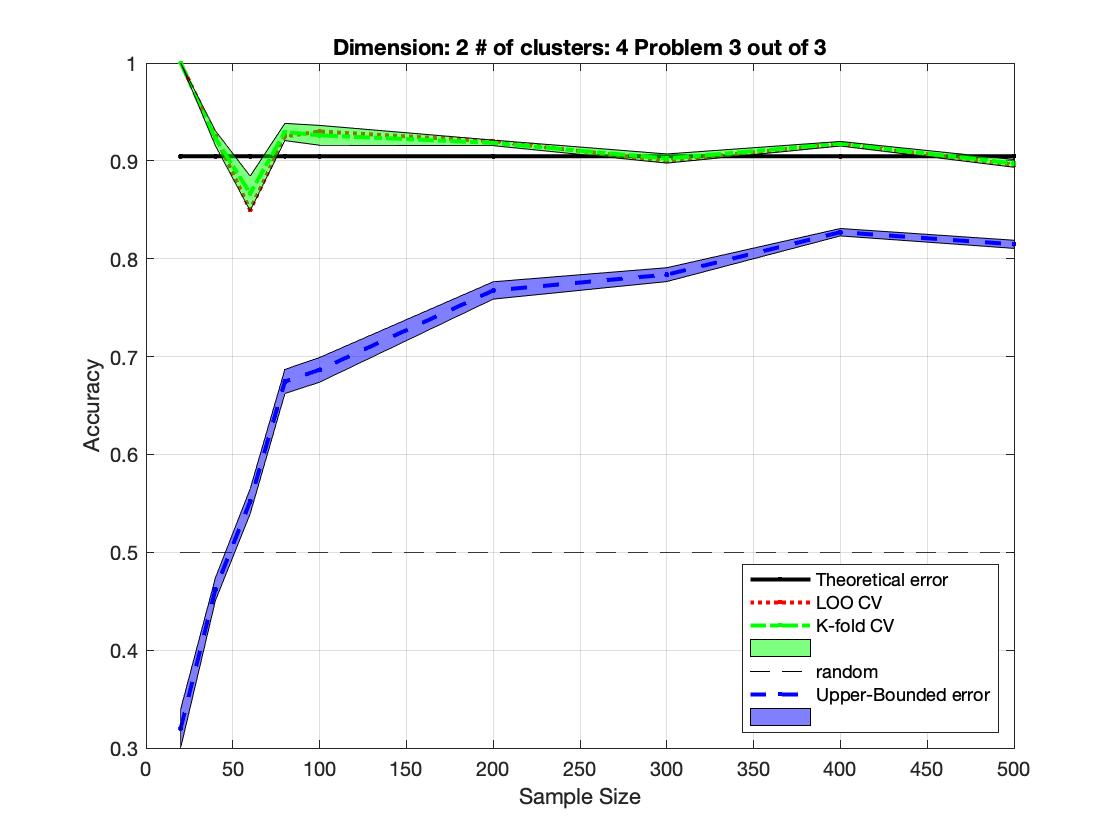}
\caption{Examples of classical permutation tests based on regular CV and CUBV decisions depending on sample variability and classifier complexity. Note that whenever there is a clear deviation between the green and black lines or a high variability in K-fold CV is obtained, the CUBV decision is below $0.5$ and thus the test does not reject the null hypothesis. This decision is warning that the green curve values obtained cannot be extrapolated to other experiments thus the analysis by another laboratory could be providing substantially different outcomes, specially in problem 2 out of 3.}
\label{fig:permtest}
\end{figure*}

\begin{figure*}
\centering
\begin{tikzpicture}
\matrix (a)[row sep=0mm, column sep=0mm, inner sep=1mm,  matrix of nodes] at (0,0) {
\includegraphics[width=0.75\textwidth]{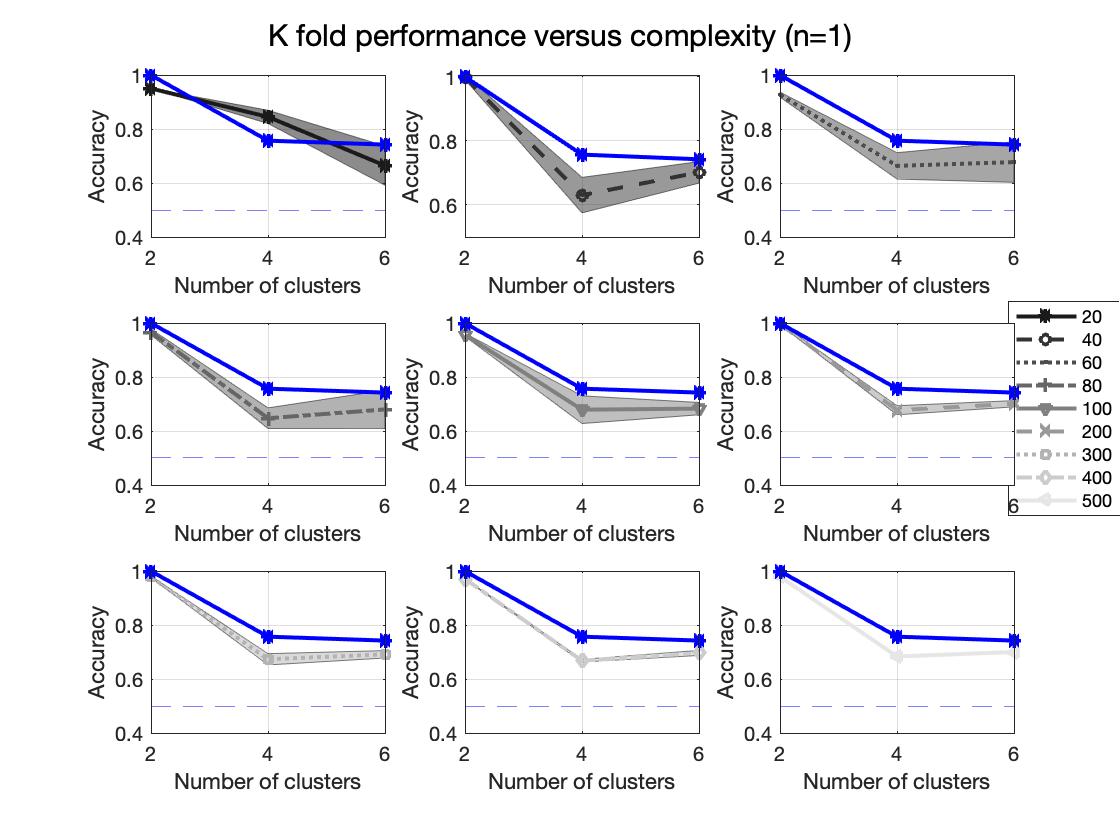}\\
\includegraphics[width=0.75\textwidth]{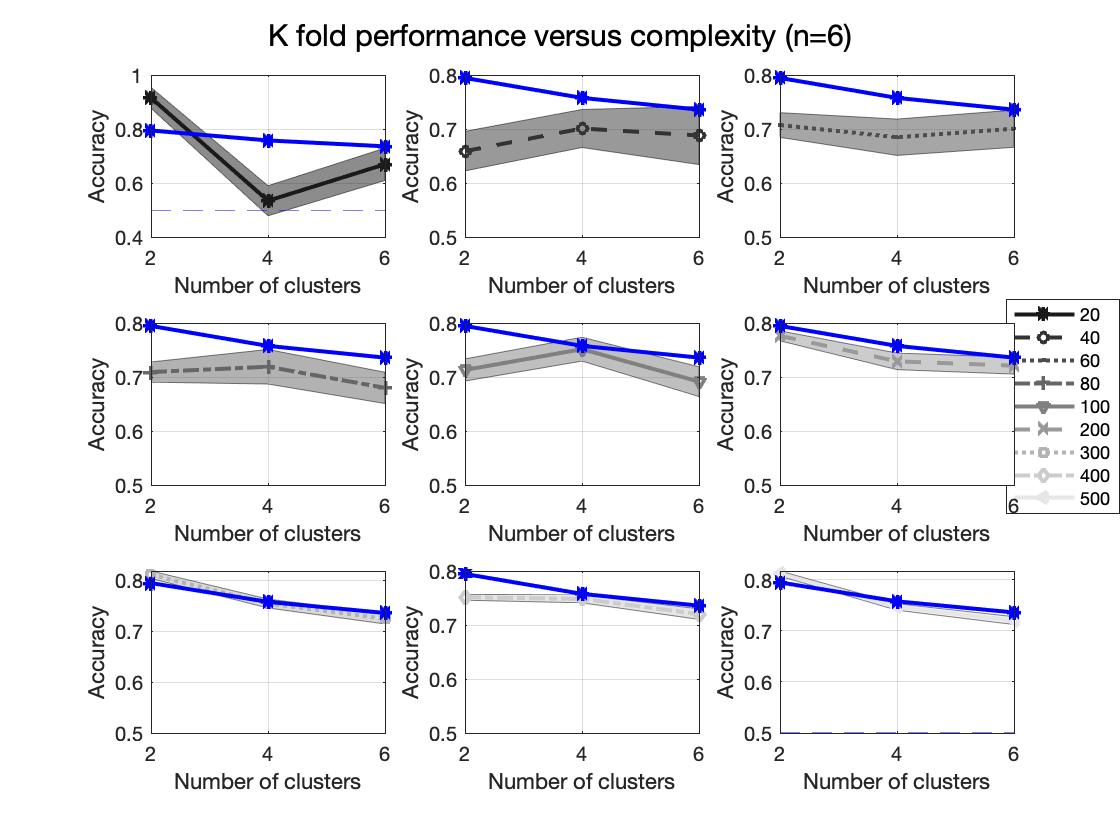}\\\\
        };
\draw[thick,gray] (a-2-1.north west) -- (a-2-1.north east);
\end{tikzpicture}
\caption{The analysis depicted in figure \ref{fig:complexity} is replicated here using a single realization. Observe the theoretical accuracy highlighted by the blue line}
\label{fig:complexityS}
\end{figure*}

\begin{figure*}
\centering
\begin{tikzpicture}
\matrix (a)[row sep=0mm, column sep=0mm, inner sep=1mm,  matrix of nodes] at (0,0) {
\includegraphics[width=0.75\textwidth]{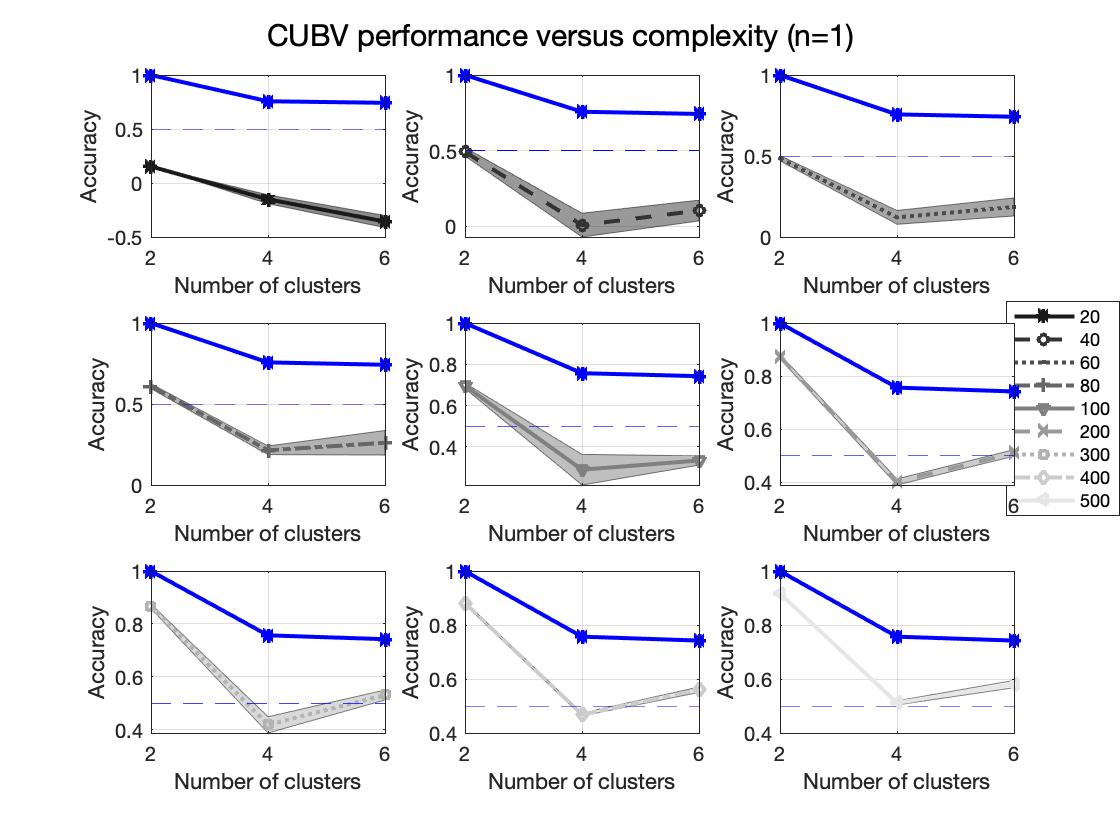}\\
\includegraphics[width=0.75\textwidth]{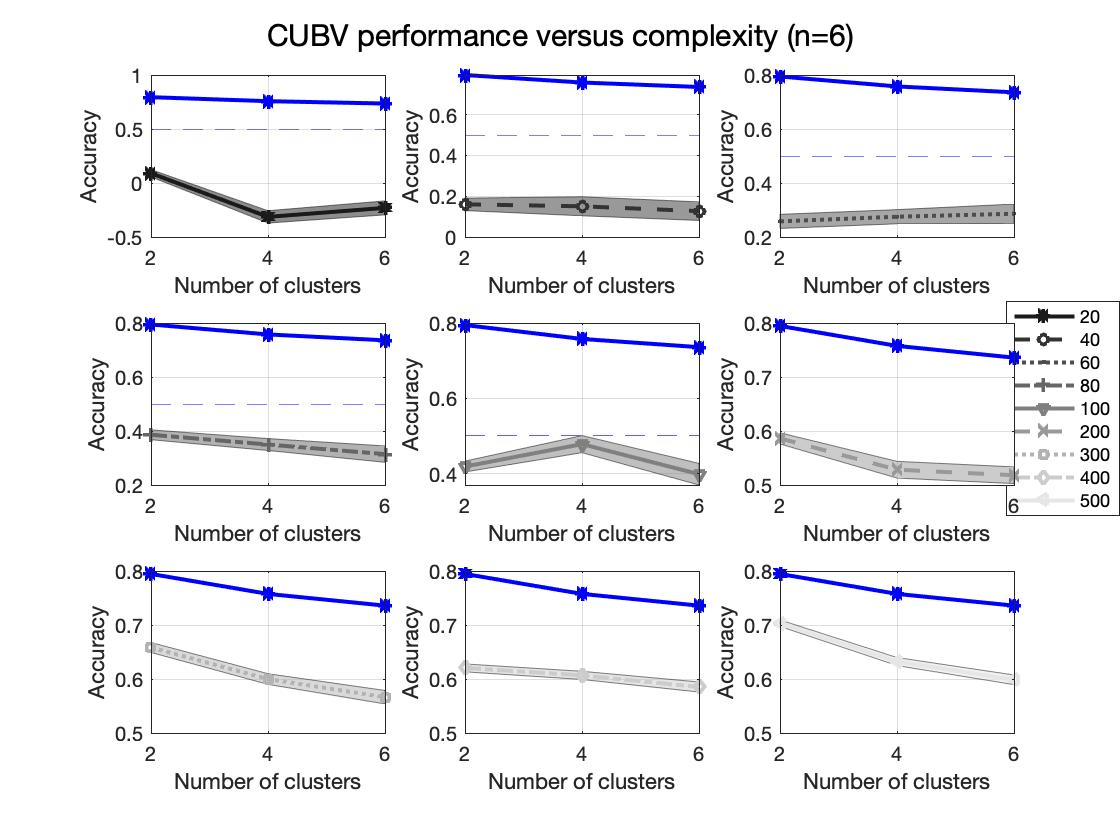}\\\\
        };
\draw[thick,gray] (a-2-1.north west) -- (a-2-1.north east);
\end{tikzpicture}
\caption{The same analysis as in figure \ref{fig:complexity2} using a single realisation}
\label{fig:complexity2S}
\end{figure*}

As previously, the behaviour of the estimator of the error and its average were not as symmetric as the ideal case, with consequences for small sample-sizes and low dimensions, as predicted by the capacity measures detailed in section \ref{sec:complex}. As an example, in figure \ref{fig:permtest} for $n=2$ and $N_{c}=4$ there was $1$ out of $3$ situations (number of cases out of different simulations that cannot be shattered) where the estimation was biased towards underestimating the real effect. In this case (2 out of 3), the test developed in section \ref{sec:novelstat} did not reject the null-hypothesis (no effect) for any value of $N$. In cases 1 and 3, significant rejection is achieved above $N=50$ samples using K-fold CUBV. In figures \ref{fig:complexityS} and \ref{fig:complexity2S} we repeated the same analysis carried out in section \ref{sec:var} with a single realisation and averaging over the set of simulations with increasing complexity.

\subsubsection{\textbf{Power and detection analysis in single mode pdf}}

\begin{figure*}
\centering
\begin{tikzpicture}
\matrix (a)[row sep=0mm, column sep=0mm, inner sep=1mm,  matrix of nodes] at (0,0) {
\includegraphics[width=0.49\textwidth]{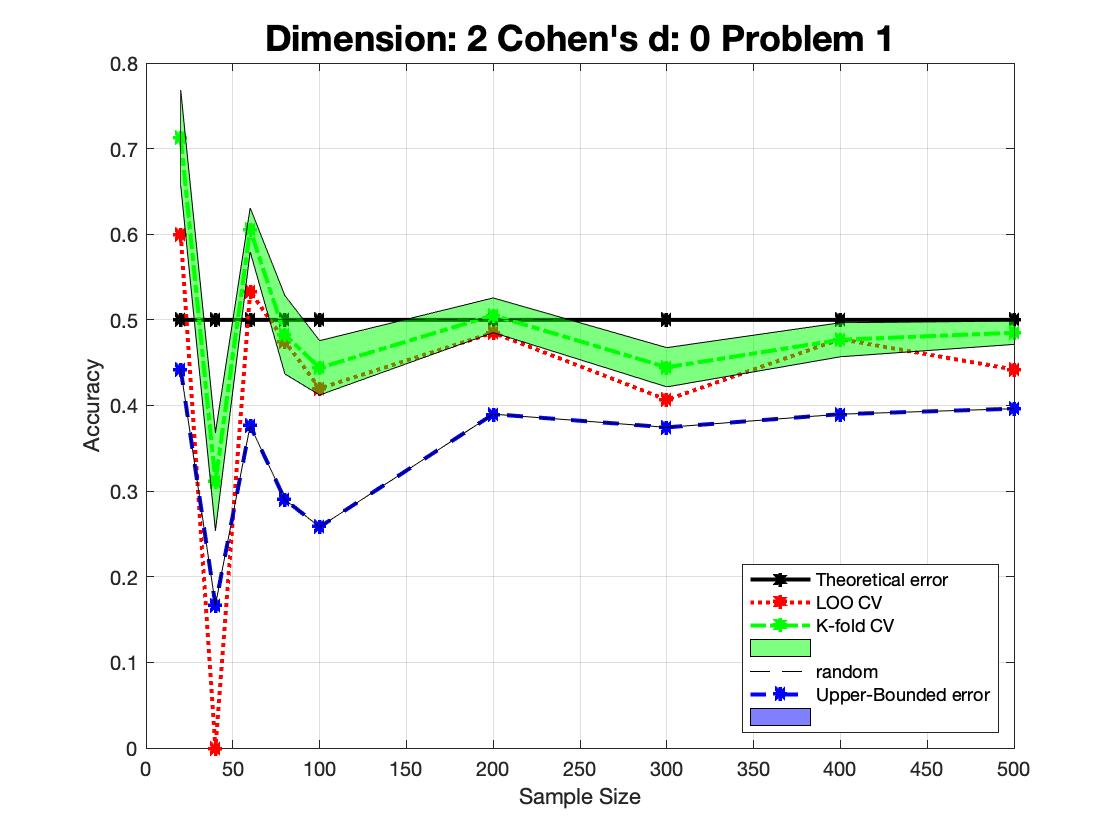} &
            \includegraphics[width=0.49\textwidth]{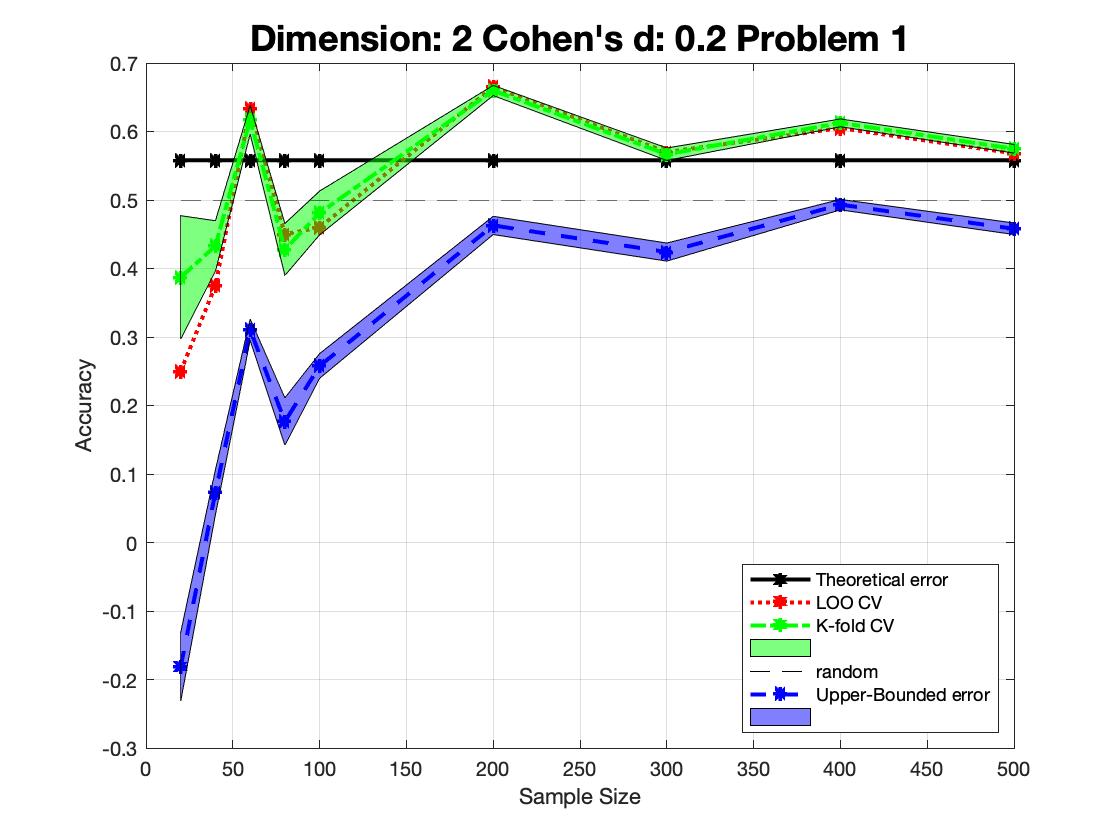} \\
            \includegraphics[width=0.49\textwidth]{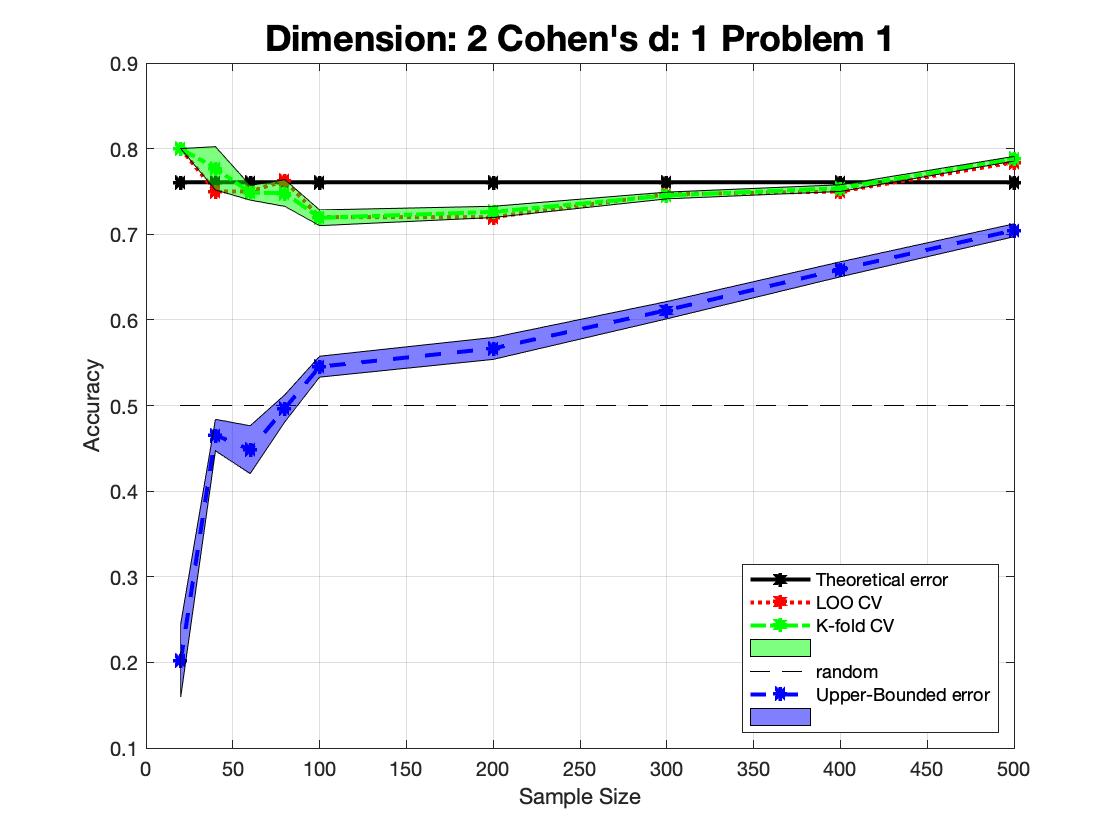} &
            \includegraphics[width=0.49\textwidth]{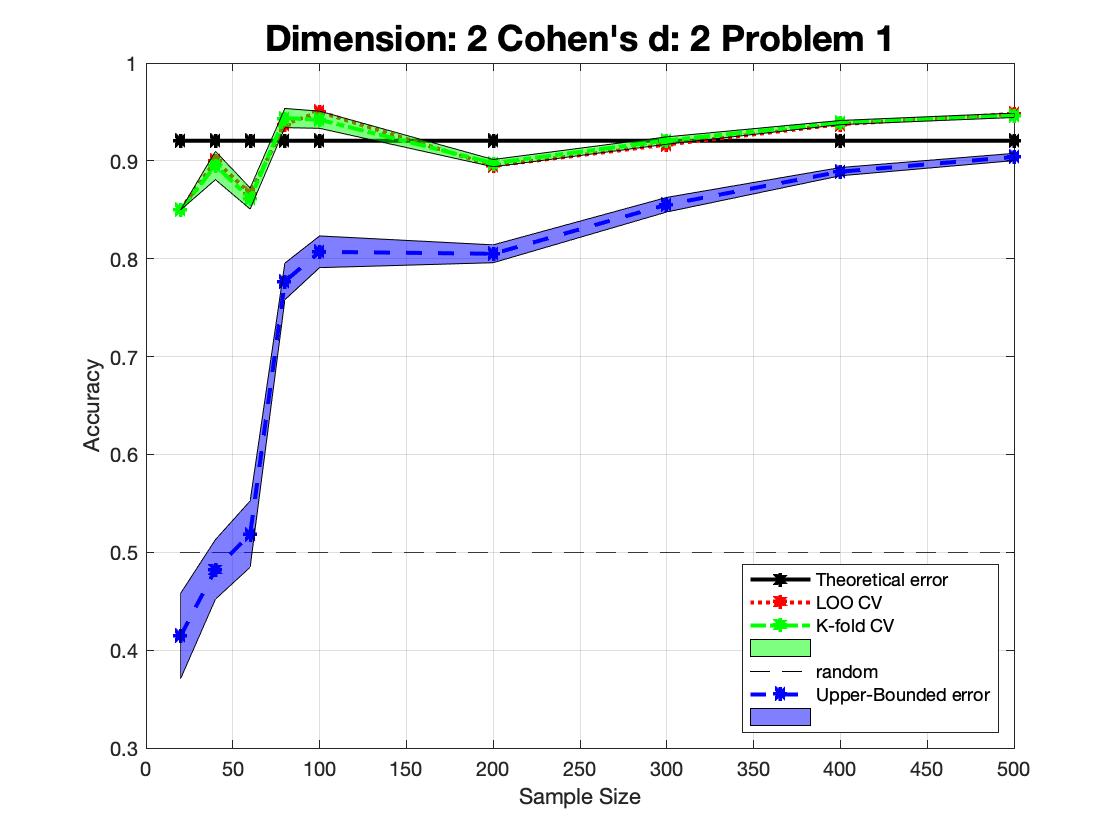} \\
            \includegraphics[width=0.49\textwidth]{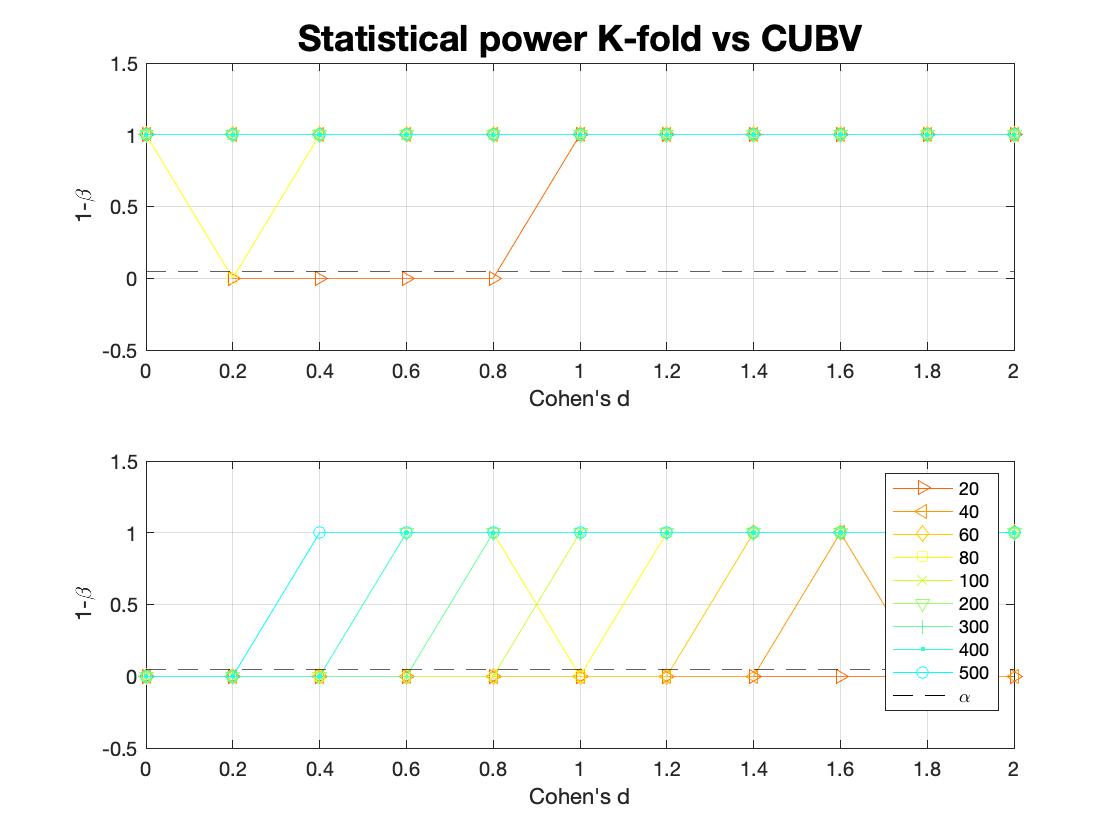} &
            \includegraphics[width=0.49\textwidth]{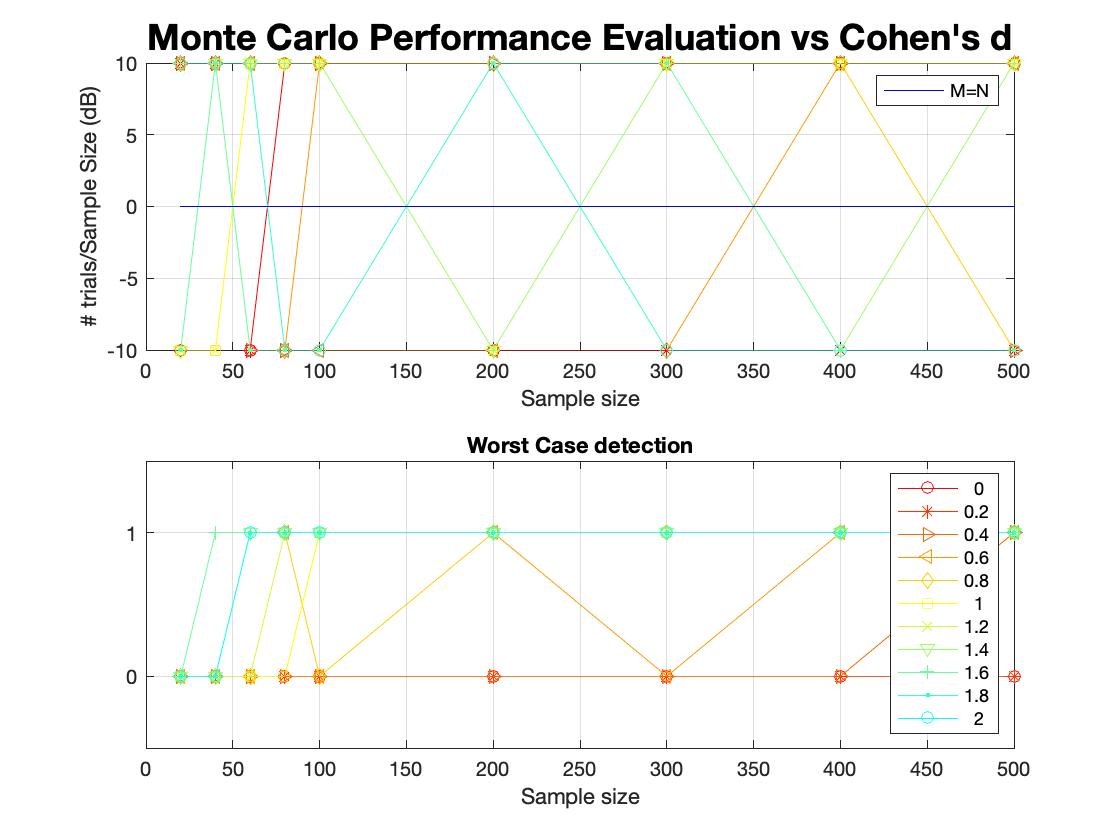}\\\\
        };
\draw[thick,blue!20] (a-1-1.north east) -- (a-3-1.south east);
\draw[thick,densely dashed,blue!20] (a-2-1.north west) -- (a-2-2.north east);
\draw[thick,densely dashed,blue!20] (a-3-1.north west) -- (a-3-2.north east);
\end{tikzpicture}
\caption{Examples, power and detection analysis in single-mode pdf using a single sample realization}
\label{fig:powertestsingle}
\end{figure*}

With the same parameter configuration at the previous section \ref{sec:powersingle}, we computed the same measures using only a single realization of the sample. In figure \ref{fig:powertestsingle} we show the expected behaviour of both approaches, K-fold CV and CUBV, varying Cohen's d and sample size in a $n=2$ classification task. Statistical power was computed by evaluating the probability that $\text{p }<0.05$  for a set of trials, $F=100$. Surprisingly, the number of FPs for a number of parameter configurations of the K-fold CV method was clearly admissible even at large sample sizes.

\subsubsection{\textbf{Power and detection analysis in multi-mode pdf}}

\begin{figure*}
\centering
\begin{tikzpicture}
\matrix (a)[row sep=0mm, column sep=0mm, inner sep=1mm,  matrix of nodes] at (0,0) {
            \includegraphics[width=0.49\textwidth]{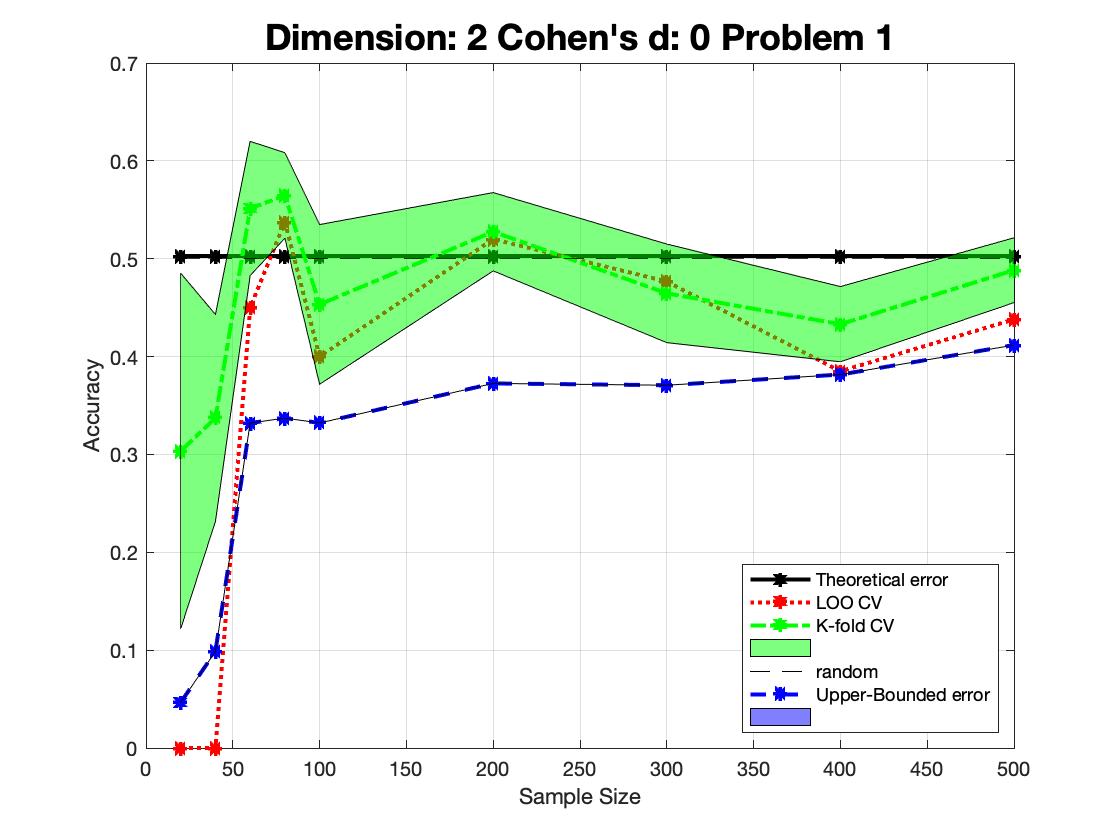} &
            \includegraphics[width=0.49\textwidth]{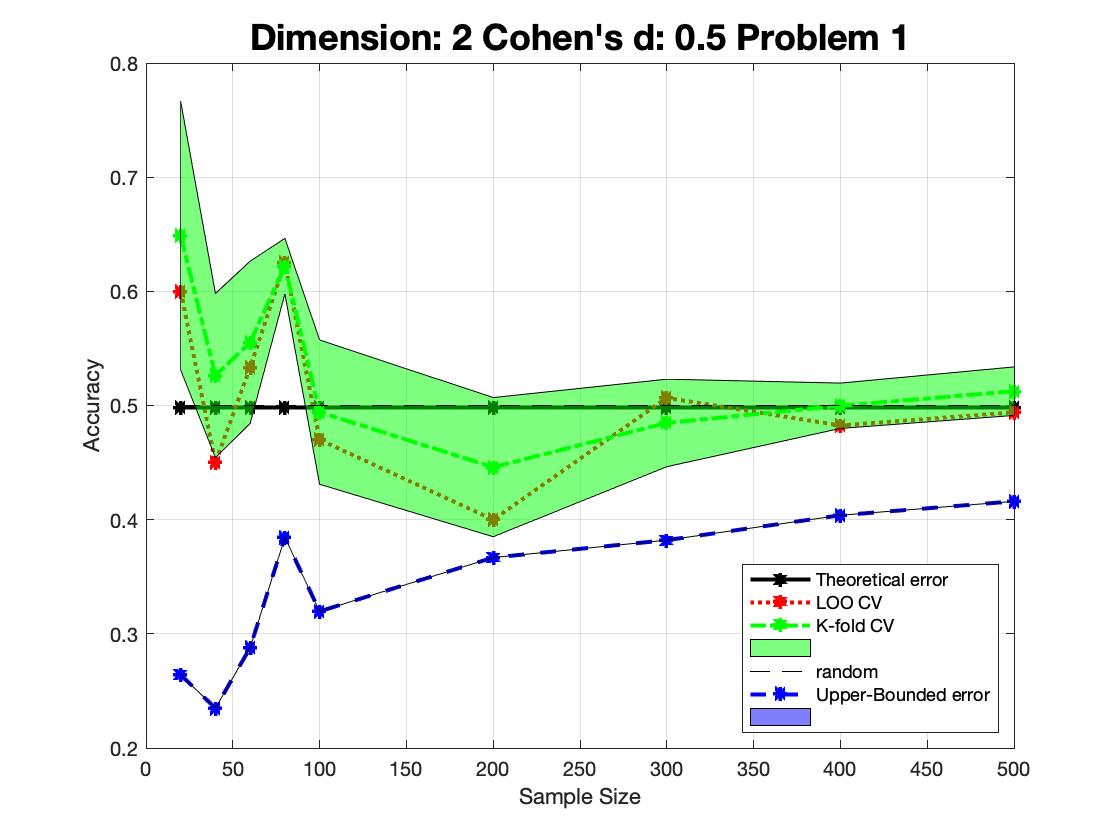} \\
            \includegraphics[width=0.49\textwidth]{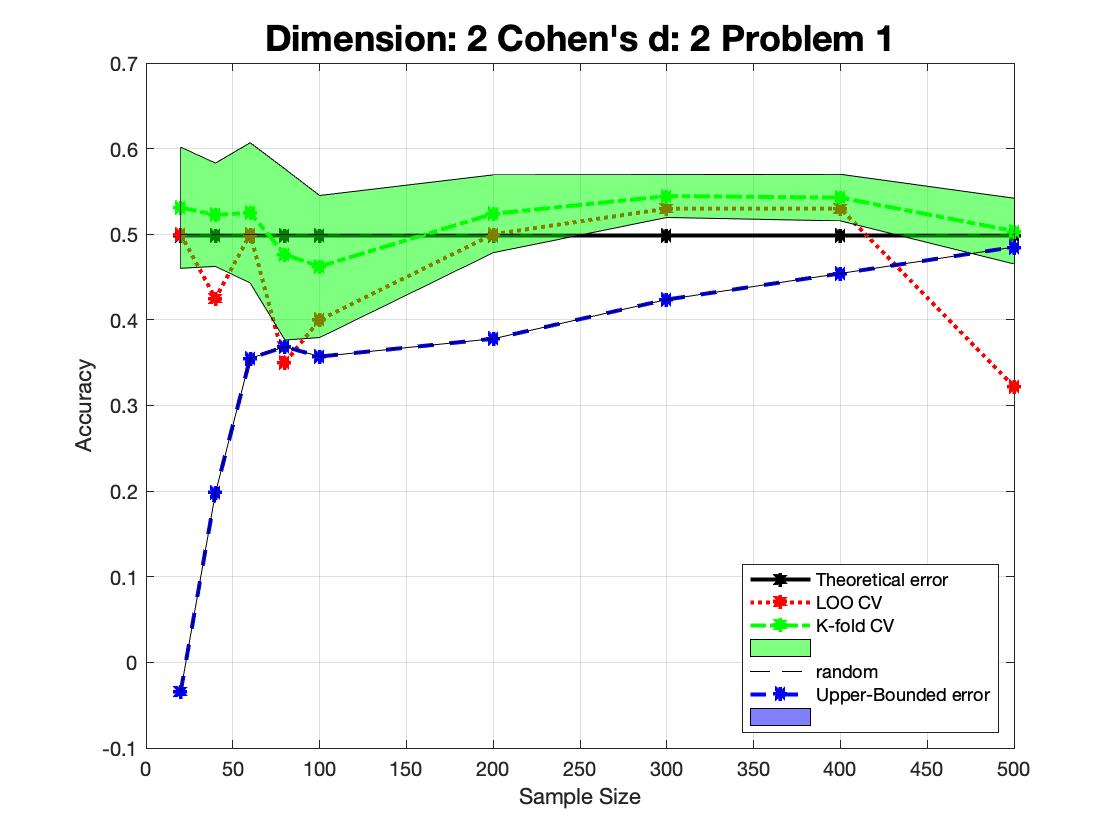} &
            \includegraphics[width=0.49\textwidth]{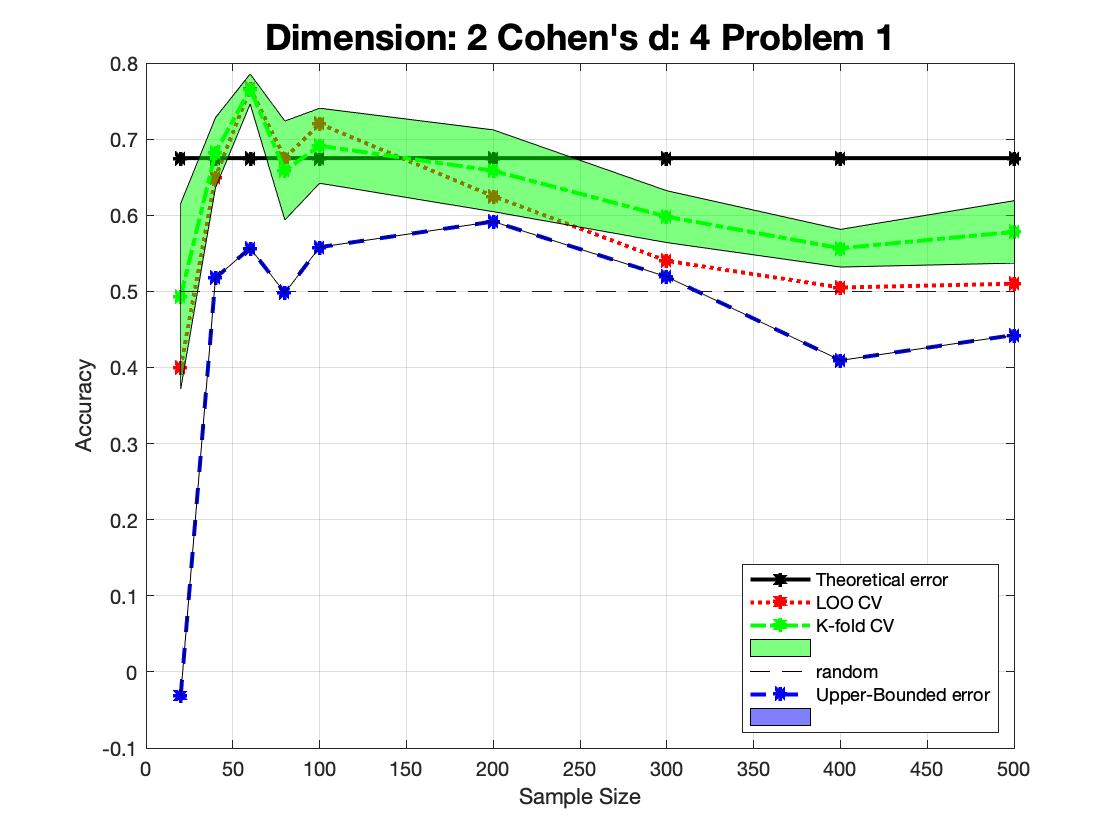} \\
            \includegraphics[width=0.49\textwidth]{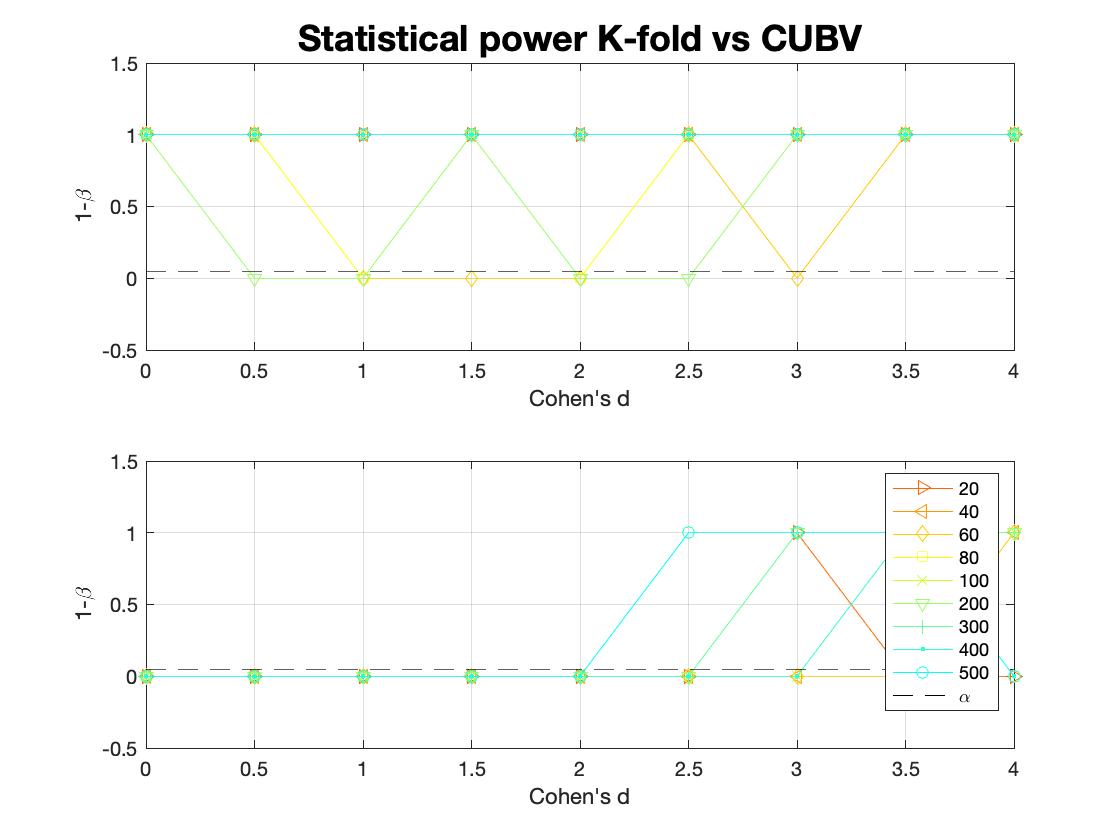} &
            \includegraphics[width=0.49\textwidth]{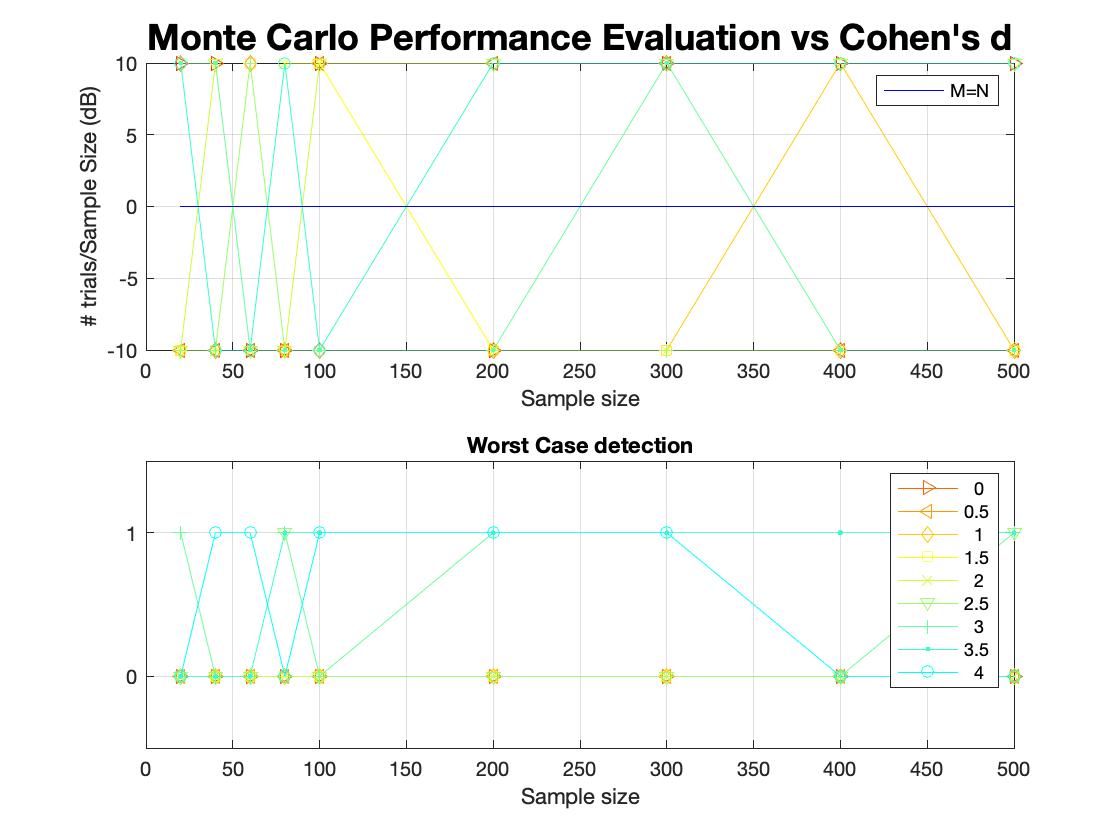}\\\\
        };
\draw[thick,blue!20] (a-1-1.north east) -- (a-3-1.south east);
\draw[thick,densely dashed,blue!20] (a-2-1.north west) -- (a-2-2.north east);
\draw[thick,densely dashed,blue!20] (a-3-1.north west) -- (a-3-2.north east);
\end{tikzpicture}
\caption{Examples, power and detection analysis in multi-mode pdf using a single sample realization}
\label{fig:powertestmulti}
\end{figure*}

With the same parameter configuration than previous section \ref{sec:powermulti} we computed the same measures using only a single realization of the sample. The power and detection analysis with increasing complexity is shown in figure \ref{fig:powertestmulti}. The results are in line with those shown in the previous section, where we highlighted the ability of the proposed CUBV method to adequately control FPs.

\subsection{Results on MRI data}

MRI datasets, described in section \ref{sec:datareal}, were evaluated with the proposed CV methods. Multiclass classification problems are usually divided into multiple, separate binary classifications. In this section we analyse three binary classification problems P1: HC+MCI vs AD+MCIc; P2: HC+MCIc vs MCI+AD; P3: HC+AD vs MCI+MCIc, that arise by combining the conditions studied in onset and progression of Alzheimer's Disease (HC, AD, MCI and MCIc). Cohen's distance $d$ for the set of problems can be seen in figure \ref{fig:cohend}. We ran $F=1000$ CV experiments and averaged the accuracy values obtained from K-fold, LOO and the proposed CV methods.

\subsubsection{\textbf{Real data analysis}}

In figure \ref{fig:cohend} we plotted the Cohen's distance for the analysed groups. Note that the null distribution is not properly modelled although $M=1000$ permutations were used to simulate the condition of no effect. Indeed there is a small effect hidden in data of about $0.1$. This can be explained by the presence of multiple clusters in each group and the similarities among them.

Data obtained after feature extraction is rather similar to that analysed in the previous simulated examples as shown in figure \ref{fig:MRI}. The presence of non-Gaussian distributions with multiple modes represents the typical scenario where standard CV methods are usually employed.

\subsubsection{\textbf{Mean classification variability and MC performance}}

\begin{figure*}
\centering
\begin{tikzpicture}
\matrix (a)[row sep=0mm, column sep=0mm, inner sep=1mm,  matrix of nodes] at (0,0) {
            \includegraphics[width=0.75\textwidth]{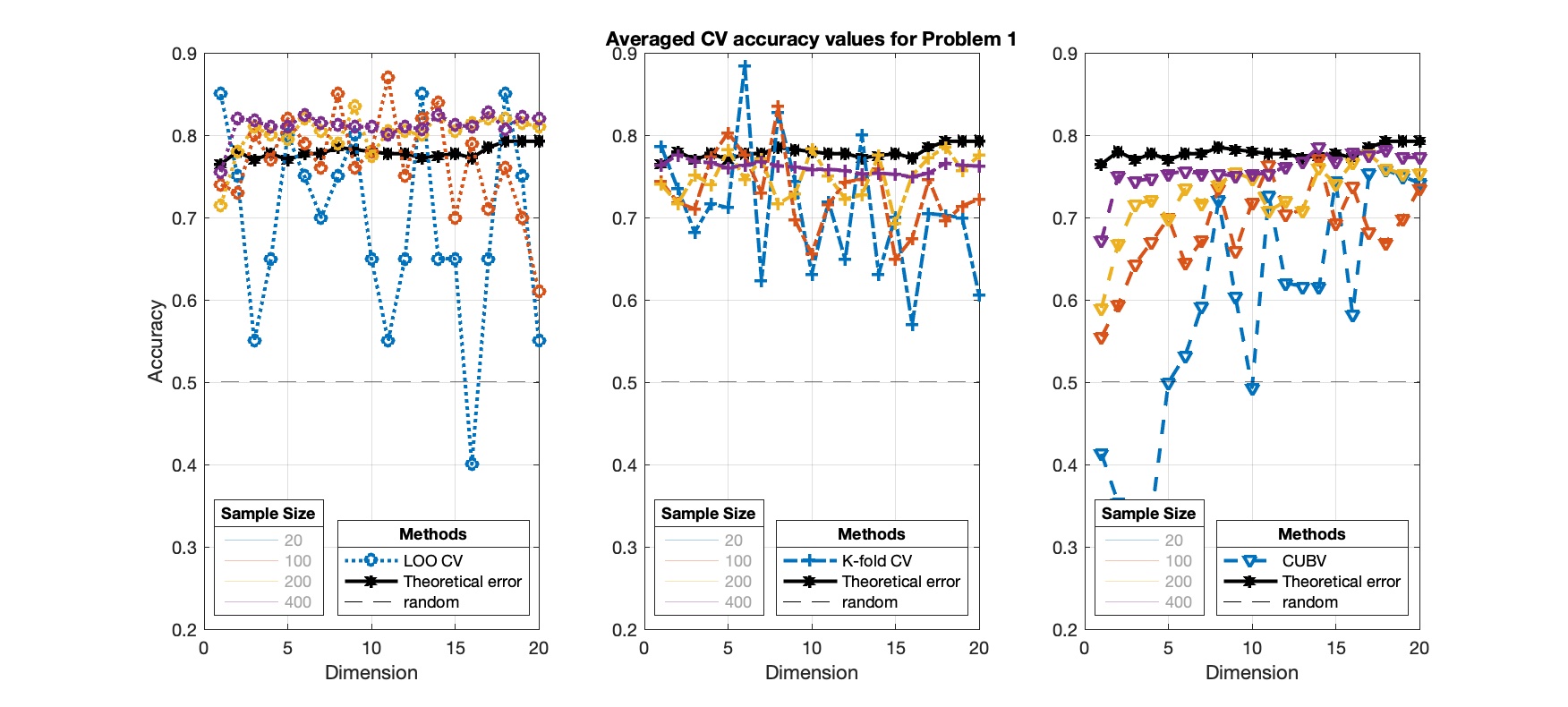}\\
            \includegraphics[width=0.75\textwidth]{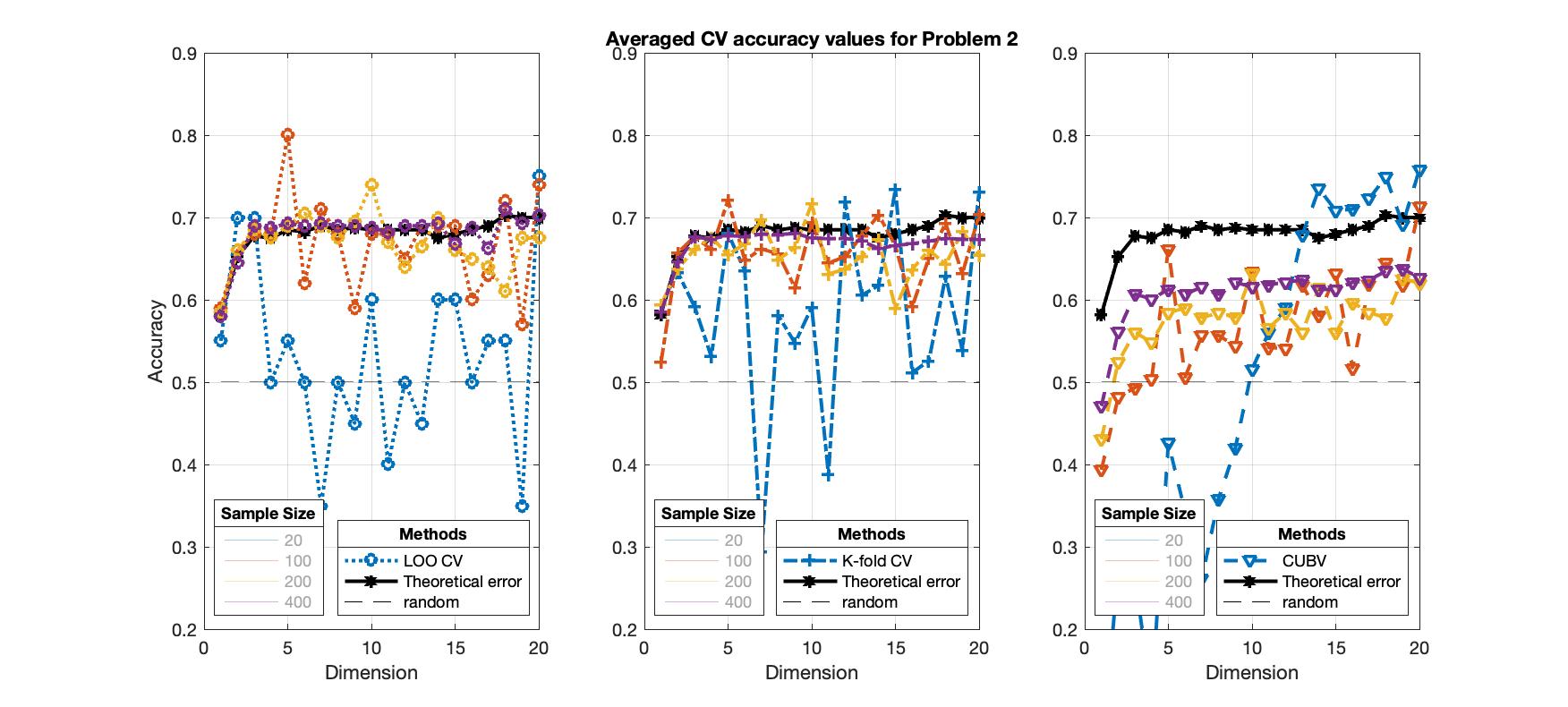}\\
            \includegraphics[width=0.75\textwidth]{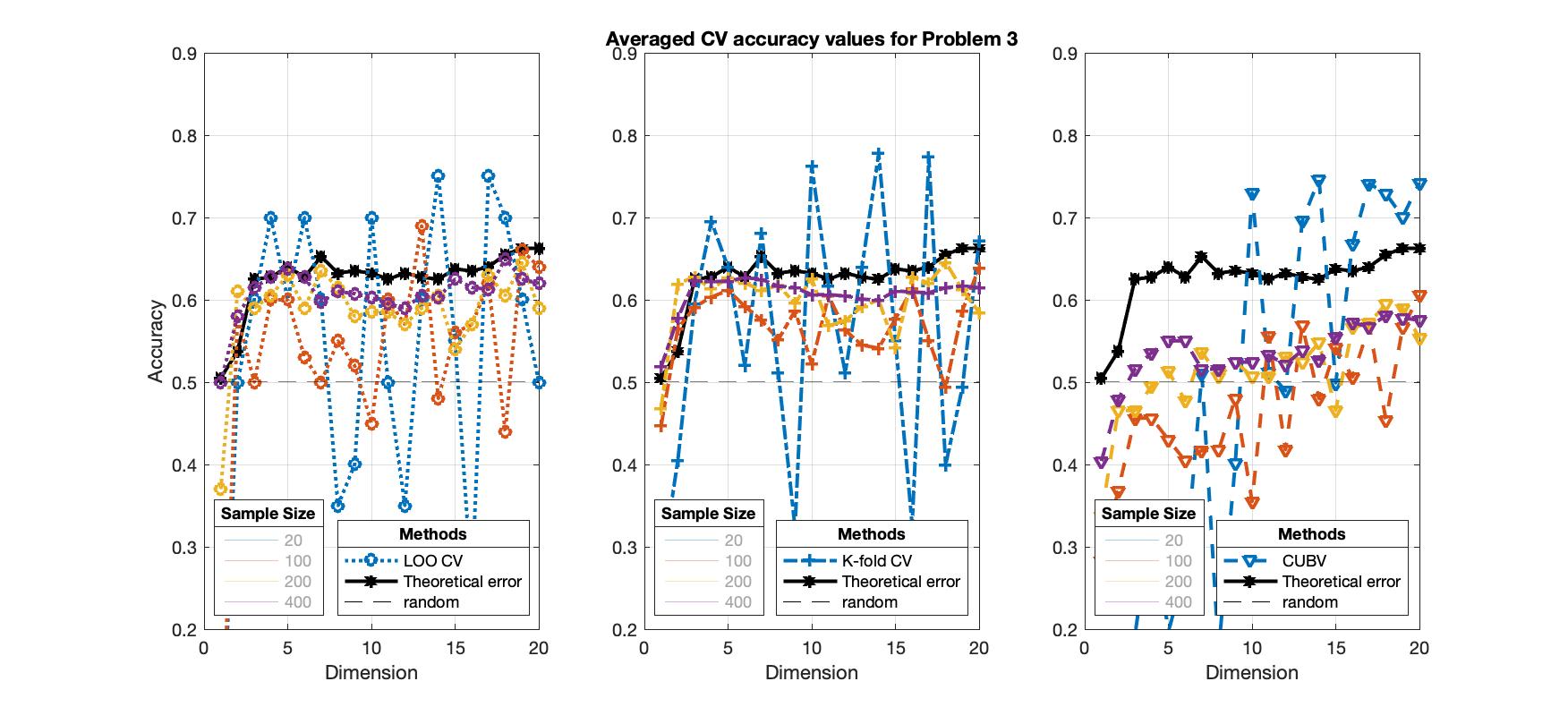}\\\\
        };
\draw[thick,blue!20] (a-2-1.north west) -- (a-2-1.north east);
\draw[thick,blue!20] (a-3-1.north west) -- (a-3-1.north east);
\end{tikzpicture}
\caption{Accuracy values for selected CV methods as a function of number of dimensions. Note the CUBV technique alerts on the variability of the results obtained by LOO and K-fold CV, especially in small effect and sample sizes, and low dimensions. With increasing sample size estimations, classical methods converge to the theoretical error rate (obtained by resubstitution of the whole dataset in the same manner as with simulated data).}
\label{fig:MRI2}
\end{figure*}

In figure \ref{fig:MRI2} we show the variability in the mean accuracy values obtained by LOO and K-fold CV ($F=1000$) that are compared with a threshold of $0.5$. The CUBV method shows an almost monotonic behaviour converging to the theoretical error value with increasing sample size and dimension. On the other hand, in figure \ref{fig:MRI2bis} the MC analysis is shown for both approaches where it can readily be seen that a larger number of trials is needed to achieve detection using the K-fold CV method (limited to 40 dB for visualization purposes and above $M=N$ at all times) and the increased probability of detection using the proposed approach with increasing dimension and sample size.

\subsubsection{\textbf{Power Analysis}}

\begin{figure*}
\centering
\begin{tikzpicture}
\matrix (a)[row sep=0mm, column sep=0mm, inner sep=1mm,  matrix of nodes] at (0,0) {
\includegraphics[width=0.75\textwidth]{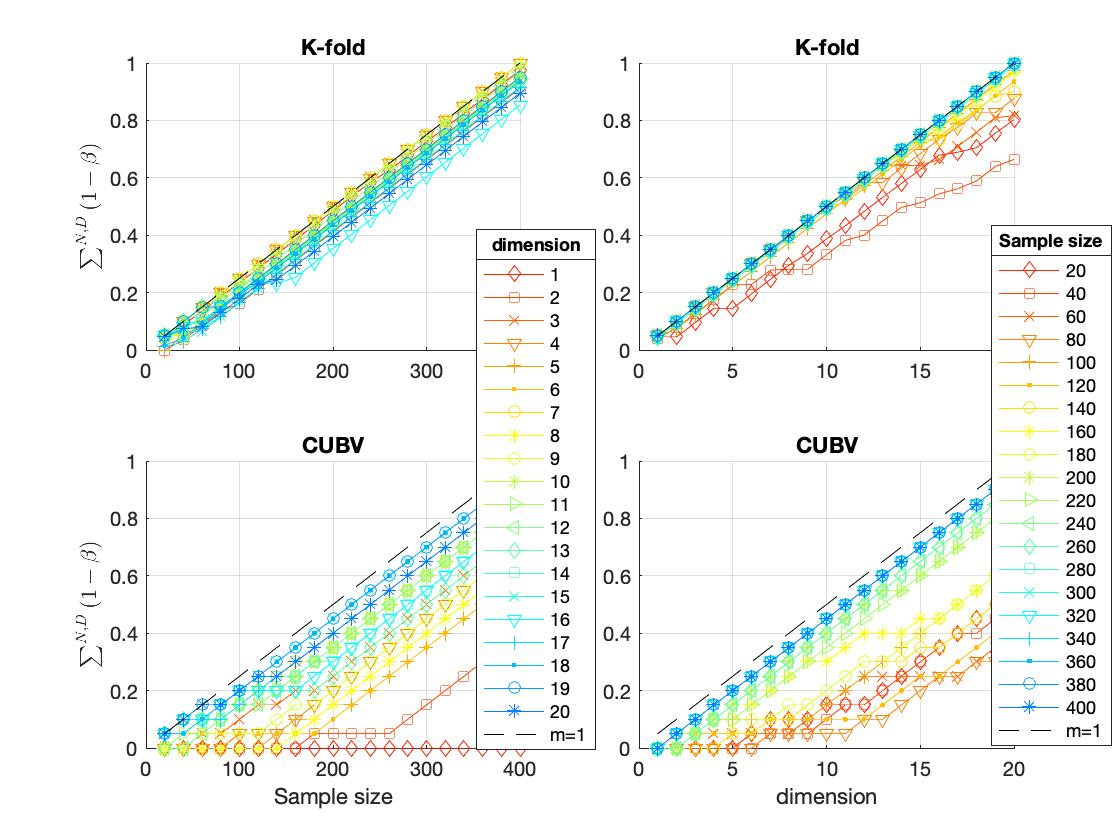}\\
\includegraphics[width=0.75\textwidth]{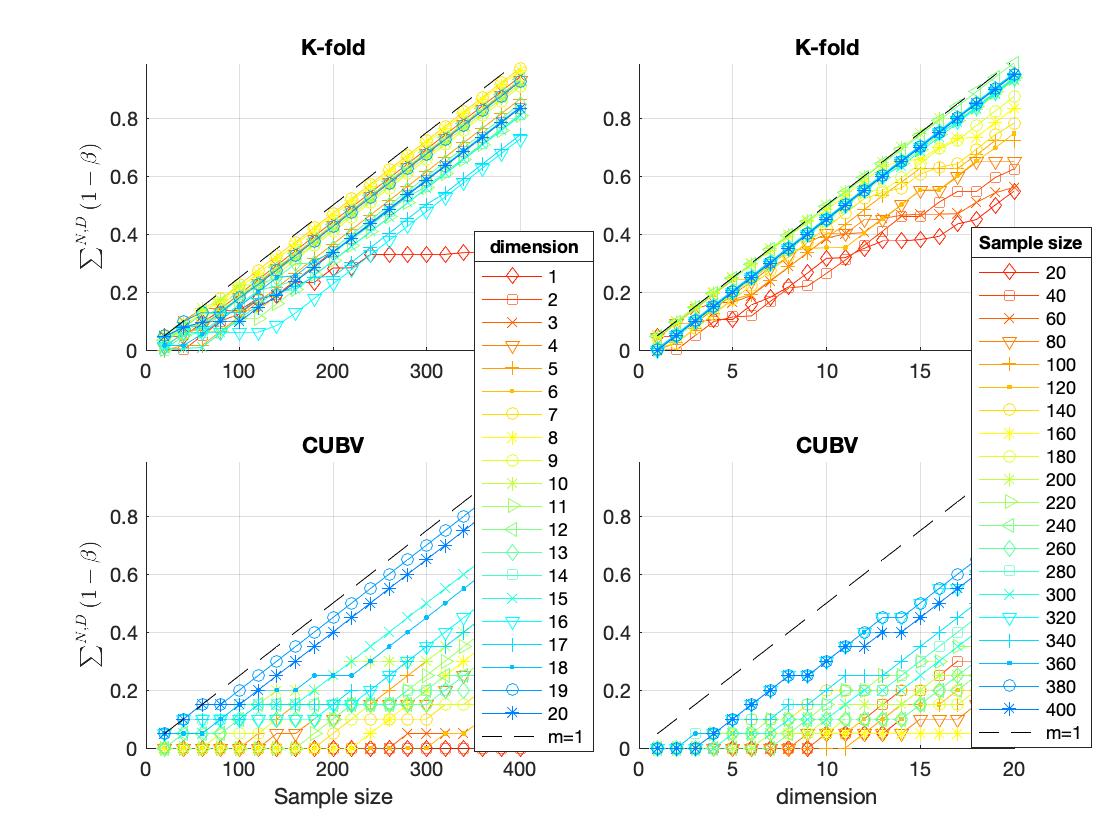}\\\\
        };
\draw[thick,blue!20] (a-2-1.north west) -- (a-2-1.north east);
\end{tikzpicture}
\caption{Normalized cumulative sum of ($1-\beta$) values for $N$ and $n$ versus dimension/sample size, respectively. Top: Problem 2; bottom: Problem 3.}
\label{fig:MRI4}
\end{figure*}

Finally, we plotted in figures \ref{fig:MRI3} and \ref{fig:MRI4} the normalized cumulative sum of beta values with increasing dimension and sample size for all the analyzed problems (P1-3), and the null experiment. Any deviation from the slope equal to 1 line in the K-fold CV means a detection with power less than 1, unlike CUBV that detects an effect when accuracy exceeds $0.5$. Observe in the null experiment that K-fold CV is mainly operating above the significance level ($\alpha=0.05$) with increasing sample size and dimension. In the latter experiment, CUBV only performs weakly at specific values of small sample sizes and high dimensions. Note that the null distribution is incorrectly modelled across sample sizes.

\section{Discussion}

From the set of experiments undertaken here, we have demonstrated that the proposed CUBV method is effective for assessing the variability of accuracy values obtained by K-fold CV. Moreover, the evidence derived from these experiments suggests that CUBV is a robust approach to statistical inference.

Whenever accuracy values were associated with a large deviation between actual and empirical risks, the CUBV method did not generate significant results over that expected by  random chance; that is, the null hypothesis was not erroneously rejected. Under the same conditions, K-fold CV provided accuracies above and below the threshold for significant across all effects sizes where CUBV indicated that the unseen CV accuracy values, drawn from an unknown pdf, were expected to have alarming uncertainty. In this sense, CUBV is a trade-off between the proper control of false positives and the power to detect true effects of the K-fold CV test. 

Classical methods to perform power calculations and evaluate the required number of trials to achieve detection (the MC evaluation method) revealed that K-fold CV usually operates far below its theoretical performance. In other words, the MC method can be considered  an over-conservative method and novel complex AI approaches that leverage these methods to a great extent cannot not be rigorously validated within these frameworks.

\begin{figure*}[h!]
\centering
\begin{tikzpicture}
\matrix (a)[row sep=0mm, column sep=0mm, inner sep=1mm,  matrix of nodes] at (0,0) {
            \includegraphics[width=0.49\textwidth]{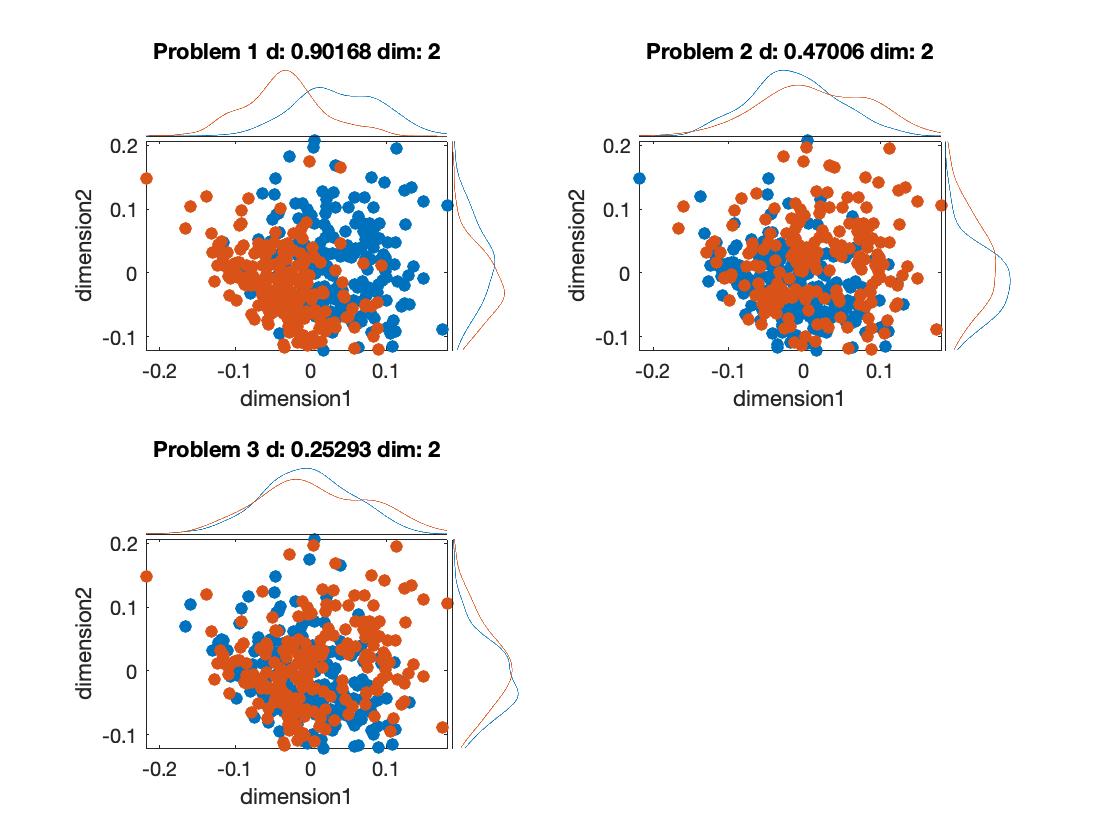}&
            \includegraphics[width=0.49\textwidth]{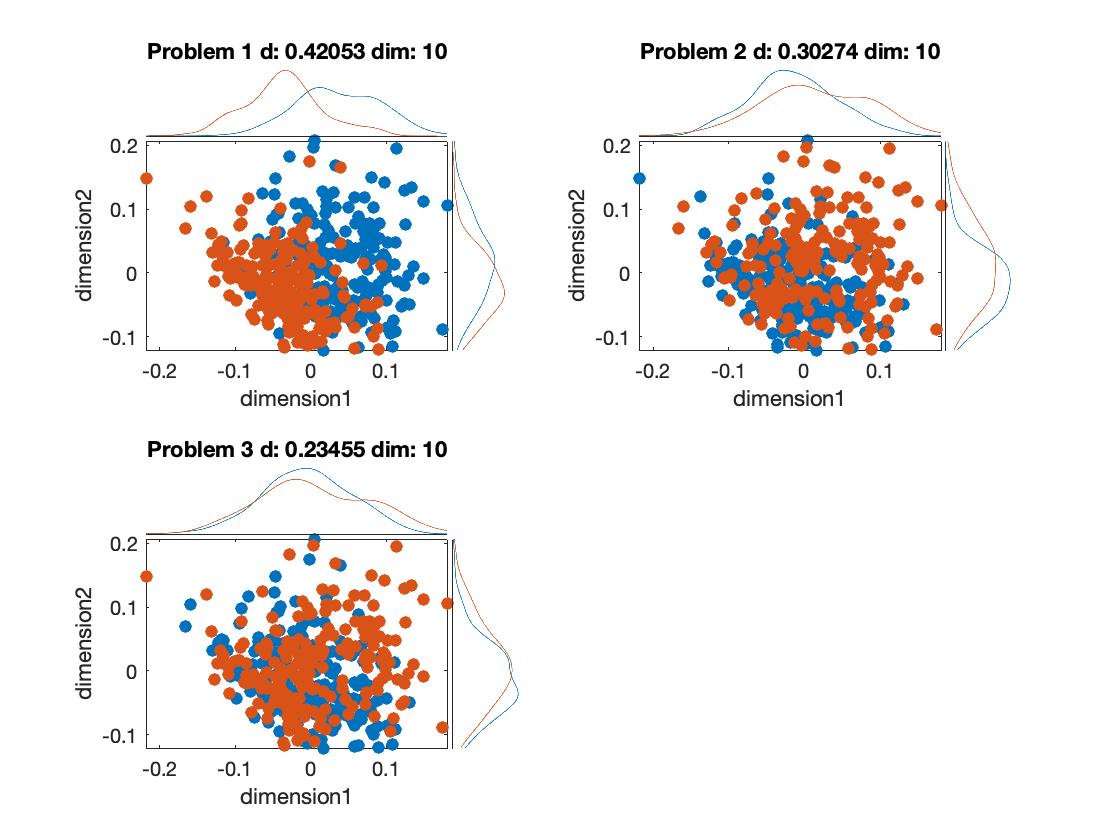}\\
            \includegraphics[width=0.49\textwidth]{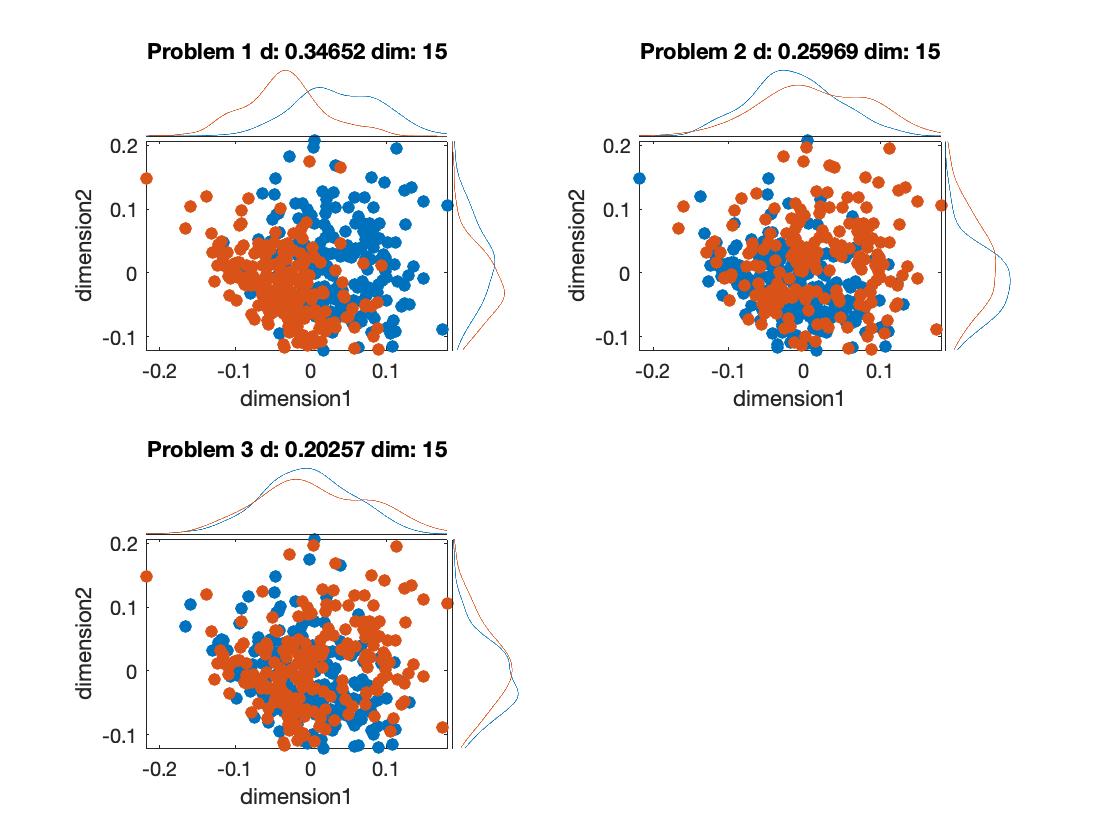}&
            \includegraphics[width=0.49\textwidth]{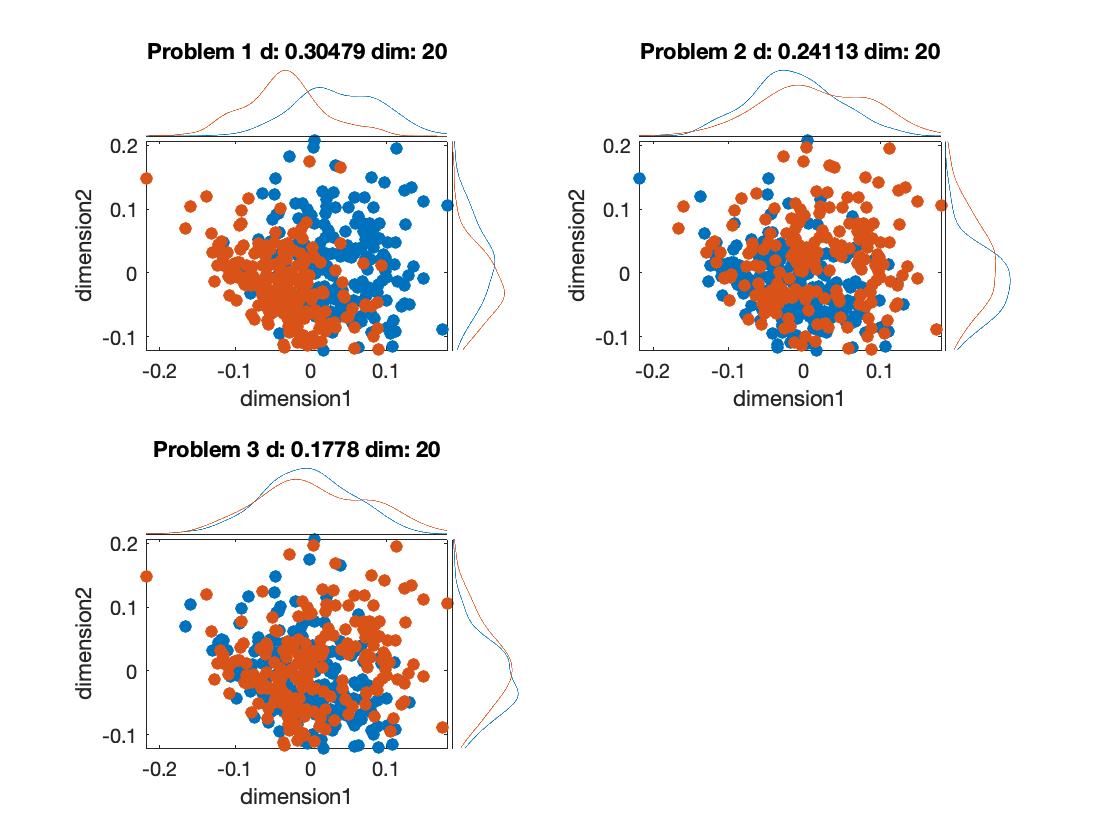}\\\\
        };
\draw[thick,blue!20] (a-1-1.north east) -- (a-2-1.south east);
\draw[thick,blue!20] (a-2-1.north west) -- (a-2-2.north east);
\end{tikzpicture}
\caption{Examples of data analyzed in this section with several dimensions and problems. Each classification problem corresponds with a different Cohen's $d$, whilst for dimension $n>2$ only the first two vector components are displayed. Note the similarities of the resulting samples, after feature extraction, with simulated data.}
\label{fig:MRI}
\end{figure*}

The theoretical findings are also applicable to the MRI-based samples from AD patients. The models devised in this paper together with the simulation of realistic datasets create suitable exemplars for  characterising performance in neuroimaging applications. A simple comparison between the datasets from figure \ref{fig:MRI} with those in figures \ref{fig:powertest2} and \ref{fig:example3} reveals the similarity in the results obtained. Nevertheless, the scatter plots and data distributions projected in the dimensions are clear examples demonstrating that the conditions to provide stable inducers are not met, and thus exploration of alternative validation methods is a priority.

\begin{figure*}[h!]
\centering
\includegraphics[width=0.75\textwidth]{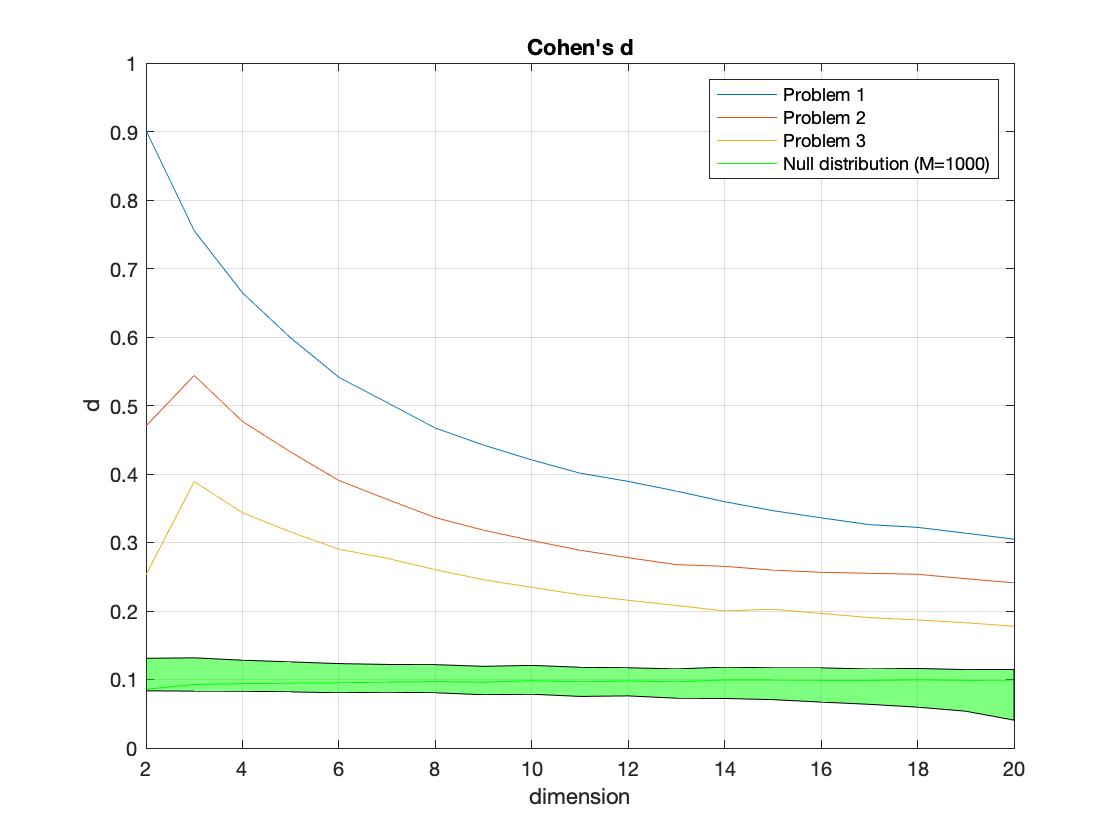}
\caption{Cohen's distance obtained from binary classes (whole datasets) versus dimension (PLS features). Note that Problem 1: HC+MCI vs AD+MCIc; Problem 2: HC+MCIc vs MCI+AD; Problem 3: HC+AD vs MCI+MCIc. The null distribution is also plotted and was obtained with $1000$ label permutations. Note that with this sample even the null distribution has a small (incorrectly modelled) effect.}
\label{fig:cohend}
\end{figure*}

\begin{figure*}
\centering
\includegraphics[width=\textwidth]{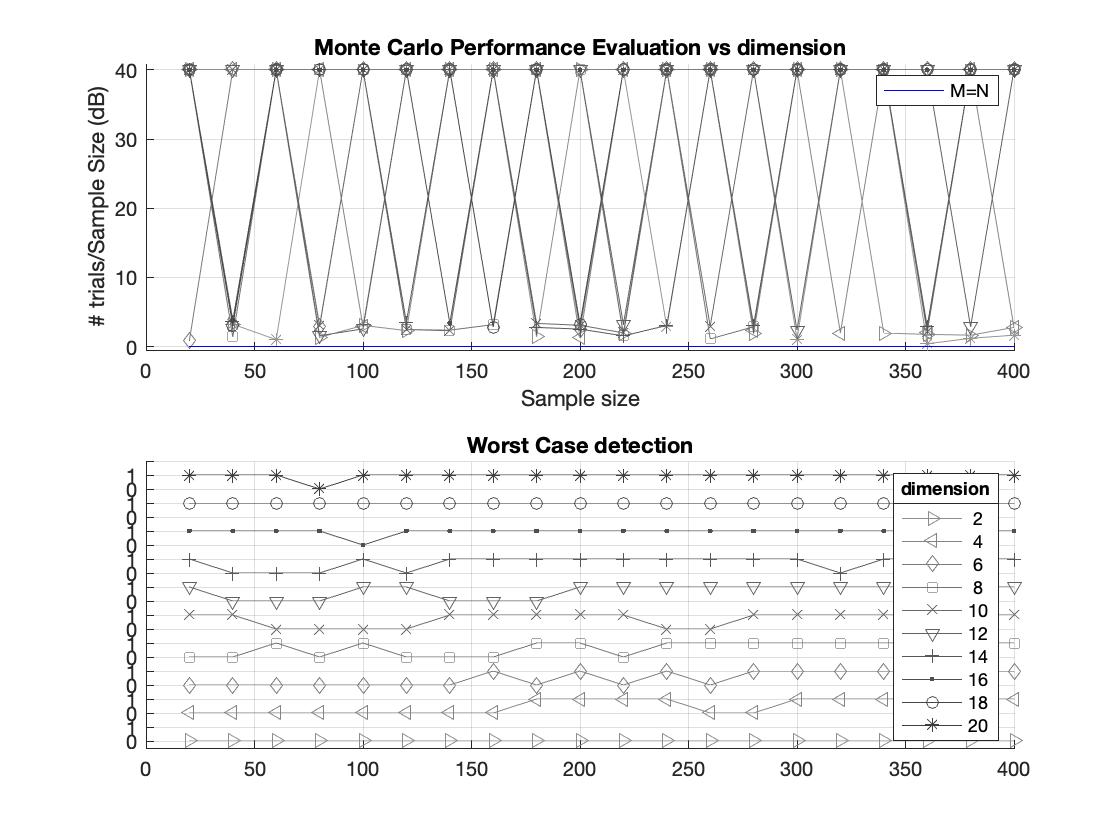}
\caption{Top: MC evaluation of K-fold CV in real datasets by averaging the results of Problems 1, 2 and 3. Note that values are limited to 40 db for visualization purposes. Bottom: detection analysis of the CUBV method versus sample size ($N$) and feature dimension ($n$).}
\label{fig:MRI2bis}
\end{figure*}

\begin{figure*}
\centering
\begin{tikzpicture}
\matrix (a)[row sep=0mm, column sep=0mm, inner sep=1mm,  matrix of nodes] at (0,0) {
\includegraphics[width=0.75\textwidth]{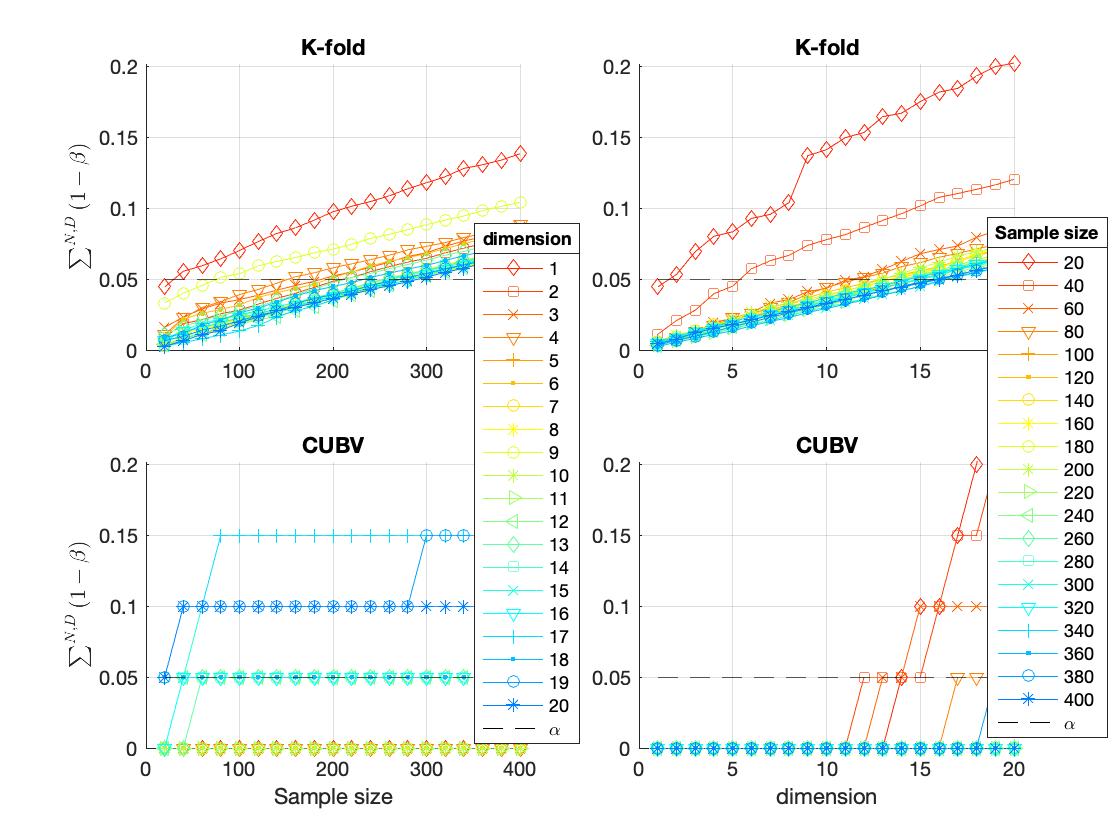}\\
\includegraphics[width=0.75\textwidth]{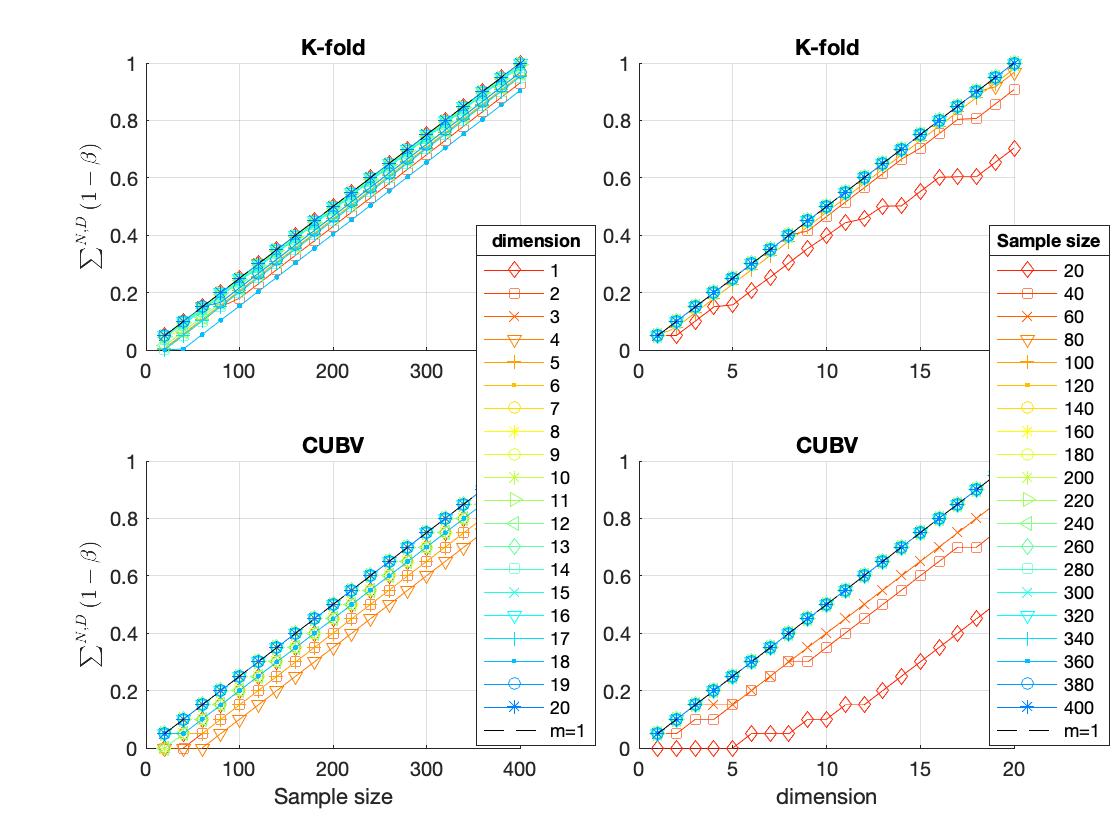}\\\\
        };
\draw[thick,blue!20] (a-2-1.north west) -- (a-2-1.north east);
\end{tikzpicture}
\caption{Normalized cumulative sum of ($1-\beta$) values ($P_c:=\frac{\sum_{j=1}^{(N,n)} (1-\beta_j)}{\#\text{experiments}}$) for $N$ and $n$ versus dimension/sample size, respectively. Top: null experiment; bottom: Problem 1}
\label{fig:MRI3}
\end{figure*}

At the final stage of any (image) ML-based processing system classifiers learn from folds of the limited amount of complex and multimodal samples. The conservative nature of the CUBV method results in robust detection that shows a monotonic behavior with sample size and feature dimension; see figure \ref{fig:MRI2bis}. The detection ability of K-fold CV methods in the search for real effects is arguable in the light of false positives from the null-experiment. As shown in figure \ref{fig:MRI3} their normalized cumulative sum of power shows a linear dependence on sample size and feature dimension. Thus, increasing sample size does not  control of the risk of false positive findings, in fact quite the opposite. One of the reasons for the poor performance is the difficulty in modelling the null distribution, as shown in figure \ref{fig:cohend}. Even with $F=1000$ label permutations and given the ``single-point'' $N=400$ sample dataset, there is an effect of about $0.1$ hidden in data that provokes a flawed statistical analysis. It is worth mentioning that, with increasing complexity of simulated data or when the problem to be solved is challenging, e.g. P2 and P3 using the MRI dataset, the CUBV method is even more useful for controlling the FP rate as shown in the experimental part. 

Finally, we emphasize the relevance of highlighting negative results in order to improve science\footnote{Please see the column in Nature about this issue https://www.nature.com/articles/d41586-019-02960-3}. Positive results are often the main goal of any research paper and we seldom evaluate our algorithms on putative task designs with no effect. This is the case described in section \ref{sec:null}. This analysis is important because it is usually employed to approximately model the null-distribution of the test-statistic in permutation analyses, e.g. performance or accuracy in a classification task using ML techniques. In permutation analysis the performance obtained from the paired data and labels is compared to that obtained by randomly permuting the group labels a large number of times, and should be distributed around $50\%$. If the distribution of the performance is non-symmetric around random chance, and is then biased, the result derived from the test data is likely to be flawed. This would mean that the distribution of data differs between groups under the null hypothesis, which violates the i.i.d. assumption and the estimation of p-values could lead to incorrect conclusions at the family-wise level \cite{Phipson2010}.

\section{Conclusions}

Standard CV methods in combination with ML were evaluated to ascertain whether statistical inferences made by data-driven approaches are sufficiently consistent. They are frequently claimed to outperform conventional statistical approaches such as hypothesis testing. However, this improvement is based on measures (e.g. accuracy) derived from ML classification tasks that do not have a parametric description and depend on the experimental setup; i.e. derived from classification/prediction folds of the dataset to establish confidence intervals. 

As shown in this paper, small sample-size datasets and learning from heterogeneous data sources strongly influences their performance and results in poor replication. A novel statistical test based on K-fold CV and the Upper Bound of the actual risk (K-fold CUBV) was proposed to tackle the uncertain predictions of ML and CV methods. The analysis of the \emph{worst case} obtained by a (PAC)-Bayesian upper bound for linear classifiers in combination with the K-fold CV estimation is a robust criterion to detect effects validating accuracy values obtained from ML models and avoiding false positives, complementing the regular K-fold method for CV.

\section*{Acknowledgments}
This work was supported by the MCIN/ AEI/10.13039/501100011033/ and FEDER ``Una manera de hacer Europa''  under the RTI2018-098913-B100 project, by the Consejer{\'i}a de Econom{\'i}a, Innovaci{\'o}n, Ciencia y Empleo (Junta de Andaluc{\'i}a) and FEDER under CV20-45250, A-TIC-080-UGR18, B-TIC-586-UGR20 and P20-00525 projects. We would also like to thank the reviewers for their contributions to improving this manuscript.

\bibliographystyle{srt}

\clearpage

\section*{Supplementary Materials}

\subsection{Remarks on ´´Common Experimental Designs'' section}\label{sec:preliminarybis}

How and when does a specific laboratory reject the null hypothesis that there is an effect? In order to test statistical significance a laboratory compares its observed accuracy with the set of values obtained by permuting conditions on its particular sample realization with the aim of modelling the null distribution. Is this sufficiently accurate? What if the sample realization is randomly providing accuracies from an overfitted solution? Even though the null distribution is properly modelled, the effect is being tested by \emph{assuming the alternative hypothesis} in a classification task, thus overfitting could affect the outcome measure. This is the main problem of using ML algorithms in group comparisons; one can never be assured of the statistical significance of the averaged accuracy values obtained from folds using empirical measures only. What if the null distribution is not properly modelled by a particular realization, e.g. by an insufficient number of permutations, $M$? This situation compounds overfitting and non-reliable conclusions could be drawn from the experiment. What if the patterns are more complex than those drawn from one-single-mode Gaussian pdf? In this case ML could be providing good classification results, distinguishing sub-patterns within the same condition in imbalanced datasets. This question is related to the assumptions, e.g. the Gaussianity and homogeneity assumptions, that are frequently violated in random effect analyses where the mixing proportion of the effect should be considered instead \cite{Rosenblatt16}.

\subsection{Nested CV intervals in low-dimensional classification and small sample sizes}\label{sec:nestedapprndix}

\begin{figure*}
\centering
\begin{tikzpicture}
\matrix (a)[row sep=0mm, column sep=0mm, inner sep=1mm,  matrix of nodes] at (0,0) {
            \includegraphics[width=0.49\textwidth]{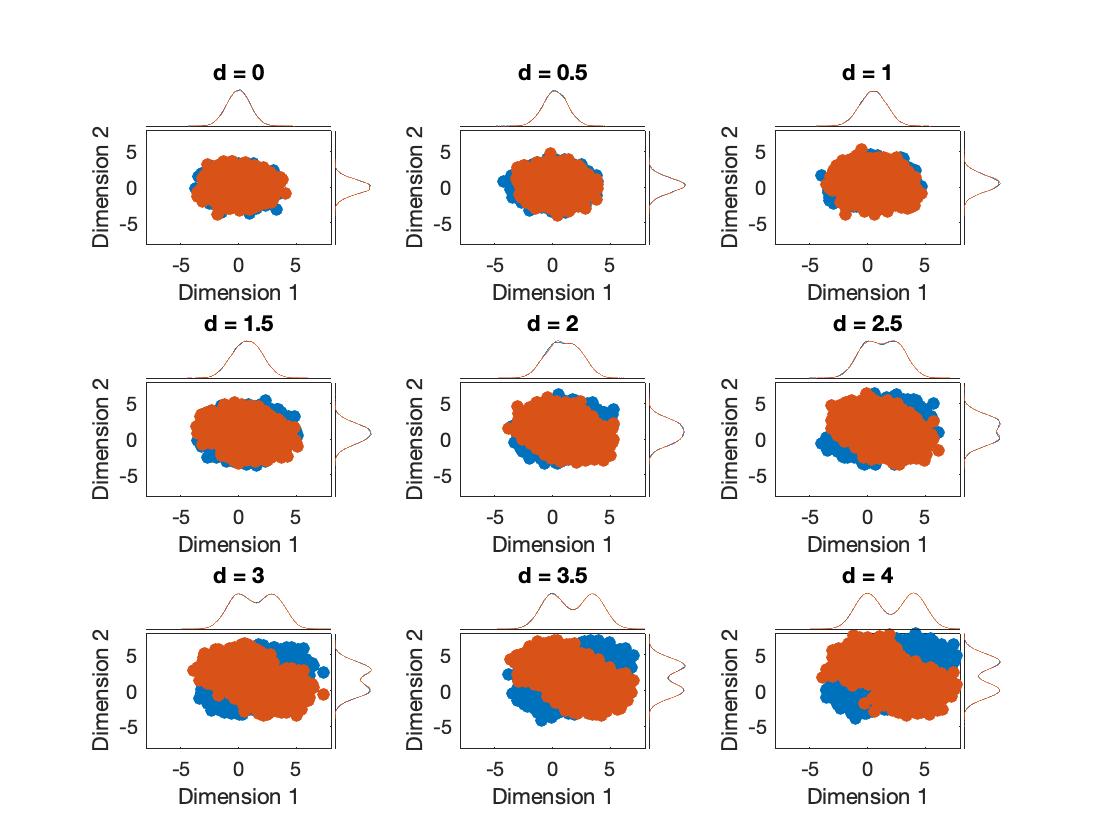} &
            \includegraphics[width=0.49\textwidth]{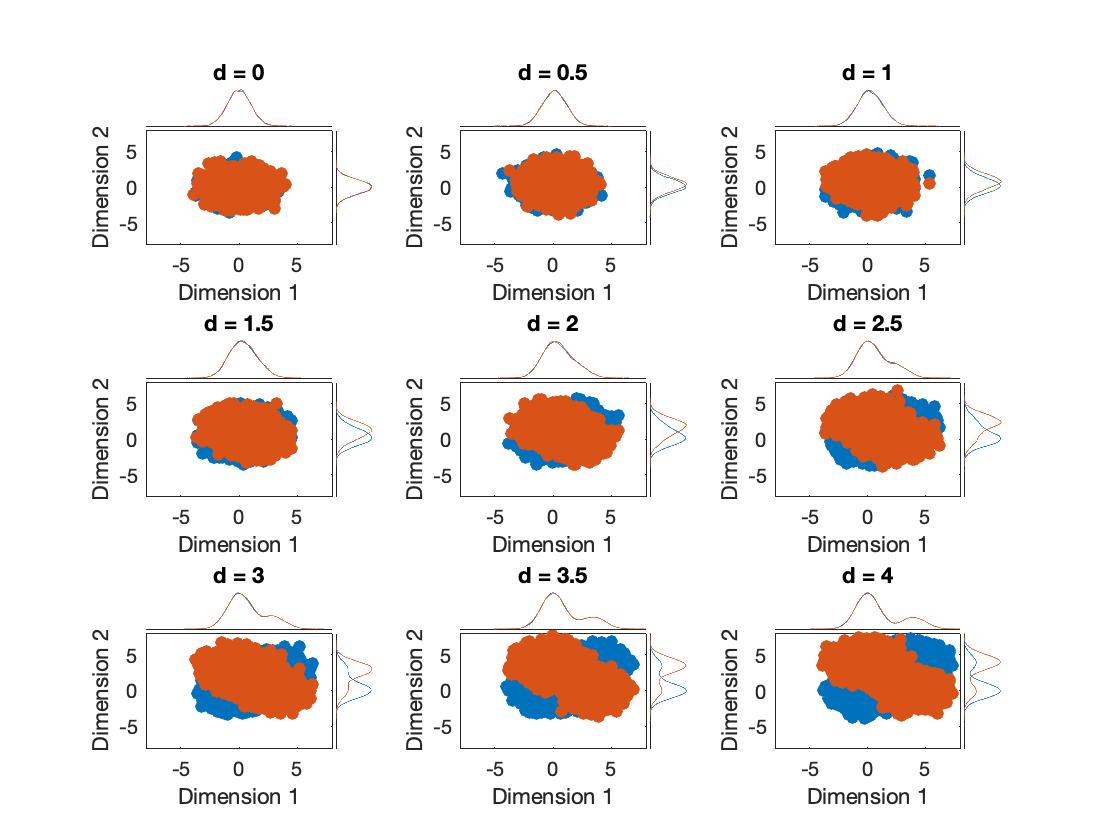} \\
            \includegraphics[width=0.49\textwidth]{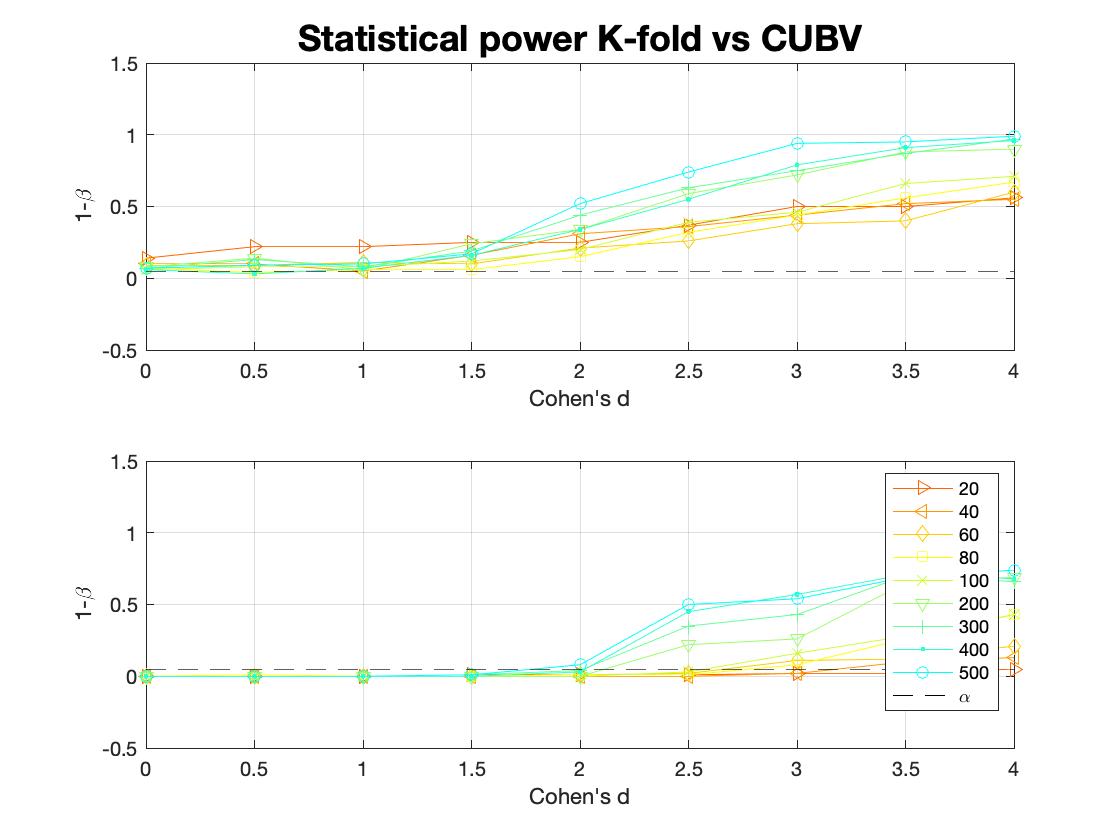} &
            \includegraphics[width=0.49\textwidth]{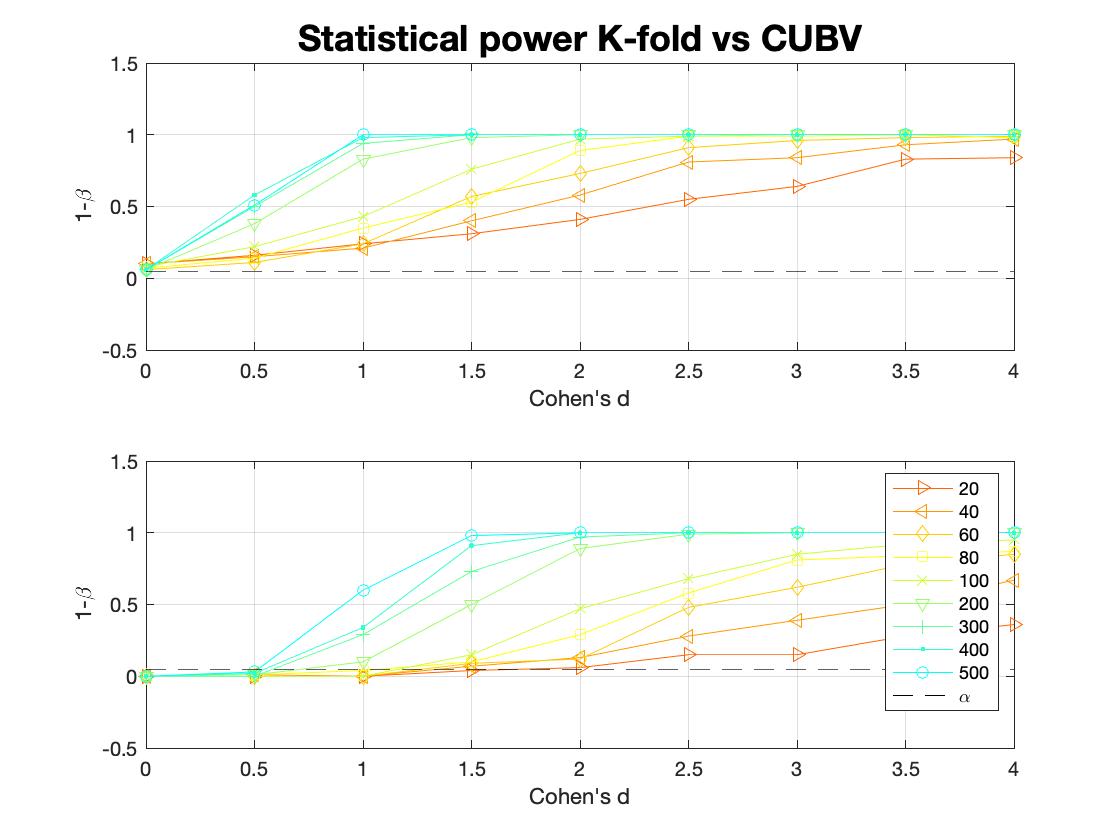} \\
            \includegraphics[width=0.49\textwidth]{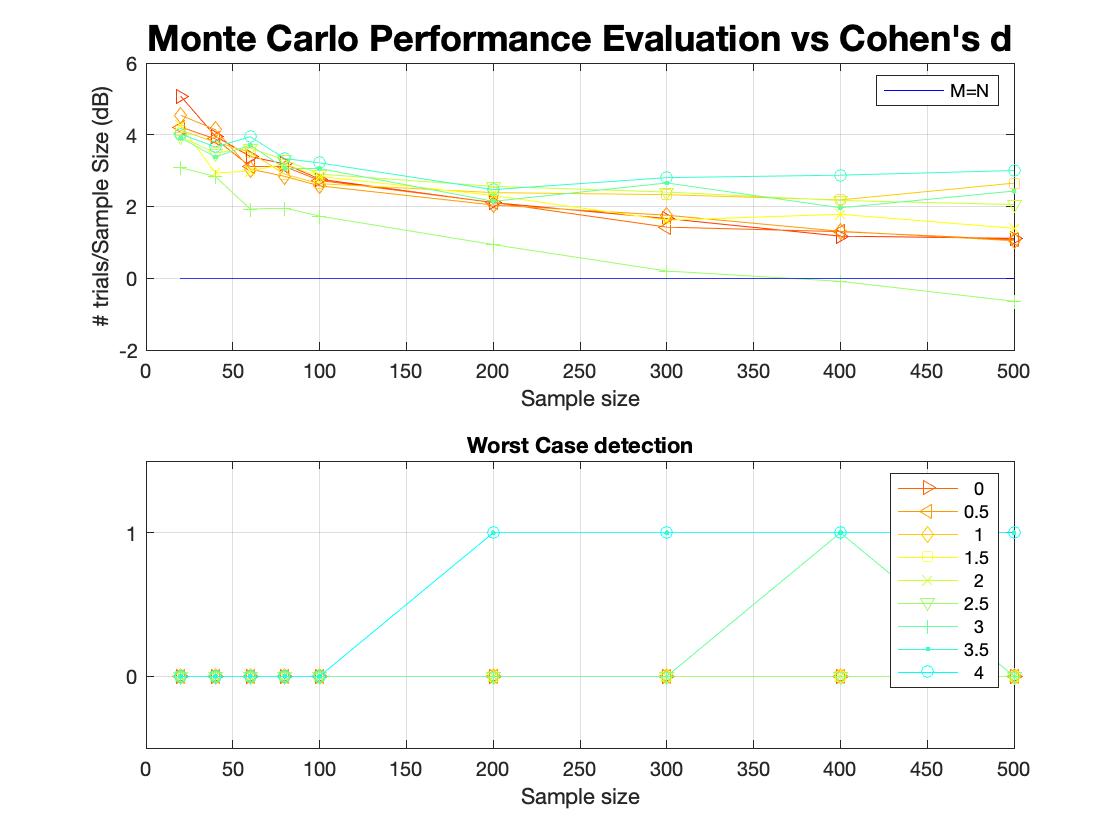} &
            \includegraphics[width=0.49\textwidth]{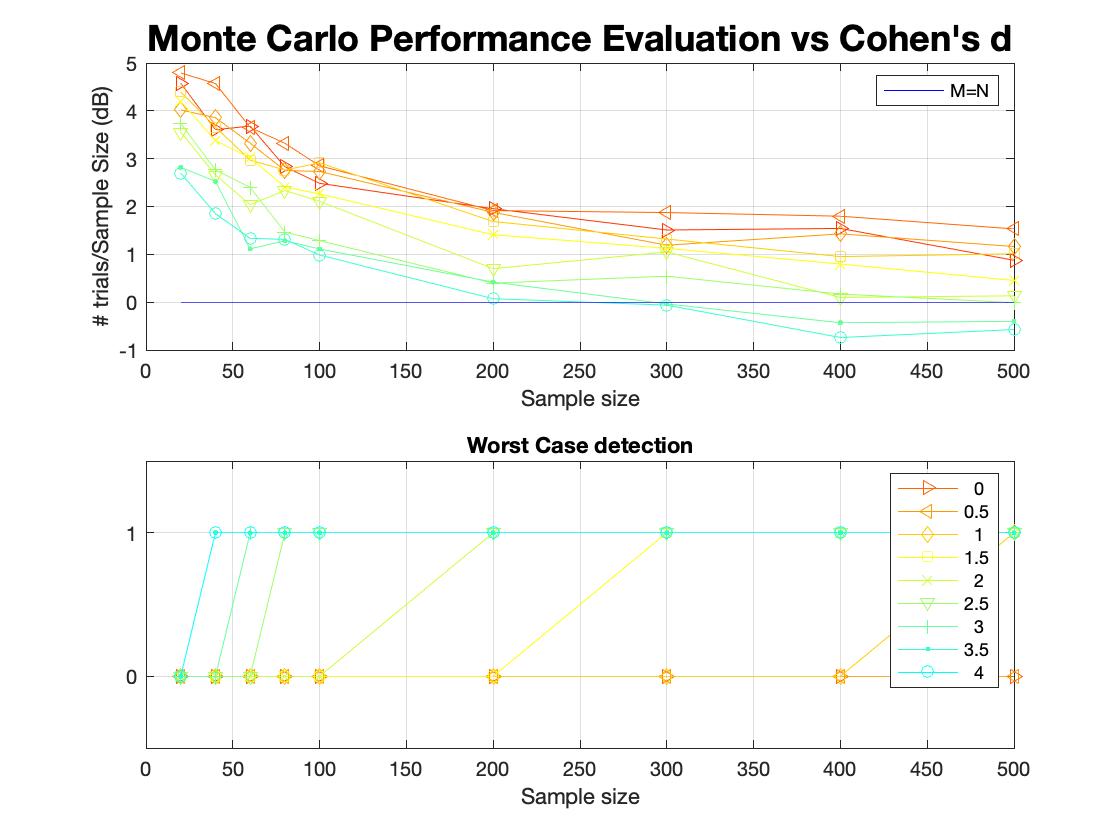}\\\\
        };
\draw[thick,blue!20] (a-1-1.north east) -- (a-3-1.south east);
\draw[thick,densely dashed,blue!20] (a-2-1.north west) -- (a-2-2.north east);
\draw[thick,densely dashed,blue!20] (a-3-1.north west) -- (a-3-2.north east);
\end{tikzpicture}
\caption{Analysis of the non-ideal case ($N_c=4$ and $n=2$). Left column: samples and $d$ values -top-; statistical power of CV permutation tests - middle up K-fold CV, middle-down CUBV; MC performance of K-fold CV -bottom up- and CUBV detection -bottom down- using a balanced dataset. Right column: the same measures using an imbalanced sample with $r=1/3$ per cluster in each group.}
\label{fig:powertest2}
\end{figure*}

We implemented the nested CV algorithm as described in \cite{Bates2023} and used it within our experimental framework to test its ability to manage low-dimensional classification problems with limited and complex samples. We compared it with naive CV, our approach based on SLT, and the theoretical upper and lower bounds \footnote{The 95\% confidence interval for a proportion is given by:
\[
\hat{P}(e) \pm Z_{\alpha/2} \sqrt{\frac{\hat{P}(e)(1 - \hat{P}(e))}{N}}
\]
where \(\hat{P}(e)\) is the estimated proportion of misclassifications, \(Z_{\alpha/2}\) is the critical value from the standard normal distribution for a given confidence level (for 95\%, \(Z_{\alpha/2} \approx 1.96\)), and \(N\) is the total number of observations.
}. First, we observed the behavior noted in \cite{Bates2023}, namely that nested CV had superior coverage compared to naive CV confidence intervals (the latter are always narrower than the model-driven theoretical bounds). However, we also observed two undesirable effects with nested CV: variability in the confidence intervals and overly conservative performance with small sample sizes. Additionally, nested CV is more computationally intensive than the other methods. Second, on average, for several effect sizes, our CUBV approach aligns with the lower limit of nested CV, thus validating both methods for establishing protective confidence intervals of the error. We observed the same behavior using the data from the multi-mode power analysis in Figure \ref{fig:powertest2}. Given the similar average performance of both naive and nested CV approaches and the ubiquitous use of naive CV, we prefer to focus on the naive CV approach for further analyses.

\subsection{Modelling data heterogeneity with SLT}

Heterogeneity in data is not always proportional to the number of (unknown) sources generating samples when, for example, they 'correlate' within a class. The concept of data-(in)dependent measures of complexity of sets of functions derived by SLT can be evaluated for this purpose \cite{Boucheron13}. In general, capacity measures, e.g., the Rademacher average, establish a connection between empirical errors (derived from the training set) and the unobserved generalization error, i.e., the actual risk. One of these measures is the well-known Vapnik-Chervonenkis (VC) dimension \cite{Vapnik82}, which we use in the experimental section to assess the degree of sample heterogeneity.

\begin{definition}\label{def:1}
A set of N vectors $Z_N = \{z_1, \ldots, z_N\}$ is in general position in n-space $Z$ if every subset of n or fewer vectors is linearly independent.
\end{definition}

\begin{definition}\label{def:2}
Sets Shattered by a Class of Functions: Let $Z$ be a set and $\mathcal{F}$ be a class of functions where $\mathcal{F}: Z \rightarrow \{0,1\}$. A subset $Z_N = \{z_1, \ldots, z_N\} \subseteq Z$ in general position is said to be \textbf{shattered} by the class of functions $\mathcal{F}$ if, for every possible binary labeling of the elements in $Z_N$, there exists a function $f \in \mathcal{F}$ that correctly assigns those labels. Formally, $S$ is shattered by $\mathcal{F}$ if:
\[
\left\{ (\mathbf{1}_{f(z_1)}, \ldots, \mathbf{1}_{f(z_N)}) : f \in \mathcal{F} \right\} = \{0,1\}^N
\]
where $\mathbf{1}_{.}$ denotes the indicator function.
\end{definition}

In other words, a subset $Z_N$ is shattered by the class of functions $\mathcal{F}$ if the class can realize all $2^N$ possible binary labelings of the $N$ elements in $Z_N$.

\begin{definition} \label{def:3}
The VC dimension of $\mathcal{F}$ is defined as:
\[
h(\mathcal{F}):=\max\{N\in\mathbb{N}:Z_N\subseteq Z\}
\]
where $Z_N$ is shattered by $\mathcal{F}$.
\end{definition}

\begin{figure*}[ht!]
\centering
\includegraphics[width=0.5\textwidth]{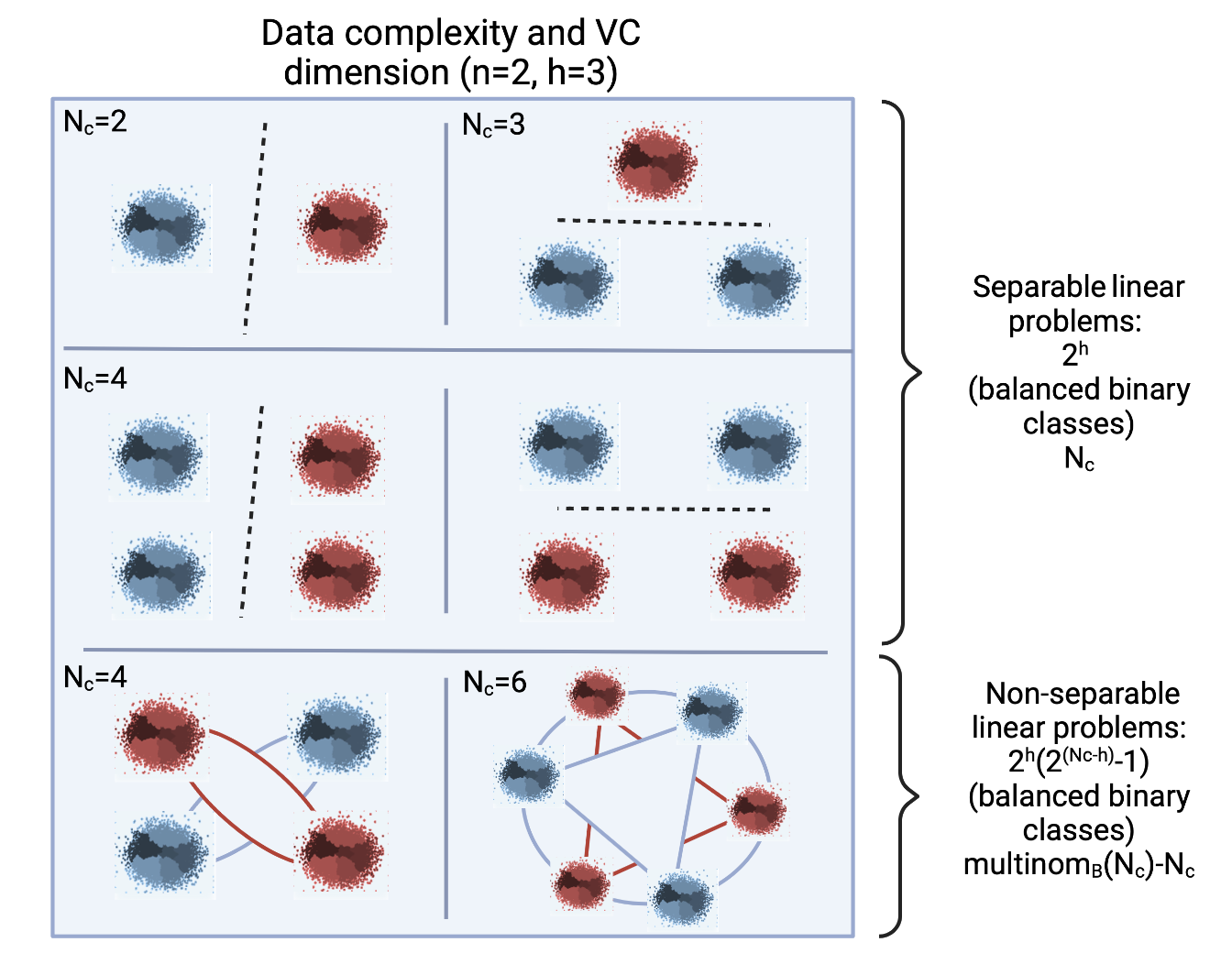}
\caption{Data complexity and VC dimension in $n=2$. In two dimensions the number of non-intersecting convex hulls is, in general, $2^h$ for a set of points or distant clusters with cardinality less than the Radon number ($n+2$) \cite{Tverberg66}. Assuming balanced sources we have in 2D only up to $\binom{h}{h/2,h/2}\sim N_c=6$ separable simulations whilst the number of non-separable simulations grow with order $\sim  \frac{2^{N_c+1}}{\sqrt{2\pi N_c}}$}
\label{fig:example3bis}
\end{figure*}

We restrict our study of data heterogeneity, without loss of generality, to the class of functions $\mathcal{F}$ consisting of linear classifiers, e.g. $f(z|w,b)=w\cdot z\overset{>}{<}b$. This focus is motivated, as mentioned in the introduction, by the observation that many machine learning systems operate effectively in low-dimensional feature spaces where linear approximations are predominant. In a nutshell, the VC dimension of the class of linear classifiers is the size of the largest set that it can shatter. For example, in two dimensions we can shatter up to three samples using linear classifiers (see figure \ref{fig:example3bis}).

\begin{proposition}\label{prop:1}
Given a set of N+1 vectors $Z_N = \{z_1, \ldots, z_N, z_o\} \subseteq Z$ in general position in n-space then the N projections onto the $n-1$ dimensional orthogonal subspace to $z_o$ are in general position.

Proof: Consider the subspace \( W \) orthogonal to \( x_o \). This subspace \( W \) is a subspace of \( Z \) of dimension \( n-1 \), defined as:
   \[
   W = \{ \mathbf{z} \in Z \mid \mathbf{z} \cdot \mathbf{z_o} = 0 \}
   \]
   where \( \mathbf{z} \cdot \mathbf{z_o} \) denotes the dot product of \( \mathbf{z} \) and \( \mathbf{z_o} \).

The orthogonal projection of a vector \( z_i \) onto \( W \) is denoted by \( z_i' \) and is calculated as:\[
\mathbf{z}'_i = \mathbf{z}_i - \frac{(\mathbf{z}_i \cdot \mathbf{z}_o)}{(\mathbf{z}_o \cdot \mathbf{z}_o)} \mathbf{z}_o
\]
   
Consider \( k \) vectors from the projected set \( Z_N' \), where \( k \leq n-1 \). Suppose they are not independent, i.e., there exist scalars \( \alpha_1, \alpha_2, \ldots, \alpha_k \), not all zero, such that:
\[
\alpha_1 \mathbf{z_1'} + \cdots + \alpha_k \mathbf{z_k'} = 0
\]
Substituting the projection formula:
\[
\alpha_1 \left( \mathbf{z_1} - \frac{\mathbf{z_1} \cdot \mathbf{z_0}}{\mathbf{z_o} \cdot \mathbf{z_o}} \mathbf{z_o} \right) + \cdots + \alpha_k \left( \mathbf{z_k} - \frac{\mathbf{z_k} \cdot \mathbf{z_0}}{\mathbf{z_o} \cdot \mathbf{z_o}} \mathbf{z_o} \right) = 0
\]
Rearranging terms:
\[
\alpha_1 \mathbf{z_{1}} + \cdots + \alpha_k \mathbf{z_{k}} - \left( \alpha_1 \frac{\mathbf{z_{1}} \cdot \mathbf{z_o}}{\mathbf{z_o} \cdot \mathbf{z_o}} + \cdots + \alpha_k \frac{\mathbf{z_{k}} \cdot \mathbf{z_o}}{\mathbf{z_o} \cdot \mathbf{z_o}} \right) \mathbf{z_o} = 0
\]
This can be written as:
\[
\alpha_1 \mathbf{z_{1}}+\cdots + \alpha_k \mathbf{z_{k}} = \beta \mathbf{z_o}
\]
with \( \beta = \left( \alpha_1 \frac{\mathbf{z_{1}} \cdot \mathbf{z_o}}{\mathbf{z_o} \cdot \mathbf{z_o}} + \cdots + \alpha_k \frac{\mathbf{z_{k}} \cdot \mathbf{z_o}}{\mathbf{z_o} \cdot \mathbf{z_o}} \right) \). Then we have:
\[
\alpha_1 \mathbf{z_{1}} + \cdots + \alpha_k \mathbf{z_{k}} = \beta \mathbf{z_o}
\]
Since \( \mathbf{z_o}, \mathbf{z_{1}}, \ldots, \mathbf{z_{k}} \) are all linearly independent (because \( k < n \)), the only way this equation can hold is if \( \alpha_1 = \cdots = \alpha_k = 0 \). This contradicts our assumption that the \( \alpha_i \) were not all zero.
\end{proposition}

An important lemma based on Lemma 1 in \cite{Cover65} and rewritten for our purposes is the following: 
\begin{lemma}\label{prop:2} The Shattering extension.
Let $Z_N = \{\mathbf{z_1}, \ldots, \mathbf{z_N}\} \subseteq Z=\mathbb{R}^n$ and $\mathbf{z_o}$ a point in $Z$, then $\{Z\cup \mathbf{z_o}\}$ is shattered by $\mathcal{F}$ if and only if $Z_N$ is also shattered with $\mathbf{w}\cdot \mathbf{z_o}=b$.

Proof: A simple proof can be derived form Lemma 1 in \cite{Cover65} but extending to linearly separable cases. An extended dichotomy $\{Z^+,Z^-\}\cup z_o$, where $Z^+$ and $Z^-$ denote vectors with different labelings, is linearly separable if a only if there exists two functions $\{f_1(z|w_1,b_1),f_2(z|w_2,b_2))\}\in\mathcal{F}$ that provide the two possible labelings for $z_o$. From them define $f^*(z|w^*,b^*)=w^*\cdot z\overset{>}{<}b^*$, where $w^*=-(w_2\cdot z_o) w_1+(w_1\cdot z_o)w_2-b_2w_1+b_1w_2$ and $b^*=-w_2z_ob_1+w_1z_ob_2$, which has a zero in $z_o$ and separates the original dichotomy $\{Z^+,Z^-\}$. Conversely, if a specific dichotomy $\{Z^+,Z^-\}$ is linear separable by a function $f^*$ with $f^*(z_o)=b$ the functions $f^*\pm \epsilon z_o$ with $\epsilon>0$ separates the extension dichotomy in both labels. To extend this result to our proposition we apply this lemma to the complete set of dichotomies that are generated by the set of functions that shatter the extension set and, conversely, to the original set when $f^*(z_o)$ lies in the separation hyperplane.
\end{lemma}
The previous lemma will be used in the following proposition, which presents a completely novel demonstration of the complexity of linear classifiers without relying on theoretical concepts of capacity measures, such as growth functions, as described in other references \cite{Vapnik98}.

\begin{proposition} \label{prop:3}
Given \( Z = \mathbb{R}^n \), the class of linear classifiers \( \mathcal{F} \) has \( h(\mathcal{F}) = n+1 \). This means that \( \mathcal{F} \) can shatter subsets of $Z$ with cardinality up to $n+1$, but no larger. 

Proof: By induction, we start with \( n=1 \). In one dimension, a linear classifier is a threshold point \( b \) on \( \mathbb{R} \), e.g., \( f(z) = w \cdot z \overset{>}{<} b \). Each class, e.g., \( \{0,1\} \), can be modeled as a closed halfspace \( \{z \in \mathbb{R} : w \cdot z \overset{>}{<} b; w, b \in \mathbb{R}\} \). The classifier can shatter any set of 2 points \( \{z_1, z_2\} \) with \( z_1 < z_2 \) if it can assign any possible combination of labels to these 2 points. For example, this can be achieved by selecting \( w = \left\{\frac{b}{z_2 + \epsilon}, \frac{b}{z_1 - \epsilon}, \frac{b}{z_1 + \epsilon}, -\frac{b}{z_1 + \epsilon}\right\} \), where \( \epsilon < z_2 - z_1 \) is a small positive constant. To conclude the base case, consider three sorted points \( \{z_1 < z_2 < z_3\} \). We need to prove that there is a combination of labels that cannot be shattered. Assume by contradiction that there exist \( w \) and \( b \) that shatter the subset with labels \( \{1, 0, 1\} \). This would require satisfying the inequalities \( \{ wz_1 > b, wz_2 < b, wz_3 > b \} \), which is not possible as it contradicts the linearity imposed by a single threshold \( b \). Thus, \( h = 2 \) with \( n = 1 \).

Now assume a linear classifier in \( n \) dimensions, i.e., \( f(\mathbf{z}) = \mathbf{w} \cdot \mathbf{z} \overset{>}{<} b \), can shatter any set of \( n+1 \) points. Formally, let:
\[
\{\mathbf{z_1}, \mathbf{z_2}, \ldots, \mathbf{z_{n+1}}\} \subseteq \mathbb{R}^n
\]
Consider a set of \( n+2 \) points in \( n+1 \) dimensions:
\[
\{\mathbf{z_1}, \mathbf{z_2}, \ldots, \mathbf{z_{n+1}}, \mathbf{z_{o}}\} \subseteq \mathbb{R}^{n+1}
\]

We need to show that for any possible labeling of these points, there exists a hyperplane that correctly separates them according to their labels. Select one point from the set, say \( \mathbf{z_{o}} \), and consider the projection of the remaining $n+1$ points onto an \( n \)-dimensional subspace orthogonal to \( \mathbf{z_{o}} \). Applying the induction hypothesis to the projected points \( \{\mathbf{z}'_1, \mathbf{z}'_2, \ldots, \mathbf{z}'_{n+1}\} \subseteq \mathbb{R}^n \), these projections can be shattered in \( \mathbb{R}^n \). Now, we extend the shattering function in \( \mathbb{R}^n \) back to \( \mathbb{R}^{n+1} \) to include \( \mathbf{z_{o}} \):
\[
\mathbf{w} \cdot \mathbf{z_{o}} = b; \quad b = b', \quad \mathbf{w} = [\mathbf{w'}^T, \frac{b - \mathbf{w}' \cdot \mathbf{z_{o}}'}{z_{o}(n+1)}]^T
\]

These hyperplanes will separate \( \mathbf{z_{o}} \) from the rest of the points according to its label while maintaining the separation for the projection of \( \{\mathbf{z_1}, \mathbf{z_2}, \ldots, \mathbf{z_{n+1}}\} \) based on their labels (Lemma \ref{prop:2}). Moreover, any dichotomy of the remaining \( n+1 \) points in \( \mathbb{R}^{n+1} \) is separable by such hyperplanes since the set of projections is separable (see proof of theorem 3 in \cite{Cover65}). To conclude the proof, \( n+3 \) points cannot be shattered in \( \mathbb{R}^{n+1} \) because, by induction, the projection of any selection of \( n+2 \) points cannot be shattered in \( \mathbb{R}^n \).

\end{proposition}

\begin{figure*}
\centering
\begin{tikzpicture}
\matrix (a)[row sep=0mm, column sep=0mm, inner sep=1mm,  matrix of nodes] at (0,0) {
            \includegraphics[width=0.49\textwidth]{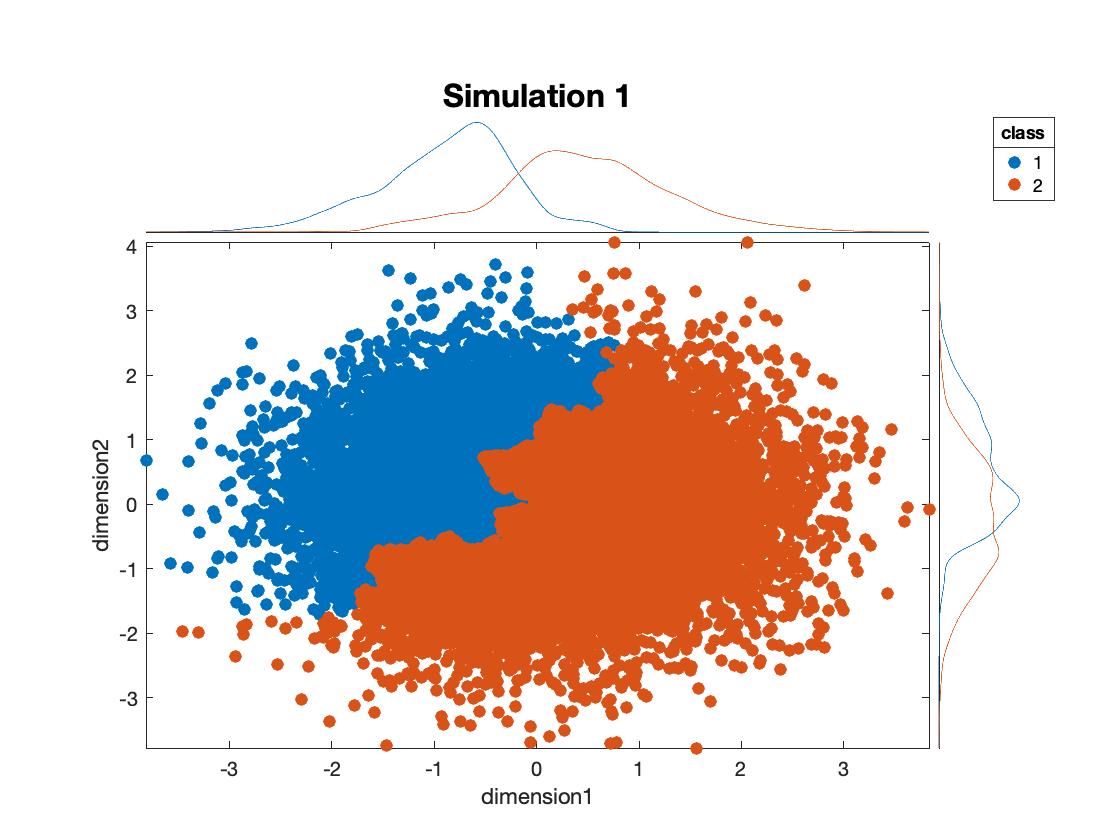}
            \includegraphics[width=0.49\textwidth]{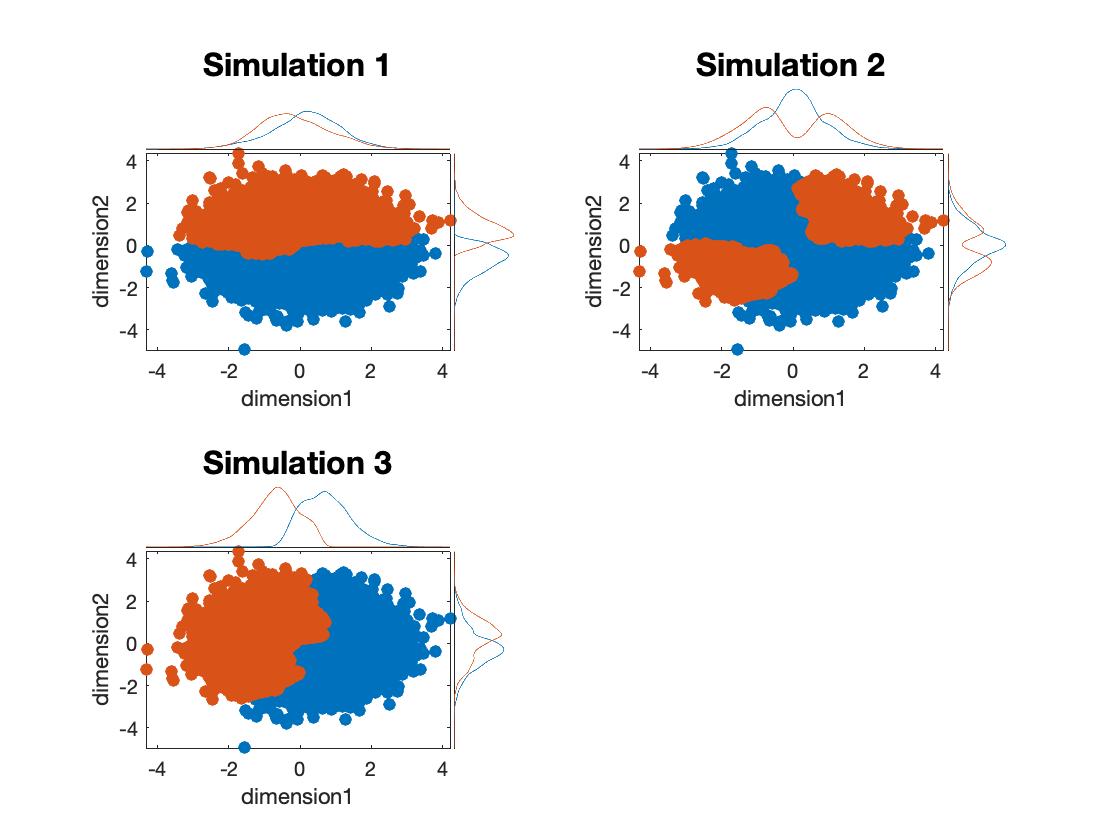}\\
            \includegraphics[width=\textwidth]{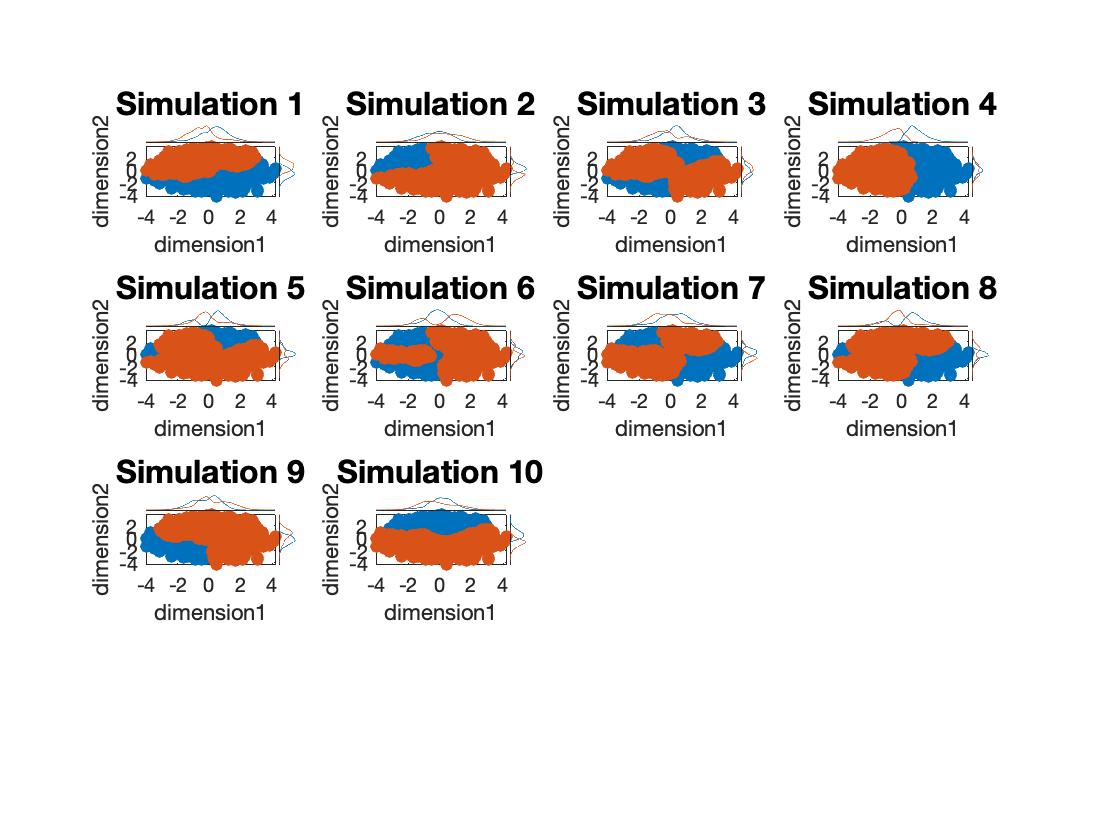}\\\\
        };
%
\draw[line width=0.5mm,blue!20] (a-2-1.north west) -- (a-2-1.north east);
\draw[line width=0.5mm,blue!20] (a-1-1.north) -- (a-1-1.south);
\end{tikzpicture}
\caption{We generate realistic datasets \cite{Gorriz19} including several modes by selecting a different number of clusters or sources generating the data sample, $N_c=\{2,4,6\}$, and setting two balanced classes in all possible configurations: $\# \text{simulations}=\frac{1}{2}\binom{N_c}{N_c/2}=\{1,3,10\}$. Bottom: three separable linear simulations 1,4,10; seven non separable linear simulations: 2, 3, 5-9.}
\label{fig:example3}
\end{figure*}

\begin{proposition} Multimodal data complexity.
Given a set of $N_c\geq h(\mathcal{F})$ points in $n$-space then the number of dichotomies $\{Z^+,Z^-\}$ each of size $N_c/2$ is $\binom{N_c}{N_c/2,N_c/2}$. Moreover, the number of linearly separable dichotomies is then $\binom{h}{h/2,h/2}$.

Proof: Assume a number, $N_c$, of points in $n$-space\footnote{They could be modelling $n$-dimensional Gaussian pdfs (clusters) from which samples are drawn with centroids randomly distributed ``far away'' from one another (non-overlapping) in the input space}. The VC dimension $h$ for linear classifiers establishes that out of $2^{N_c}$ different dichotomies, linear classifiers can shatter only $2^{n+1}$. The multinomial coefficient is the number of ways to put $m$ clusters into $k$ classes, $\binom{m}{m_1,m_2,\ldots,m_k}:=\frac{m!}{m_1!m_2!\ldots m_k!}$, where $m_j$ is the number of clusters in class $j$. Thus, if the clusters are grouped into balanced binary classes $\{0,1\}$ we have only $\binom{N_c}{N_c/2,N_c/2}$ different dichotomies instead of $2^{N_c}$. In the same manner, the number of ways to put $h$ separable clusters into a binary dichotomy of size $h/2$ follows the same rule.
\end{proposition}
For example, generating $6$ clusters ($3$ per class) in $n=2$ dimensions, we have $20$ different dichotomies or classification simulations out of $64$. Removing the inverse simulations (inverting labels) only $10$ remain and among them $7$ represent a problem that a linear classifier cannot shatter (those described by multimodal pdfs or subsets with intersecting convex hulls). This example is depicted in figures \ref{fig:example3bis} and \ref{fig:example3} (bottom) for $n=2$. This procedure allows us to select which simulated data samples represent a realistic data pool for subsequent simulations. In general, any data-dependent capacity measure detects if the sample is likely to be drawn from a multimodal pdf. In this case, we can visually/analytically identify the number of simulations following this realistic pattern.

\newpage

\subsection{Figures}

\begin{figure*}
\centering
\includegraphics[width=0.49\textwidth]{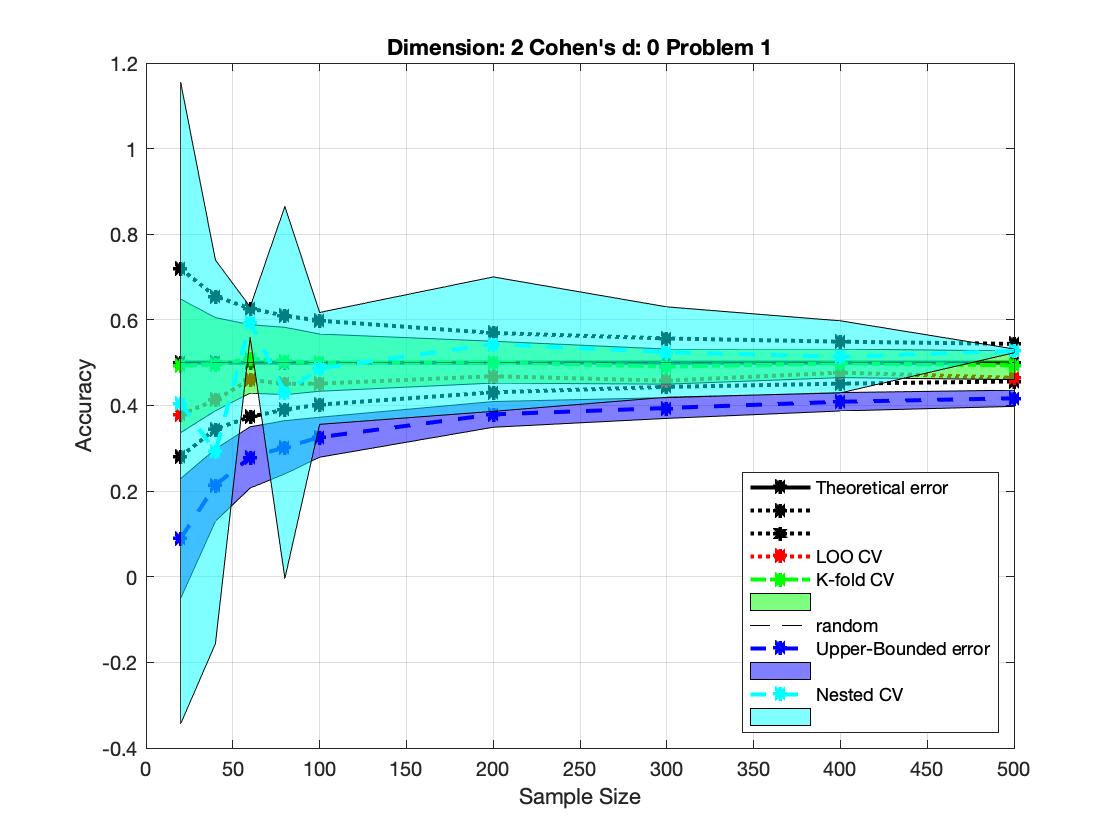}
\includegraphics[width=0.49\textwidth]{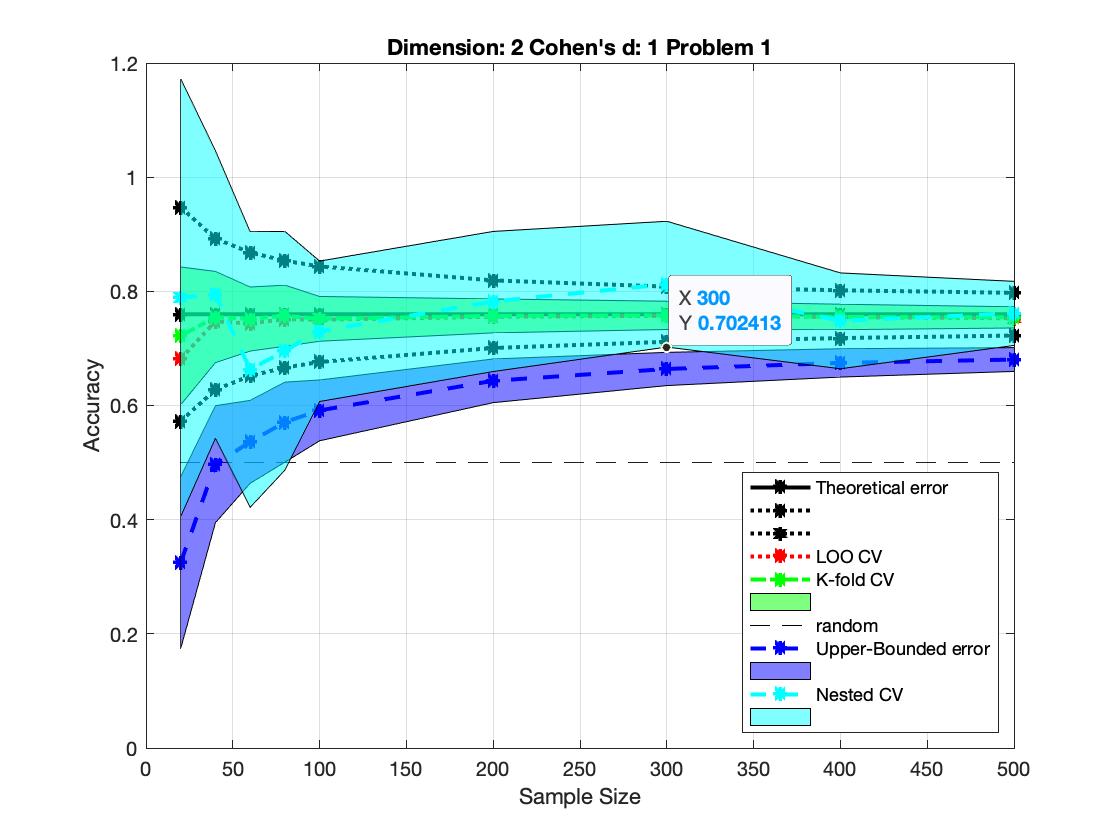}
\includegraphics[width=0.49\textwidth]{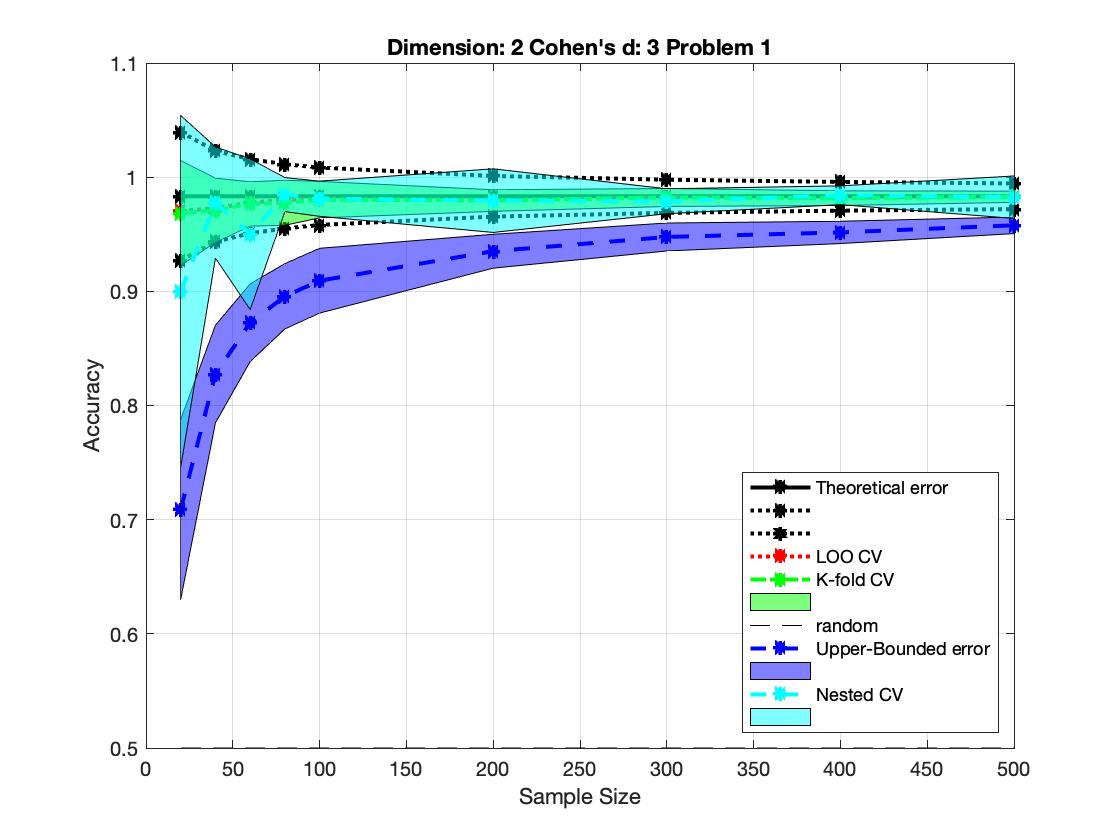}
\caption{Performance of nested CV, naive CV and the proposed K-fold CUBV test. We show the model-driven upper and lower bounds for the theoretical accuracy using 2D Gaussian data.}
\label{fig:nestedCV}
\end{figure*}

\begin{figure*}
\centering
\includegraphics[width=0.49\textwidth]{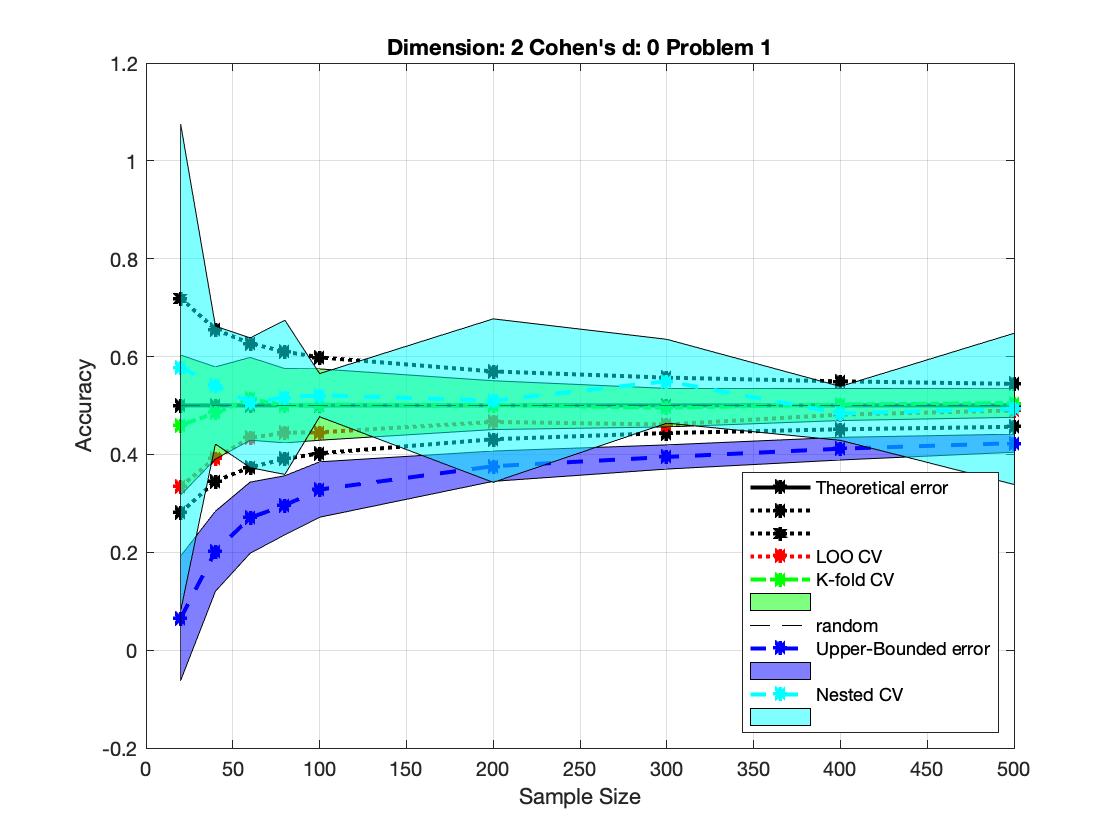}
\includegraphics[width=0.49\textwidth]{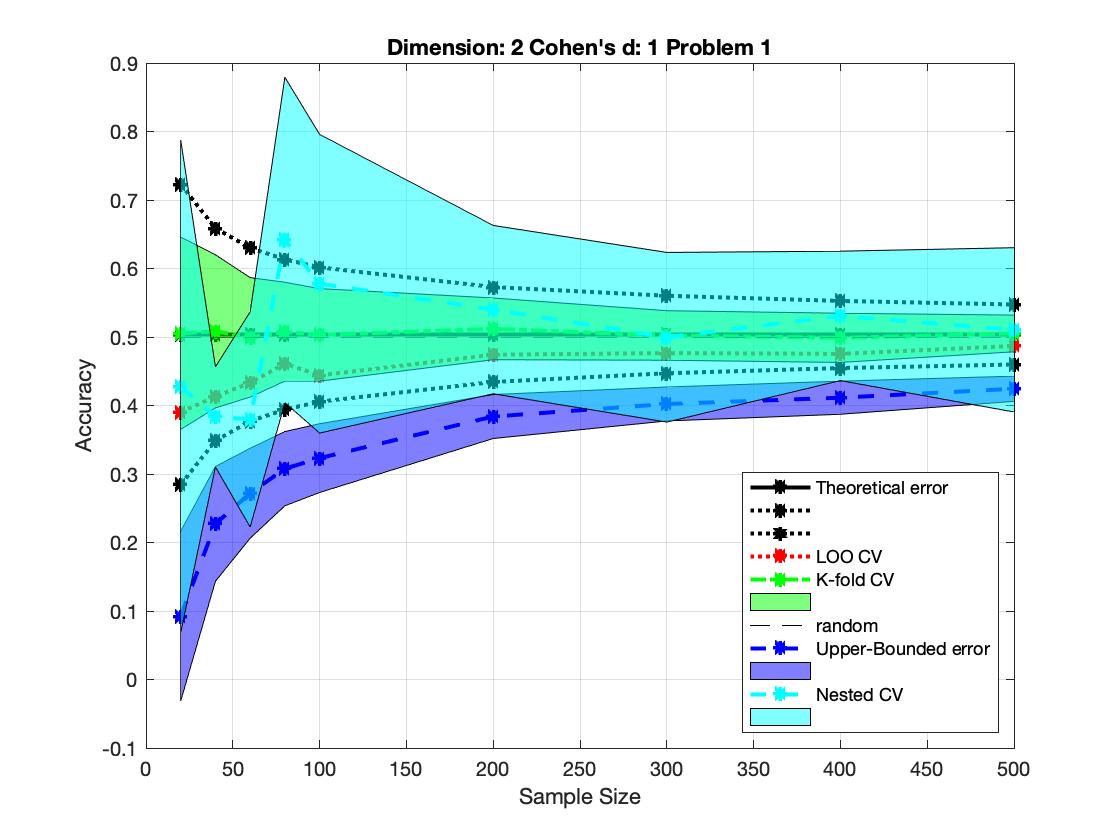}
\includegraphics[width=0.49\textwidth]{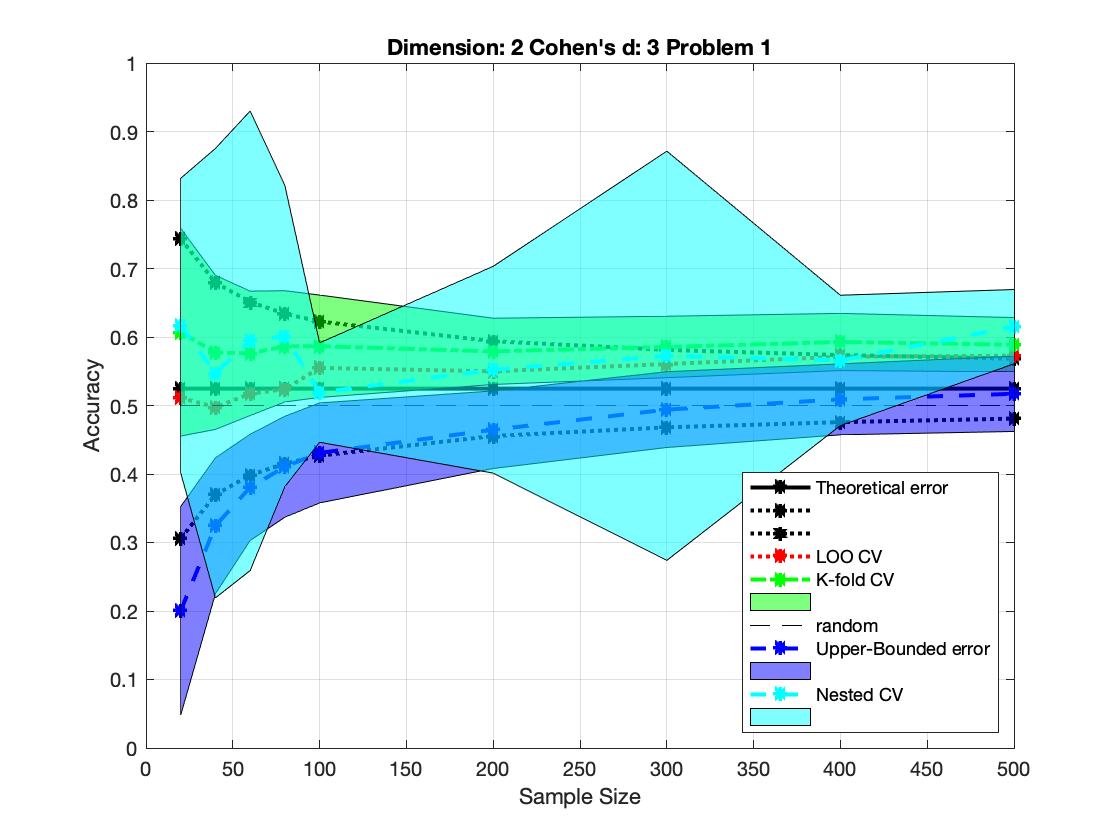}
\caption{Performance of nested CV, naive CV and the proposed K-fold CUBV test. We show the model-driven upper and lower bounds for the theoretical accuracy using 2D Gaussian data and 2 modes per cluster.}
\label{fig:nestedCV2}
\end{figure*}

\end{document}